\documentclass[nofootinbib]{AIAArevtex4}

\usepackage{amssymb,amsmath}
\usepackage{algorithm,algpseudocode}
\usepackage{color,xspace}
\usepackage{graphicx}
\usepackage{subfigure}
 \usepackage{bibentry}
\usepackage{enumerate}
\let\oldReturn\Return
\renewcommand{\Return}{\State\oldReturn}

\newtheorem{theorem}{Theorem}[section]
\newtheorem{lemma}[theorem]{Lemma}

\renewcommand{\theenumi}{(\roman{enumi})}

\newcommand{\real}{\mathbb{R}}

\newcommand{\setdef}[2]{\left\{#1 \; | \; #2 \right \}}
\newcommand{\realpositive}{\mathbb{R}_{>0}}

\newcommand{\realnonnegative}{\mathbb{R}_{\ge 0}}

\newcommand{\map}[3]{#1: #2 \rightarrow #3}

\newcommand{\oprocendsymbol}{\hbox{$\bullet$}}
\newcommand{\oprocend}{\relax\ifmmode\else\unskip\hfill\fi\oprocendsymbol}

\newcommand{\until}[1]{\{1,\dots, #1\}}
\newcommand{\argmin}{\operatorname{argmin}}

\newcommand{\stated}[3]{x_{#1 #2}(#3)}
\newcommand{\statepre}[3]{\bar{x}_{#1 #2}(#3)}
\newcommand{\statepost}[3]{\hat{x}_{#1 #2}(#3)}
\newcommand{\covd}[3]{P_{#1 #2}(#3)}
\newcommand{\covpre}[3]{\bar{P}_{#1 #2}(#3)}
\newcommand{\covpost}[3]{P_{#1 #2}(#3)}
\newcommand{\meas}[2]{y_{#1}(#2)}
\newcommand{\neigh}[1]{\mathcal{N}_{#1}}

\newcommand{\Exp}{\operatorname{E}}
\newcommand{\Cov}{\operatorname{Cov}}
\newcommand{\trace}{\operatorname{trace}}
\newcommand{\diag}{\operatorname{diag}}
\newcommand{\elem}[2]{\operatorname{elem}(#1,#2)}
\newcommand{\row}[2]{\operatorname{row}(#1,#2)}

\usepackage{setspace}
\begin{document}

\title{\Large Cooperative Robot Localization \\ Using Event-triggered Estimation}

\author{Michael Ouimet \footnote{Robotics Research Engineer, SPAWAR Systems Center Pacific, San Diego, CA 92152}} 
\affiliation{SPAWAR Systems Center Pacific, San Diego, CA 92152} 
		\author{David Iglesias \footnote{Graduate Research Assistant, Ann and H.J. Smead Aerospace Engineering Sciences Department, University of Colorado Boulder, Boulder, CO
  80309}}\affiliation{University of Colorado Boulder, Boulder, CO
  80309} 
	\author{Nisar Ahmed \footnote{Assistant
    Professor, Ann and H.J. Smead Aerospace Engineering Sciences Department, University of Colorado Boulder, Boulder, CO
  80309, AIAA Member}} \affiliation{University of Colorado Boulder, Boulder, CO
  80309} 
	\author{Sonia Mart\'inez \footnote{Professor, Department of Mechanical and
    Aerospace Engineering, University of California San
  Diego, La Jolla, CA 92093}} \affiliation{University of California San
  Diego,La Jolla, CA 92093}

\vspace{-0.25in}
\begin{abstract}
  This paper describes a novel communication-spare cooperative
  localization algorithm for a team of mobile unmanned robotic vehicles.  Exploiting an
  event-based estimation paradigm, robots only send measurements to
  neighbors when the expected innovation for state estimation is high.
  Since agents know the event-triggering condition for measurements to
  be sent, the lack of a measurement is thus also informative and
  fused into state estimates.  The robots use a Covariance
  Intersection (CI) mechanism to occasionally synchronize their local
  estimates of the full network state. In addition, heuristic
  balancing dynamics on the robots' CI-triggering thresholds ensure that,
  in large diameter networks, the local error covariances
  remains below desired bounds across the
  network.  
  Simulations on both linear and nonlinear dynamics/measurement models
  show that the event-triggering approach achieves nearly optimal
  state estimation performance in a wide range of operating
  conditions, even when using only a fraction of the communication
  cost required by conventional full data sharing.  The robustness of
  the proposed approach to lossy communications, as well as the
  relationship between network topology and CI-based synchronization
  requirements, are also examined.
\end{abstract}

\maketitle

\section{Introduction}
The decrease in the price and weight of robotic hardware (wireless
communication, sensor suites, actuators) can make possible the
autonomous deployment of large teams of unmanned aerospace vehicles in
surveillance, exploration, search-and-rescue, and cargo-transportation missions.
While today's technology has advanced tremendously, algorithms that
can endow robotic teams with the desired autonomy are still lacking.
In particular, the successful execution of higher-level tasks often
relies on accurate robot position information; e.g.~to be used in path
planning or decision-making routines. In the case that full position
information is unavailable, a relevant question is how vehicles can
exploit their partial access to information to produce joint
\textit{Cooperative Localization} (CL) algorithms.

The CL problem arises when mobile robots try to localize themselves with respect to
other mobile robots, whom they can also communicate with. 
In CL, robots typically obtain and share measured/estimated relative pose information to improve their own local pose estimates (which may possibly be augmented with the pose estimates of other robots to obtain a `moving map'). 
This approach has many close connections to the well-known simultaneous localization and mapping (SLAM) problem \cite{Thrun-ProbRoboticsBook-2005}, where in CL the `map features/landmarks' are dynamic and can also actively provide information to one another. 
As such, many different statistical state estimation techniques for \textit{decentralized CL} have been developed to most efficiently leverage the sensing and computing capabilities of multiple robots. 
However, state of the art CL algorithms make many significant trades in terms of required communication bandwidth, computational processing, localization accuracy, network topology constraints, and robustness to statistical inconsistencies (which can easily arise in networked distributed data fusion problems). In particular, the problem of minimizing required communication bandwidth and messaging frequency for CL in networks with arbitrary relative measurement and communication topologies remains quite challenging, as the localization performance, statistical robustness, and computational efficiency of each robot's state estimator depend heavily on these characteristics.  

\begin{figure}[htb]
  \centering
  \includegraphics[width=.5\linewidth]{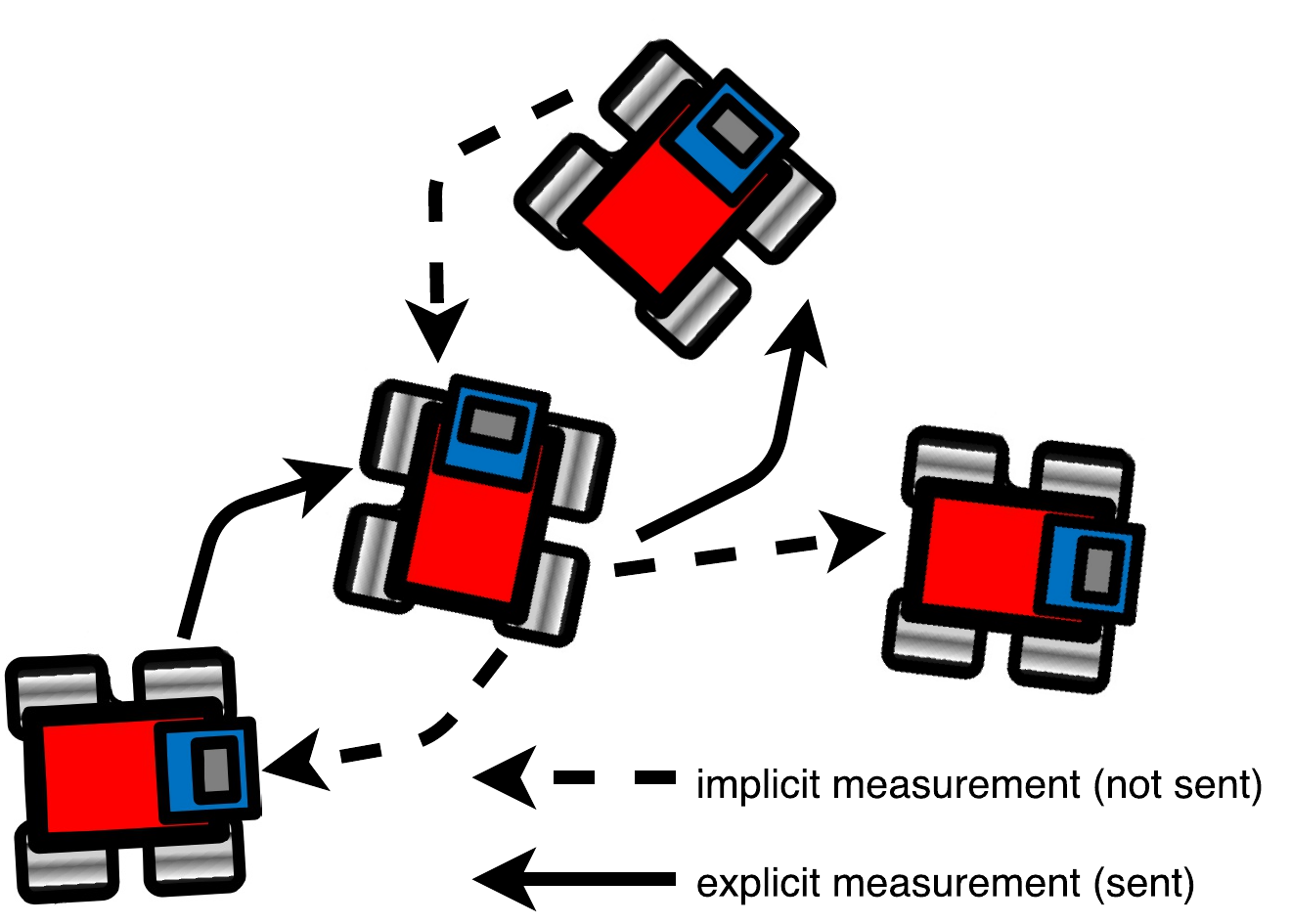}
\vspace{0ex}
  \caption{Event-triggered CL: {\it sufficiently
      informative} data is sent {\it explicitly}; otherwise, it is censored/{\it implicit
      data}. 
		}\label{fig:schematic}
\end{figure}

This work studies a novel Kalman filter-based CL strategy
that trades off estimation performance with communication cost. Using
an event-triggering strategy, the scheme determines when a robot
should pass {\it explicit measurements} to neighbors, and then
combines these with {\it implicit measurements} to produce a network
state estimate. As shown in Figure~\ref{fig:schematic}, this framework is
very applicable to unmanned ground vehicles (UGVs) operating in planetary exploration missions without GPS aiding,
unmanned aerial vehicles (UAVs) or space vehicles in GPS-denied environments, 
or unmanned underwater vehicles (UUVs). UUVs, for instance, 
which typically have inertial measurement units (IMUs) and GPS as well
as an acoustic modem for low-bandwidth communication.  The vehicles
receive GPS measurements only at the water surface, but they can get
time-of-flight ranging measurements from acoustic message
transmissions to one another while underway.  For best performance,
the network should share their occasional GPS measurements and regular
ranging measurements at every opportunity. However, transmitting data
underwater via acoustic waves requires considerably more energy than
using electromagnetic waves through the air.  By intelligently
controlling the rate with which the robots exchange explicit data
messages, the trade off between best estimation performance and
increased mission duration (or more bandwidth for other data) can be
balanced. 
\vspace{-1ex}

\vspace{-1ex}
\vspace{-1ex} This paper details a novel CL algorithm for a team of
robotic agents to estimate the state of the entire network.  Employing an
event-based approach, agents only send measurements to their
neighboring agents when the size of the measurement innovations
(i.e. the amount of `surprising' new sensor information) is large
relative to some known threshold. 
Because agents all know the event-triggering condition for
measurements to be sent explicitly, the lack of a measurement is
informative, and can thus be properly fused into state estimates as
`implicit data' using set-based observation updates. This results in a
\textit{decentralized event-triggered filter}, which each robot
reproduces locally while handling the explicit/implicit
measurements. In order to keep a local covariance error metric bounded
whenever they do not receive direct measurements from each other, the
robots occasionally combine their state estimates via Covariance
Intersection (CI) fusion over the whole network state, to fully
benefit from estimate correlations. When the communication network
diameter is large, fusion updates may still result in state estimation
uncertainties (covariances) that are larger than practically desirable
bounds for certain agents.
Thus, balancing dynamics are also defined on the individuals'
event-triggering thresholds, which allows better-informed robots to
aid poorly informed robots to achieve desired covariance bounds.
Simulations with linear and non-linear dynamical models and
measurements show that, under ideal communication conditions, the
event-triggered CL approach performs nearly as well as the
idealized fusion approach based on full measurement sharing, but
requires only a fraction of the communication cost. 
Simulations under non-ideal communication conditions (namely, with complete packet losses and incomplete observability/communications between each robot) show that the proposed event-triggered approach can still provide reasonably reliable and robust localization performance under certain circumstances. 

This paper is based on previous work presented by the authors
in~\cite{Ouimet-MFI-2015} and extends that work in several important
ways. First, this paper generalizes the formulation of the
event-triggered filter to non-linear dynamical systems, and
specifically examines the effectiveness of the filter in robot
networks defined by nonlinear Dubins dynamics and nonlinear range and
bearing measurements in simulations. Secondly, given that the
event-triggered algorithm assumes measurements expected but not
received were intentionally censored, this work also investigates the
sensitivity of the proposed algorithm's performance to random
communication losses between agents. Finally, this paper includes
simulations on various communication topologies to examine relative
performance and necessity of sporadic CI-based synchronization in the
proposed algorithm.



\section{Literature review and related work}

\subsection{Related work: Multi-robot SLAM}
Much research in the robotics and autonomous vehicles literature has focused heavily on localization and mapping problems, with simultaneous localization and mapping (SLAM) being an important special case \cite{Thrun-ProbRoboticsBook-2005}. In SLAM, a robot tracks its own pose in an unknown map, represented by landmarks (or more general map features) whose locations are sensed and tracked online via relative sensing data. Following seminal work in the 1990s and early 2000s \cite{JH-HFDW:91, GD-PN-HFDW-SC-MC:01}, the statistical state estimation framework for SLAM soon became widely adopted, and matured quite rapidly to accommodate a wide variety of robotic sensors and vehicle types. 

Single robot SLAM algorithms are especially mature, and can be generally categorized as either filter-based algorithms or batch methods. Filter-based methods focus on obtaining estimates of an environmental map and the most current robot state only. These are typically based on the Kalman filter variants, such as the extended Kalman filter (EKF) or unscented Kalman filter (UKF) \cite{Holmes-ICRA-2008}, or approximations to the more general Bayes filter (e.g. particle filters \cite{Thrun-AI-2001, Montemerlo-ICRA-2003, Grisetti-RAS-2007}, Gaussian sum filters \cite{Kwok-ICRA-2005}). 
In contrast, batch methods seek estimates of the map along with the robot's entire state trajectory over time, and tackle the problems of robot state filtering and smoothing together. Batch methods are typically formulated as non-linear optimization problems, and efficiently implemented as non-linear least squares, maximum likelihood or Bayesian maximum a posteriori (MAP) estimators via specialized solvers \cite{Sunderhauf-IROS-2012, Mur-TRO-2015}. Filter-based methods are often computationally cheaper to implement and can be implemented online through recursive propagation of state-map estimates and uncertainties (covariances). Batch methods are more demanding computationally, but can provide very high fidelity maps and state estimates, since non-linear motion/sensing models, dense data sources (e.g. camera imagery), and subtle constraints such as loop closure can be easily handled. A key assumption for tractability in both filter-based and batch single robot SLAM methods is that the map itself always remains observable and static, i.e. only the robot is assumed to move, while the map features do not change or move. 

Multi-robot SLAM methods draw heavily from single robot SLAM techniques, but also offer a richer set of strategies for exploiting the sensing and computing abilities of multiple robots. Ref. \cite{PR-MA-BK-MK:16}, for instance, describes a distributed approach in which a single processing hub is responsible for fusing sensor data from multiple robots into a consistent map; this hub then acts as a global map server for each robot, which periodically downlinks the updated map to correct its private onboard SLAM solution. Alternatively, completely decentralized approaches such as those described in \cite{Jones-Allerton-2011} and \cite{Cunningham-ICRA-2013} can also be used, where each robot remains fully capable of performing SLAM on its own but occasionally fuses its estimated map with its neighbors' maps. As with single robot SLAM methods, these multi-robot SLAM techniques also rely on the persistent availability of static map features; more importantly, all robots must be able to sense, identify and reason over a common set of features.  
An important drawback of these and other similar approaches is the high computational expense of marginalizing out robot poses to communicate and fuse large scale maps. 
Such approaches must also contend with issues of landmark/feature association and coordinate frame alignment across multiple partially overlapping maps \cite{Indelman-CSM-2016}.  

\subsection{Multi-robot Cooperative Localization}
Cooperative Localization (CL)~\cite{SIR-GAB:02} describes
methodologies where robots aim to localize themselves with respect to
a combination of moving and static objects/robots.  In CL, a team of
mobile robotic agents, each with equivalent computing and sensing
capabilities, use relative distance measurements with respect to each
other to jointly estimate the pose of the network, resulting in a more
accurate state estimate.  
Loosely speaking, the CL problem can be framed as `multi-robot SLAM with moving landmarks'. 
Due to the coupling between the estimates of all agents' positions, this technique largely benefits from sporadic
GPS (or other inertial) measurements, effectively pinning down the relative position information into absolute global coordinate frame
positions. If robots are truly cooperative (i.e. part of the same team) and can actively communicate with one another, then the problem of landmark/feature association can also be solved more easily. Furthermore, rather than requiring robots to maintain estimates of an entire map, robots need only maintain position estimates of (some/all) other robots. Robots can thus can be opportunistic about how and when to share information to improve each other's localization performance. 
As in multi-robot SLAM, the distributed nature of CL also affords many opportunities to exploit the sensing and computing capabilities of multiple robots for improved scalability. 
While centralized single-point processing approaches to CL could be considered (wherein all robots send their information to a single data fusion center), these are generally vulnerable to communication dropouts as well as failure of the central processing point. As such, distributed CL algorithms that exploit decentralized local processing at each robot are the focus here. 

Distributed CL methods can be broadly categorized as either \textit{weakly coupled} or \textit{strongly coupled}, depending on how the state estimator for each robot is configured. In strongly coupled methods, each robot augments its own pose estimate with pose estimates of some other robots (any subset of the full network). That is, each robot explicitly maintains a `moving map' of other robots, in order to fully track and exploit all dependencies on other robot states which get updated via relative measurement and/or state estimate exchanges. 
In weakly coupled methods, each robot does not augment its state estimator with other vehicle states, but rather only estimates its own pose with relative measurement and/or state estimate exchanges. These methods avoid the full computational overhead of tracking a moving robot map, but are often less accurate and `lossy' compared to strongly coupled methods, since they do not fully exploit local statistical correlations between neighboring robots that develop over time. 
In both sets of methods, different trades are made in terms of balancing requirements for communication bandwidth and onboard computing, as well as in terms of addressing the more subtle issue of statistical consistency. 
Consistency issues arise in the CL problem if one or more robots in the network fail to account for dependencies between different robot state estimates and/or measurements. 
This can lead to the `data incest' problem, were information is incorrectly double-counted such that state estimates become incorrectly overconfident in the presence of large errors throughout the network. 
Since exact tracking of information dependencies is generally an intractable problem for multi-robot networks (except in special cases) \cite{Campbell-CSM-2016}, both strongly and weakly coupled methods use conservative data fusion approximations like Covariance Intersection (CI) \cite{SJ-JU:07} or other alternatives to address this issue. 
In the following, several examples of strongly coupled and weakly coupled distributed CL algorithms are highlighted. 
While not exhaustive, this brief review is meant to provide a representative sample of existing CL strategies. 

\subsubsection{Strongly Coupled Methods:} Many examples of strongly coupled CL algorithms can be found in the aerospace literature for vehicle formation control; these approaches are often based on decentralized Kalman/information filters for joint minimum mean squared error (MMSE) network state estimation \cite{Ferguson-GNC-2003, Smith-IET-2007,McLoughlin-JGCD-2007}. 
Filtering-based CL algorithms are particularly amenable to vehicle formation problems, since good priors can usually be determined to
model the dynamics and control inputs of each vehicle in the team, and since the problem can be easily solved in a relative frame of
reference with respect to designated `lead' vehicles. 
However, such conditions can be relaxed for strongly coupled CL, and approximations to decentralized MMSE state estimators can also be used. For instance, Arambel, et al. \cite{POA-CR-RKM:01} developed a CI-based sub-optimal conservative data fusion strategy for cooperative satellite localization, where each satellite tracks it own pose and velocities relative to inertial frame, along with separation distances and rates relative to all other vehicles. 

A variety of other strongly coupled CL algorithms have also been developed in the robotics and sensor networks literature. 
Ref. \cite{SIR-GAB:02} describes a strongly coupled algorithm in which robots use a multi-centralized filtering algorithm to broadcast measurements one at a time to the whole network. Leung, et al. developed an algorithm that guarantees equivalence to an idealized centralized state estimator for all robots \cite{KYKL-TDB-HHTL:10}. In this approach, robots must only consider their knowledge of the full network topology and share appropriate local information with neighbors (including odometry data, relative position measurements and/or state estimates) as determined by temporal `checkpoints', which helps ensure that each robot's state estimators remain statistically consistent. While this approach is robust to dynamic topologies without guarantees of full connectivity, it also requires high volume information transfer between robots to exchange sensor measurements and state estimator outputs (means and covariances for robot poses). 
In contrast, ref. \cite{NK-FC-RA-RC:06} describes an approach based only on local state estimator output exchange between vehicles. Each robot in this approach maintains two sets of extended Kalman filter state estimates for the entire robot network: one that can be updated by direct fusion with state estimates provided by other robots (but cannot be shared further with other vehicles); and another that can be updated by the ego robot's sensor measurements and shared with other robots. This exploits the fact that Kalman filter mean and covariance estimates summarize all information from previously fused sensor measurements; the use of two filters on each robot avoids information double-counting, and thus ensures statistical consistency. However, this approach still imposes heavy communication requirements between robots. 

To address excessive communication overhead, ref.\cite{NT-SIR-GBG:09} develops an alternative measurement-only exchange approach in which real-valued sensor data are quantized into binary messages. For example, using 1 bit to represent relative measurements, it is shown that `sign of innovation' EKF and MAP estimators can be obtained for each robot that provide reasonable estimation performance in many situations. This approach is perhaps most closely related qualitatively to the event-triggered approach proposed here, with key differences being that: (i)  event-triggering does not depend on the sign of innovation, but rather on expected values and covariances of the innovations (which in turn are a function of motion and sensor models); and (ii) rather than always transmitting a fixed-size quantized measurement, an event-triggering sender robot either sends no data (0 bits) or full real-valued data to the receiver robot. 

\subsubsection{Weakly Coupled Methods:} 
Ref. \cite{SEW-JMW-LLW-RME:13} formulates a relatively simple `one-way fusion' approach in which a designated vehicle (server) broadcasts its position estimates to other robots (clients), which then update their pose estimates via extended information filters. Kia, et al. ~\cite{SSK-SFR-SM:14} describe a more sophisticated decentralized EKF approach in which mobile robots with correlated states exchange different sets of variables to update their local correlations, only when there is a new relative measurement between them (rather than at each time step). 
Ihler, et al. \cite{ATI-JWF-RLM-ASW:05} exploited a similar idea for node self-localization in static sensor networks, by using probabilistic nonparametric (Monte Carlo) belief propagation to fuse non-linear relative position measurements with respect to other network nodes. 
The main drawback of these approaches is that they require restricted network topologies for interagent measurements and communications, to guarantee correct convergence and avoid data incest issues. 

Other weakly coupled CL methods overcome these limitations by sharing both measurement data and state estimates together. 
In \cite{AB-MRW-JJL:09}, each robot maintains two independent EKF tracks for each other robot it communicates with: one in which its own pose estimate is not updated with information received from the other robot, and another in which its pose is updated by this information. By exchanging relative measurements and pose information from both EKFs, and by tracking the pedigree of information received from all other agents, pairs of robots can be guaranteed to avoid data incest. 
Approaches that use other strategies to simplify each robot's state estimator requirements have also been explored.  Prorok, et al. \cite{AP-AB-AM:12} developed a particle filtering approach in which each robot shares with detected neighbors relative range/bearing measurements and compressed sets of local particle estimates of their own states. 
Knuth and Barooah \cite{JK-PB:09} and more recently Luft, et al \cite{LL-TS-SIR-WB:16} developed similar strategies using parametric EKFs; in the former, robots opportunistically exchange relative pose measurements and pose estimates at each time step, while robots in the latter also maintain estimates of cross correlations of pose estimation errors for other robots that have been communicated with previously. 
In these approaches, each robot uses the received information to derive sub-optimal approximate statistical updates for its own pose estimate. 
In many cases, however, these methods require expensive fusion update operations for each robot's state estimate and lead to message sizes that do not scale well to large networks. 

Other weakly coupled techniques have been developed to address scaling and data incest issues simultaneously, using techniques that are guaranteed to avoid data incest without extra bookkeeping or topology restrictions. \v Curn, et al. \cite{JC-DM-NO-VC:13} developed a CL approach based on the Common Past Invariant Ensemble Kalman filter (CPI-EnKF), a variant of the Monte Carlo-based Ensemble Kalman filter that is widely used for large scale data assimilation. Hao and Nashashibi \cite{HL-FN:13} developed a method based on Split Covariance Intersection, which addresses the over-conservativeness of classical CI by decomposing covariances for each robot's pose into two exclusive components: one independent of and another dependent on relative measurement information sent by other robots. 
Refs. ~\cite{LCCA-EDN-JLG-SIR:13} and \cite{HM-DGE:14} describe yet another related approach, where a robot uses its own state estimate to send a `pseudo state estimate' to the another robot involved in a relative measurement; the receiving robot uses CI to fuse its current estimate with the new pseudo estimate in a statistically conservative (but memoryless) manner. 
This approach is applicable to dynamic/ad hoc network topologies and only requires robots to estimate their own poses; this results in lower computational complexity, but at the expense of sacrificing information from joint estimate correlations. 

\subsubsection{Batch CL methods:} The majority of techniques developed for CL are filter-based. However, batch estimation methods for both weakly and strongly coupled decentralized CL can also been considered. 
In early work, Kurazume, et al. proposed a `cooperative positioning system', in
which a subset of robots in the network would take turns to
deliberately remain stationary in order to act as static landmarks for
the other mobile robots, which then fuse their relative
measurements via batch least squares \cite{Kurazume-ICRA-1996, Kurazume-ICRA-1998}.  
Building on the success of batch SLAM techniques, Nerurkar, et al. devised a distributed MAP CL estimator for multiple moving vehicles, using distributed data allocation and distributed conjugate gradient optimization methods \cite{EDN-SIR-AM:09}. 
Ref. \cite{Paull-ICRA-2015} proposed an alternative decentralized batch CL approach for GPS-denied underwater vehicle localization based on a dual factor graph formulation, which accounts for inter-vehicle localization on top of local vehicle SLAM. 
Ref. \cite{Wymeersch-ProcIEEE-2009} also considered a factor graph CL approach based on wireless network
signaling. While factor graphs and other related modern batch approaches often cope well with highly non-linear dynamics and measurement models, they require sophisticated optimization solvers and are often difficult to implement online for constrained onboard computing hardware. As such, filtering-based CL is the focus of this work. 
 
%
\subsection{Sparse communications and event-triggered algorithms}
A major feature of both weakly and strongly coupled CL algorithms
developed to date is their reliance on frequent explicit information
sharing between agents. However, this also leads to significant practical limitations in multi-robot systems,
since it requires each robot to expend considerable energy and consume
significant communication bandwidth to maintain good CL performance.
For applications such as those involving UUVs, this limitation can be
particularly costly and severe, and motivates consideration of
alternative cooperative state estimation strategies that require less
energy expenditure and communication bandwidth.

Recent work in the networked controls and cyberphysical systems literature has shown that \emph{event-triggered algorithms} are an attractive means for significantly reducing communication overhead in distributed control and estimation problems. 
The key idea is that, unlike typical periodic (time-triggered) control and estimation algorithms that require explicit information transmission and acquisition at regular time intervals, event-triggered algorithms require explicit information transmission only on the occurrence of certain events -- usually if some signal related to states of interest fall outside a certain threshold. Hence, information is only communicated when needed and as needed, and control/estimation routines can still be performed in the absence of explicit information being sent/received by processors in the network -- since the lack of any explicit information being received is itself still an informative event (provided all processors are aware of each other's event-triggering thresholds).   

Algorithms for centralized event-triggered estimation in distributed systems are fairly well established, e.g. see the survey in \cite{Liu-SSCE-2014} and the study in \cite{Trimpe-EBCCSP-2015} for a comparison of possible thresholding mechanisms. Yet, theory and techniques for \textit{decentralized} event-triggered estimation are not as well developed.  
In particular, the literature to date on distributed event-triggered estimation deals with either theoretical estimation error analysis for linear systems \cite{Yan-NonlinearDyn-2014, Trimpe-CDC-2014}, or one of the following application areas (with emphasis again on linear dynamics and sensor models): 
(i) wireless sensor networks where a single centralized state estimator receives measurements from multiple sensor nodes \cite{JS-ML:09, DS-TC-LS:14, Liu-TAC-2015}; 
(ii) control networks where multiple sensors/estimators communicate with each other over a common bus \cite{Trimpe-IFAC-2011, Trimpe-TAC-2014, Muehlebach-ACC-2015}; or 
(iii) consensus-based Kalman filtering algorithms \cite{Battistelli-ACC-2016, Li-IET-2016}. 
By definition, area (i) is not directly applicable to the decentralized CL problem considered here. Methods in area (ii) make the critical assumption that the communication bus is always reliable and that all data on the bus is accessible to each distributed state estimator, and thus is not generally well-suited to robotic CL problems. While the consensus methods of area (iii) could in principle be adapted to robotic CL, these generally tend to have intrinsically higher communication overhead and worse estimation accuracy than other forms of decentralized data fusion \cite{Chong-FUSION-2016}. 

A natural starting point to develop event-triggered algorithms for decentralized CL is to adapt existing centralized event-triggered methods.  
For instance, refs. ~\cite{JS-ML:09, DS-TC-LS:14} develop event-triggered estimation algorithms in the context of distributed sensing for a single partially observable dynamic process model, where a single centralized fusion center fuses the explicit and implicit measurements sent to it. In this paper, the centralized event-triggered estimation techniques from these works are adapted and evolved further into a novel strongly coupled decentralized CL framework, in which multiple partially observable dynamic processes (robot pose states) are estimated simultaneously.  
%

\section{Technical Preliminaries}

To build up to our formal problem statement
(Section~\ref{se:prob-stat}) and technical approach (Section~\ref{se:ETCL}), this part presents useful notation and
concepts from probability theory.

Let $\real$ be the set of real numbers and $\realnonnegative$ the set
of positive or zero real numbers.  For a vector $v \in \real^d$, $d>
0$, let $\diag(v) \in \real^{d \times d}$ be the diagonal matrix with
vector $v$ along the diagonal. 
Let $|v|$ be the Euclidean $2$-norm of $v$. For a set of
matrices $\{M_i\}_{i \in \until{N}}$, each in $\real^d \times
\real^d$, let $\diag(M_i)$ be the block-diagonal matrix where the
$i$-th diagonal block is matrix $M_i$. The matrix $I$ is the identity
matrix of the appropriate size.

Consider a multi-variate random variable $Z$, and let
$\map{f}{\real^d}{\realnonnegative}$ denote its probability density
function (pdf). Let $\Exp(Z) = \int_{\real^d} z f(z) \text{d} z$ and
$\Cov(Z) = \Exp[(z - \Exp(Z))(z - \Exp(Z))^T]$ denote its expected
value and covariance, respectively.  Given $Z$ with pdf $f$ and a
subset $\Omega \subseteq \real^d$, define a conditional pdf
$\map{f_{\Omega}}{\real^d}{\real}$ as $f_\Omega(z) =f(z \,|\, z \in
\Omega)$; that is, $f_\Omega(x) = \frac{f(x)
  1_{\Omega}(x)}{P(\Omega)}$, where $1_\Omega$ is the indicator
function and $P$ is the probability mass of $\Omega$ computed using
$f$. With a slight abuse of notation, the conditional random variable
with pdf $f_\Omega$ is referred to as the ($f$-distributed) random
variable $Z$ with truncated support $\Omega$.
The one-dimensional normal probability density function,
$\map{\phi}{\real}{\realnonnegative}$, is defined as $ \phi(z) =
\frac{1}{\sqrt{2 \pi}} \text{exp}(-\frac{1}{2}z^2),$ $ z \in \real.$
The normal distribution's tail probability, i.e., the probability that
the random variable is larger than a quantity $z$, is denoted by
$\map{\mathbf{Q}}{\real}{[0,1]}$, and defined as
$\mathbf{Q}(z) = \int_{z}^{+\infty}\phi(s)\text{d}s. $

The next result~\cite{NLJ-SK-NB:95} describes how the statistics of a
Gaussian random variable change when conditioned on having a bounded
support interval. 
\begin{lemma} \label{le:truncGauss} Let $Z$ be a univariate Gaussian
  random variable with mean $\mu$ and variance
  $\sigma^2$. Then, the mean and variance of $Z$
  with truncated support $\Omega = \setdef{z \in \real}{a\leq z \leq
    b}$ satisfy $\Exp(Z) = \mu + \bar{z}$ and $\Cov(Z) = (1 -
  \vartheta)\sigma^2$, where 
\begin{align*}
  \bar{z} & = \frac{ \phi( \frac{a -\mu}{\sigma}) - \phi( \frac{b
      -\mu}{\sigma}) }{ \mathbf{Q}( \frac{a -\mu}{\sigma}) -
    \mathbf{Q}( \frac{b -\mu}{\sigma})}\sigma\,,
\end{align*}
and 
\vspace{-2ex}
\begin{align*}
  \qquad \quad \vartheta = \Big(\frac{ \phi( \frac{a -\mu}{\sigma}) -
    \phi( \frac{b -\mu}{\sigma}) }{ \mathbf{Q}( \frac{a -\mu}{\sigma})
    - \mathbf{Q}( \frac{b -\mu}{\sigma})} \Big)^2 - \frac{ (\frac{a
      -\mu}{\sigma})\phi( \frac{a -\mu}{\sigma}) - (\frac{b
      -\mu}{\sigma})\phi( \frac{b -\mu}{\sigma}) } {\mathbf{Q}(
    \frac{a -\mu}{\sigma}) - \mathbf{Q}( \frac{b
      -\mu}{\sigma})}\,. \qquad \oprocend
\end{align*}
\end{lemma}

\section{The Cooperative Localization Problem}\label{se:prob-stat}
This section formally describes the robots' dynamical model, sensing,
and communication models, and the CL problem. 
The fundamental concepts behind event-triggered state estimation with
innovation thresholds are then presented.

\subsection{Robot dynamics}
Assume $N$ robots are moving in a particular environment and that
their dynamics are modeled as a discrete-time nonlinear system. In
this way, at each time step $k$,
\begin{align}
  x_{i}(k+1) &= f_i(x_i(k),u_i(k)) + w_i(k),
\label{eq:nl-dynamics}
\end{align}
where $x_i \in \real^d$ represents the state of robot $i$, $f_i$ is
the (nonlinear) discrete-time state transition function, and $u_i \in
\real^n$ is its control input, for $i \in \until{N}$.  Here, the noise
$w_i(k)$ is assumed to be normally distributed with zero mean and
covariance $Q_i(k)$, and uncorrelated in time. One may linearize the
nonlinear dynamics model $f_i$ about previous estimates to obtain the
system matrices $A_i(k), B_i(k)$, for all $k \ge 0$, see
e.g.~\cite{SSK-SFR-SM:14}.  Here, $A_i(k) = \frac{d f}{d
  x_i}(\hat{x}_i(k))$ and $B_i(k) = \frac{d f}{d u_i}(\hat{x}_i(k))$
are the Jacobians evaluated at the current best state estimate
$\hat{x}_i(k)$.  This gives the time-varying linear model
\begin{align*}
  x_{i}(k+1) \approx A_i(k) x_i(k) + B_i(k) u_i(k) + w_i(k), \qquad i
  \in \until{N}.
\end{align*}
 
%
For each robot pair $i$ and $j$, the corresponding process noises are
assumed mutually uncorrelated. Let $x(k) \in \real^{Nd}$ be the vector
of all robots' locations. It is further assumed that the robots are
realizing a cooperative task with known assigned motions, and so, the
control strategies of all robots are known.  When performing
Extended-Kalman-Filter (EKF) updates for state estimation, the
nonlinear model \eqref{eq:nl-dynamics} is used to propagate the state
estimate during the prediction step, while $A_i$ and $Q_i$ are used to
propagate the covariance in the prediction
step~\cite{YBS-XRL-TK:01}.  

\subsection{Sensing and communication}\label{se:sensing-comms}
Robot $i$ can occasionally take a set of $m_i$ local scalar
measurements stacked into $y_i \in \real^{m_i}$,
\begin{align*}
y_i(k) = h_i(x(k)) + v_i(k), \qquad i \in \until{N}.
\end{align*}
These may include the components of the relative position between
itself and the robots in its view (i.e.~nonlinear range and bearing),
as well as absolute position~measurements (i.e.~GPS).  When performing
EKF updates for state estimation, the measurement model is linearized
to compute measurement innovation covariances, using $C_i(k) = \frac{d
  h_i}{d x}(\hat{x}(k))$, giving
\begin{align*}
y_i(k) \approx C_i(k)x(k) + v_i(k), 
\end{align*}
where $C_i(k) \in \real^{m_i \times Nd}$, $i\in \until{N}$, is the
measurement matrix that transforms the state into measurement space.
The measurement noise $v_i(k)$ is normally distributed with zero mean,
diagonal covariance $R_i(k)$, and uncorrelated in time. The
measurement noise is uncorrelated to the process noise, and
uncorrelated with other measurement noises produced by different
robots. It is assumed throughout that outliers and corrupt data from invalid 
measurements are discarded or not present.  

A point-to-point communication protocol is assumed to be running on
the network that efficiently allows each robot to know which other
robots it can communicate~with (i.e.~the desired network topology is
known and fixed in advance).  The subset of robots that $i$ can
communicate with is denoted as
$\neigh{i}$. 
If robot $i$ can communicate with robot $j$, then it is assumed that
$j$ can also communicate with $i$.  Furthermore, it is also assumed
that robots obtain relative position measurements of each other only
if they can communicate with each other.  Finally, the neighbor set
for each robot remains
fixed, 
such that the number of neighbors for each robot is small compared to
the size of the full network (i.e.~$O(1)$ as opposed to $O(N)$) to
limit the number of messages exchanged and processed at each step. For
simplicity, messages are assumed to be sent and received synchronously
within a given time step, up to some fixed range $r_c$ between robots.

\subsection{Cooperative localization}\label{se:CL}
For each robot $i \in \until{N}$, the current state and covariance
estimate of a subset of $N_i$ other robots in the team at time $k$ is
denoted by $(\statepost{i}{i}{k},\covd{i}{i}{k}) \in \real^{{N_i}d}
\times \real^{{N_i}d\times {N_i}d}$. For simplicity, let $N_i = N$,
the total number of robots, for all robots $i$; that is, assume each
robot is interested in the state of all robots in the network, even if
only a subset of all $N$ robots are in $\neigh{i}$.  

In the CL problem considered here, robot $i$ shares some or all of its
local measurement vector $y_i(k)$ with robots in $\neigh{i}$. Each
robot fuses information received from its local sensors and other
robots to minimize $\trace(\covd{i}{i}{k})$, $i \in \until{N}$. Since
measurements are not broadcast to everyone, each robot will obtain a
different network state estimate.  In particular, exchanging relative
range and bearing measurements between robots $i$ and $j$ causes the
$(i,j)$ block elements of $\covd{i}{i}{k}$ and $\covd{j}{j}{k}$ to
become non-zero, due to the correlation of estimation errors for $i$'s
and $j$'s states in both $\statepost{i}{i}{k}$ and
$\statepost{j}{j}{k}$. On the other hand, if a third robot $r$ never
processes relative measurements between $i$ and $j$, then the $(i,j)$
block of $\covd{r}{r}{k}$ remains zero (assuming $r$ has no prior
information about this correlation).
%
%
If robot $i$ receives an absolute position measurement, improves its
absolute state estimate, and shares the measurement with its
neighbors, the cross-correlations allow the members of $\neigh{i}$ to
improve their local own state estimates as well. Robot $i$ updates
$(\statepost{i}{i}{k},\covd{i}{i}{k})$ by fusing data $y_i(k)$ from
its onboard sensors with measurements $y_j(k)$ received directly from
neighbors $j \in \neigh{i}$. It should also be noted that robots do
not `forward' data received from neighbors to other neighbors, i.e. if
$r \in \neigh{j}$ but $r \notin \neigh{i}$, then robot $i$ does not
receive any $y_r(k)$ data via $j$, and, likewise, robot $r$ does not
receive any $y_i(k)$ data via $j$ (although both $r$ and $i$ will
receive $y_j(k)$ data, including any relative ranges and bearings to
both $r$ and $i$ from $j$).

In this paper, standard Extended-Kalman-filter (EKF) state prediction
and measurement updates are used to fuse measurements
sequentially. The generic form of these updates are shown in
Algorithms~\ref{algo:KFpred} and~\ref{algo:KFmeas}; the latter is
applied in turn to each $y$ in a measurement queue for any particular
robot's local state estimator, by computing the Kalman gain
$\covpost{}{}{k}C(k)^T(C(k)\covpost{}{}{k}C(k)^T +R(k))^{-1}$ and
measurement innovation $z=(y- h(\statepost{}{}{k}))$ for each
$y$. 
This work extends the notion of CL to handle situations where the
robots use an event-triggering rule to decide when to send
measurements to neighbors.

\begin{algorithm}[htb]
  \caption{Generic Kalman Filter Dynamics Prediction Update }
  \label{algo:KFpred}
  \begin{algorithmic}[1]
\Require $\statepost{}{}{k-1}, \covpost{}{}{k-1}, f$, $u(k-1)$ and $Q(k-1)$
\State $\statepre{}{}{k} \leftarrow f(\statepost{}{}{k-1}, u(k-1))$
\State $\covpost{}{}{k} \leftarrow  A(k-1)\covpost{}{}{k-1}A(k-1)^T +
Q(k-1)$
\Return  $\statepre{}{}{k},\covpre{}{}{k}$
  \end{algorithmic}
\end{algorithm}

\begin{algorithm}[htb]
  \caption{Generic Sequential Kalman Filter Measurement Update}
  \label{algo:KFmeas}
  \begin{algorithmic}[1]
    \Require $\statepost{}{}{k}, \covpost{}{}{k}, C(k), R(k),$ and $y$
    (measurement obtained locally or from neighbor) \\
    \If{$y$ is first measurement fused at time $k$}
    \State {$\statepost{}{}{k} \leftarrow \statepre{}{}{k}$, \ \
      $\covpost{}{}{k} \leftarrow \covpre{}{}{k}$} \ \ (from
    prediction update)
\EndIf
\State $\statepost{}{}{k} \leftarrow \statepost{}{}{k} +
\covpost{}{}{k}C(k)^T(C(k)\covpost{}{}{k}C(k)^T +R(k))^{-1}(y-
h(\hat{x}(k)))$
\State $\covpost{}{}{k} \leftarrow \big(I -
\covpost{}{}{k}C(k)^T(C(k)\covpost{}{}{k}C(k)^T + R(k))^{-1}C(k) \big
)P(k)$ \Return $\statepost{}{}{k},\covpost{}{}{k}$
  \end{algorithmic}
\end{algorithm}


\subsection{Event-triggered estimation}\label{se:ETE}
\begin{figure}[htb]
  \centering
  \includegraphics[width=.5\linewidth]{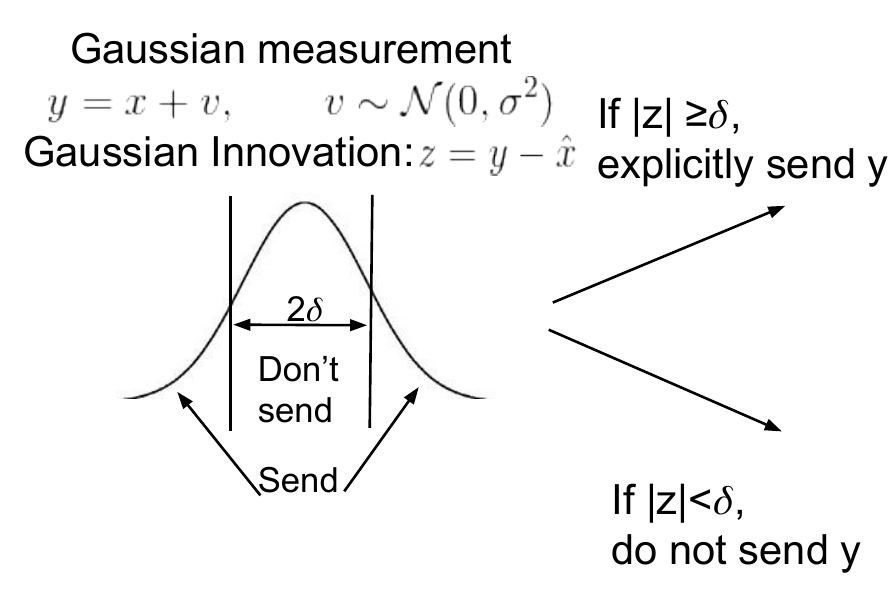}
  \caption{Schematic of an robot using an innovation-based
      triggering strategy.}\label{fig:EvI-send}
\end{figure}

This section describes event-triggered estimation using innovation
thresholds. Figure~\ref{fig:EvI-send} illustrates the basic
event-triggering concept for a measurement taker $i$, which must
decide whether to send its measurement $y$ to neighbor $j$ at a given
time step $k$. In this simple example, suppose $y$ is a direct
measurement of a scalar state $x$, i.e.~$h(x) = x$, 
and for now assume both $i$ and $j$ possess an identical copy
the same local estimate $\hat{x}$ of $x$ just before $i$ measures $y$.
Robot $i$ can compute the measurement innovation
$z=y-\hat{x}$. 
According to Algorithm~\ref{algo:KFmeas}, for a given Kalman gain, a
smaller innovation $z$ will tend to produce a smaller effect on the
update for $\hat{x}$. That is, the closer $y$ is to its expected value
($\hat{x}$ in this case), the smaller the `surprise factor' for the
measurement update step becomes, meaning that $\hat{x}$ will not
change by much. More generally, if the Kalman gain is such that a wide
range of $y$ and $\hat{x}$ values will typically (in a statistical sense) lead to small
$\hat{x}$ updates on the RHS of line~5 
in Algorithm~\ref{algo:KFmeas}, then large shifts in $\hat{x}$ can only
occur when $z$ is relatively large, i.e.~when $y$ disagrees
significantly with $\hat{x}$. If $\hat{x}$ provides a good estimate of
$x$ in the MSE sense, then such major disagreements will be
statistically infrequent, and $z$ tends to remain relatively
small. 

This insight motivates the idea behind event-triggered filtering:
robot $i$ only sends measurement $y$ to $j$ if the magnitude of the
`surprise factor' from the innovation $z$ is larger than some
parameter threshold $\delta$. Thus, measurements come in at a regular
interval to robots, and robots have the opportunity to share or censor
each of these measurements over time. Importantly, since the robots
are aware of $\delta$ and share a common pre-fusion $\hat{x}$, the
lack of a measurement from $i$ can be used by $j$ to infer the
possible values of $y$ obtained at $i$. As such, $j$ can still carry
out a local fusion update for $\hat{x}$ using \emph{implicit
  information} about $x$ in the form of \emph{set-valued knowledge}
for $y$.

\begin{figure}[htb]
  \centering
  \includegraphics[width=.98\linewidth]{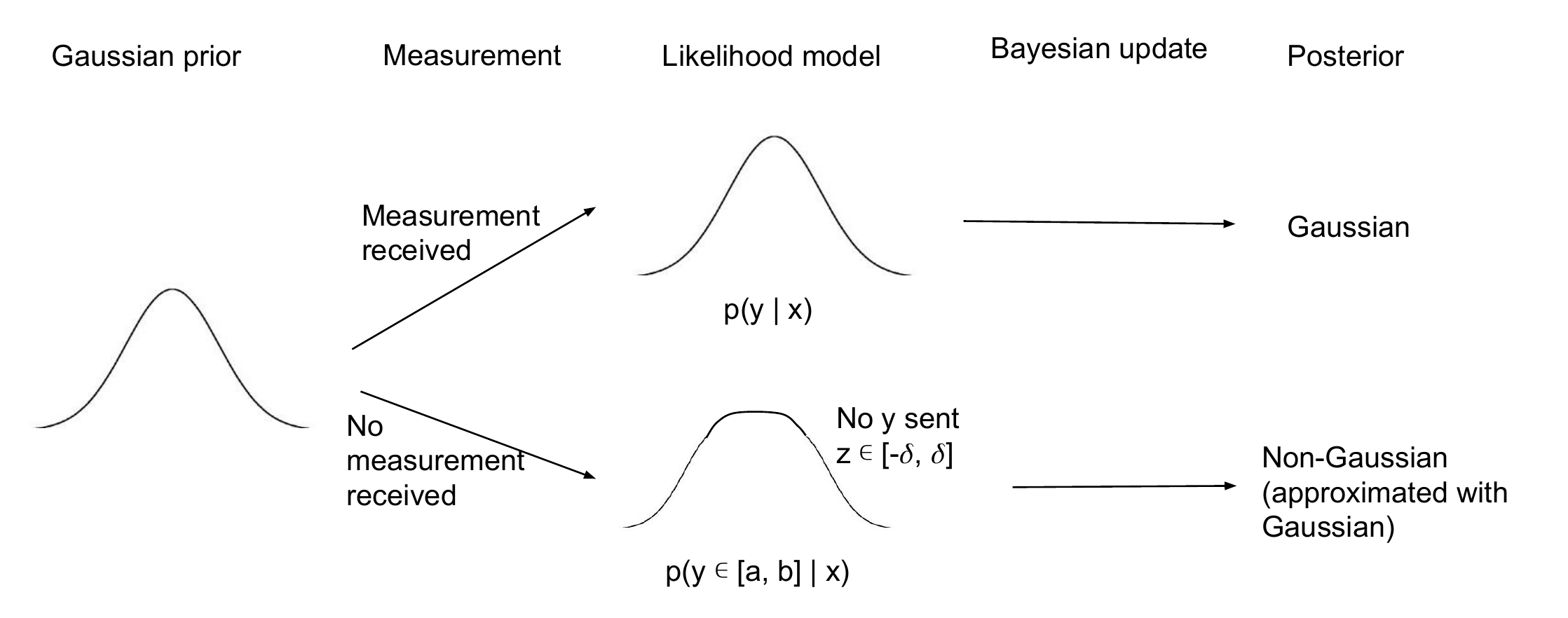}
  \caption{ Schematic of an robot's fusion strategy for
      implicit and explicit measurements received. }
  \label{fig:EvI-receive}
\end{figure}

Figure~\ref{fig:EvI-receive} illustrates how receiving robot $j$
handles a measurement communicated by robot $i$ at a given time step.
If robot $i$ does send the measurement explicitly, it can be fused in
the usual way via the EKF described in Section~\ref{se:CL}. However, if no
explicit measurement is received from robot $i$, robot $j$ reasons
over the set of measurements that would cause $i$ to censor the
measurement (i.e.~such that the innovation is less than $\delta$).
This corresponds to a non-Gaussian measurement model, and assumes that
robot $i$ and $j$ have a common $\hat{x}$ to compute the innovation
from.  

Now, in a decentralized setting with $N\geq 2$ robots, any pair of communicating robots $i$ and $j$ may also communicate 
with other robots that are not in $\neigh{j}$ or $\neigh{i}$, respectively, and which may
therefore contribute additional information that is never received by the other robot in the pair. 
Therefore, the assumption of a common reference estimate $\hat{x}$ being available to all robots
becomes very difficult to enforce (without resorting to techniques like consensus, which as mentioned earlier leads to very high
communication overhead and inefficient estimates). 
Instead, the common $\hat{x}$ assumption is relaxed and adapted to an event-triggering strategy where a
common reference estimate must exist only between \emph{pairs} of communicating robots $i$ and $j$. 
As such, each communicating robot pair must maintain a separate state
estimate that tracks only the (explicit and implicit) information that is actually communicated between them. 
This `extra bookkeeping estimator' strategy shares similarities with other strongly coupled CL methods described earlier, and 
also bears resemblance to the channel filter algorithm used in Bayesian decentralized data fusion, which tracks 
and removes common information exchanged by pairs intelligent sensor nodes to avoid data incest \cite{SG-HRDW:94, Campbell-CSM-2016}. 
 
Section~\ref{se:ETCL} describes the full details of this adapted
event-triggering strategy for decentralized CL. 
In short, each robot $i \in \until{N}$, propagates a current best state and
covariance estimate of the whole network at each timestep $k$,
$(\stated{i}{i}{k},\covd{i}{i}{k}) \in \real^{Nd} \times
\real^{Nd\times Nd}$, and also maintains `common estimates' with each
other robot $j$ that it shares measurements with, denoted by
$(\stated{i}{j}{k},\covd{i}{j}{k})\in \real^{Nd} \times
\real^{Nd\times Nd}$.  These common estimates can be interpreted as
$i$'s estimate of $j$'s estimate of the state of the network.  In
general, this is a conservative estimate because it assumes that $j$
does not receive any other information from the world, i.e.~it only
accounts for information exchanged between $i$ and $j$ exclusively.
The update laws enforce $\stated{i}{j}{k} = \stated{j}{i}{k}$ and
$\covd{i}{j}{k} = \covd{j}{i}{k}$, for all communicating pairs $i, j$
and all times $k \geq 0$, allowing $i$ and $j$ to have a common state
and covariance estimate to compare measurement innovation with the
innovation threshold parameter $\delta$.

Note that this implicit measurement (later formalized
in~\eqref{eq:imp-transform}) has a non-Gaussian measurement likelihood
model, making exact Bayesian fusion of such information analytically
intractable since it does not admit a closed-form solution.  Thus,
following the approach in~\cite{DS-TC-LS:14} to fuse set-valued
measurements, a Gaussian distribution is used to approximate the
posterior pdf on the state (which in this case will be a truncated
Gaussian).  After an implicit measurement is taken, the first two
moments of the prior Gaussian are corrected by weighting its moments
with those of a Gaussian with truncated support
(Lemma~\eqref{le:truncGauss}).  The correction is implemented using a
modified Kalman gain in the Kalman filter state estimator update for
robot $j$.  Algorithm~\ref{algo:imp-meas} describes concisely how this
is done.

Since the Gaussian approximation to the implicit measurement update
only has a closed-form solution for implicit measurements, the
measurement covariance $R$ is assumed to be diagonal (as in
Section~\ref{se:sensing-comms}) to allow for each measurement to be
treated with independent scalar updates.  The use of scalar sequential
updates is necessitated by the fact that the formulas for determining
the moments of a truncated Gaussian pdf are available only in the
scalar case (hence, elements of the measurement vector are censored
and/or fused one element at a time). 
In principle, implicit measurement fusion updates can be numerically approximated for
non-scalar measurements, e.g. using sparse grid quadrature integration methods 
for the necessary moment calculations  \cite{Heiss-JEcon-2007, Jia-JGCD-2011}; 
however, for simplicity, these updates are not developed here. 

\section{Event-triggered Cooperative Localization}\label{se:ETCL}
This section describes the proposed event-triggered cooperative localization algorithm, beginning with an overall description of the algorithm.  
\subsection{Description of event-triggered cooperative localization algorithm}
\begin{figure}[htb]
\centering
\includegraphics[width=.65\linewidth]{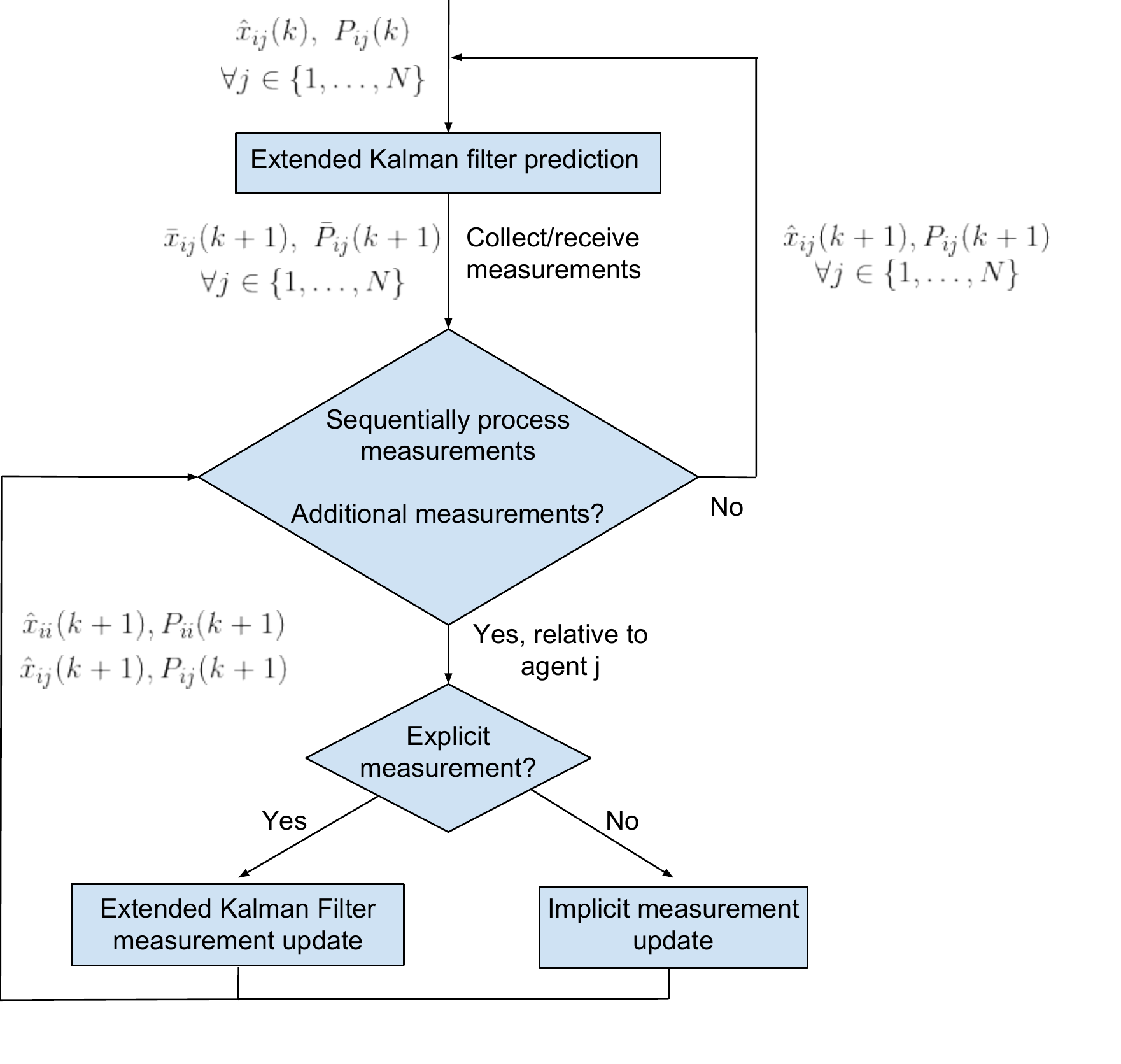}
\vspace{0ex}
\caption{Description of the overall event-triggered cooperative localization flow.}
\label{fig:overall-ETCL}
\end{figure}

%
For a robot to reduce the communication cost of transmitting its
measurements at every time step, it is desirable to employ an
event-triggered strategy to be used within a Kalman filtering
framework.  The idea is that a robot $i$ only broadcasts the $\ell$-th
component of its current measurement vector $\elem{\meas{i}{k}}{\ell}$
at time step $k$ to a specific neighbor $j$ if the norm of that
measurement component's innovation $|\elem{\meas{i}{k} -
  h(\stated{i}{j}{k}}){\ell}|$ (with respect to their
common pair-wise filter estimate $\stated{i}{j}{k}$) 
is larger than a known threshold parameter $\delta$. The absence of a measurement
component from a robot gives implicit information about that
measurement.  This thresholding based on the innovation is applied to
all measurements collected by each agent, i.e. both absolute
measurements (such as GPS) as well as relative measurements (such as
range/bearing).  Each robot's state estimation filter must therefore
be adapted so that implicit information can be included.
Figure~\ref{fig:overall-ETCL} describes the overall flow of the
proposed event-triggered cooperative localization algorithm.

The algorithm described here is a new decentralized adaptation of the
centralized event-triggered filter in~\cite{DS-TC-LS:14}. This
new distributed algorithm provides a tunable method to tradeoff between the
quality of state estimation and costly communication of measurements.
At every time step, each agent $i$ updates its own estimate of the
current state of the network as well as its estimate of each other
agent $j$'s estimate of the network (where $j \neq i$).  Agent $i$
receives measurements from its sensor suite and performs a Kalman
filter update to its own posterior state estimate.  Using its a priori
common state with each other agent $j$, i.e.~$\stated{i}{j}{k}$, it
determines which, if any, components of its measurement to send to
each neighbor $j$ based on an innovation triggering threshold. Agent
$i$ receives explicit and implicit measurements from other
agents. Implicit measurements are defined to be the knowledge gained
when a measurement is not sent (i.e.~when the innovation is smaller
than the threshold). Explicit measurements are fused using a Gaussian
measurement model via a Kalman filter update, while implicit
measurement updates are performed using Gaussian moment matching
approximations for set-based measurement updates. All measurements are
used to update $\stated{i}{i}{k}, \covd{i}{i}{k}$, whereas only
measurements sent between $i$ and $j$ (explicitly or implicitly) are
used to update $\stated{i}{j}{k}, \covd{i}{j}{k}$.

\subsection{Event-triggered information fusion}
The components of the algorithm are now described in detail.  The
algorithm is initialized with $\statepost{i}{j}{0} =
\statepost{j}{i}{0}$ and $\covpost{i}{j}{0}=\covpost{j}{i}{0}$, for
all $i,j \in \until{N}$. At timestep $k$, agent $i$ first propagates
the state estimates for each $j \in \until{N}$: $ \statepre{i}{j}{k} =
f(\statepost{i}{j}{k-1}, u(k))$, $\statepost{i}{j}{k}
=\statepre{i}{j}{k}$ and $ \covpre{i}{j}{k} =
A(k)\covpost{i}{j}{k-1}A(k)^T + Q(k)$, $\covpost{i}{j}{k} =
\covpre{i}{j}{k}$.  Note that $A(k) = \diag(A_i(k)) $ and $Q(k) =
\diag(Q_i(k))$.

At each timestep $k$, robot $i$ obtains a measurement $\meas{i}{k} \in
\real^{m_i}$ from its sensor suite. Given a $\delta \in
\realpositive$, for each $j \in \neigh{i}$, robot $i$ determines for
which measurement vector component $\ell \in \until{m_i}$ the expected
innovation will be large enough, i.e., $|\elem{\meas{i}{k} -
  h(\statepre{i}{j}{k}}){\ell}| > \delta$. 
	Then, it sends the measurement components for which this holds to agent $j$. In turn, robot $i$ will
receive explicit measurements from other robots $l$ for which a
similar condition holds.

\subsubsection{Explicit measurement fusion}For each $\ell \in
\until{m_i}$ for which the corresponding innovation is larger than
$\delta$, agent $i$ sends measurement component $\ell$ to $j$.  Robot
$j$ performs a Kalman update (c.f.~Algorithm~\ref{algo:KFmeas})
with the measurement component
$\elem{\meas{i}{k}}{\ell}$ and corresponding measurement model
	to its own estimate
$(\statepost{j}{j}{k},\covpost{j}{j}{k})$ as well as to the common
state $(\statepost{j}{i}{k},\covpost{j}{i}{k})$.  Agent $i$ makes the
same Kalman update to $(\statepost{i}{j}{k},\covpost{i}{j}{k})$,
keeping the common state estimate for $i$ and $j$ identical.

\subsubsection{Implicit measurement fusion}
For each measurement component $\ell \in \until{m_i}$ for which the
corresponding predicted innovation is smaller than $\delta$, agent $i$
does not send that measurement component to $j$.  Therefore, $j$
`receives' the implicit information from $i$ that $|\elem{\meas{i}{k}
  - h(\statepre{i}{j}{k})}{\ell}| \leq \delta$,
or
\begin{align}\label{eq:imp-transform}
  &\elem{\meas{i}{k} - h(\statepre{j}{j}{k})}{\ell} \in \nonumber \\
  &[-\delta + (\row{h(\statepre{j}{i}{k})}{\ell} - \row{h(\statepre{j}{j}{k})}{\ell}),
    \delta + (\row{h(\statepre{j}{i}{k})}{\ell} - \row{h(\statepre{j}{j}{k})}{\ell})].
\end{align}
Agent $j$ fuses the implicit measurement of component $\ell$ into its
estimate, $(\statepost{j}{j}{k},\covpost{j}{j}{k})$, and both $i$ and
$j$ also fuse it into their shared estimate,
$(\statepost{j}{i}{k},\covpost{j}{i}{k})$ and
$(\statepost{i}{j}{k},\covpost{i}{j}{k})$, using
Algorithm~\ref{algo:imp-meas}.    
Specifically, $x_{\text{ref}}$ refers to the specific common shared state
which is kept by an
agent; that is, if the measurement was implicitly sent from $i$ to
$j$, then $x_{\text{ref}}= \statepre{i}{j}{k}$. Other variables refer
to the states, covariance, measurements, and error
statistics of the problem.  

\begin{algorithm}[htb]
  \caption{Implicit Measurement Update for Kalman Filter}
  \label{algo:imp-meas}
  \begin{algorithmic}[1]
    \Require $\statepre{}{}{k}, \covpre{}{}{k}, \statepost{}{}{k}, \covpost{}{}{k}, x_{\text{ref}},  C(k), R(k), \delta$ 
    \State $\mu \leftarrow h(\statepost{}{}{k}) - h(\statepre{}{}{k})$
    \State $Q_e \leftarrow C(k)\covpre{}{}{k}C(k)^T + R(k)$
    \State $\alpha \leftarrow h(x_{\text{ref}}) - h(\statepre{}{}{k})$
    \State $\bar{z} \leftarrow \frac{ \phi( \frac{-\delta+\alpha
        -\mu}{\sqrt{Q_e}}) - \phi( \frac{\delta+\alpha
        -\mu}{\sqrt{Q_e}}) }{ \mathbf{Q}( \frac{-\delta+\alpha
        -\mu}{\sqrt{Q_e}}) - \mathbf{Q}( \frac{\delta+\alpha
        -\mu}{\sqrt{Q_e}})}\sqrt{Q_e}$
    \State $\vartheta \leftarrow \Big(\frac{ \phi(
      \frac{-\delta+\alpha -\mu}{\sqrt{Q_e}}) - \phi(
      \frac{\delta+\alpha -\mu}{\sqrt{Q_e}}) }{ \mathbf{Q}(
      \frac{-\delta+\alpha -\mu}{\sqrt{Q_e}}) - \mathbf{Q}(
      \frac{\delta+\alpha -\mu}{\sqrt{Q_e}})} \Big)^2 $ 
%
    $- \frac{ (\frac{-\delta+\alpha -\mu}{\sqrt{Q_e}})\phi(
      \frac{-\delta+\alpha -\mu}{\sqrt{Q_e}}) - (\frac{\delta+\alpha
        -\mu}{\sqrt{Q_e}})\phi( \frac{\delta+\alpha -\mu}{\sqrt{Q_e}})
    } {\mathbf{Q}( \frac{-\delta+\alpha -\mu}{\sqrt{Q_e}}) -
      \mathbf{Q}( \frac{\delta+\alpha -\mu}{\sqrt{Q_e}})} $
    \State $ K \leftarrow
    \covpost{}{}{k}C(k)^T(C(k)\covpost{}{}{k}C(k)^T + R(k))^{-1}$
    \State $\statepost{}{}{k} \leftarrow\statepost{}{}{k} + K\bar{z}$
    \State $\covpost{}{}{k} \leftarrow\covpost{}{}{k} - \vartheta
    \covpost{}{}{k}C(k)^T(C(k)\covpost{}{}{k}C(k)^T +
    R(k))^{-1}\covpost{}{}{k} $ 
   \Return $\statepost{}{}{k},\covpost{}{}{k}$
  \end{algorithmic}
\end{algorithm}

Algorithm~\ref{algo:imp-meas} shows the modified KF update that
results in the optimal posterior state estimate following fusion of an
implicit scalar measurement (which follows from Lemma
\ref{le:truncGauss} and Theorem~$7$ in
\cite{DS-TC-LS:14}). Furthermore, $\vartheta \in (0,1)$, leading to a
strict decrease in the covariance.

\section{Event-triggered Cooperative Localization with Covariance Intersection}
\begin{figure*}[tb]
 \centering \subfigure[Implicit/explicit data flow to robot $i$]{
   \includegraphics[width=.37\linewidth]{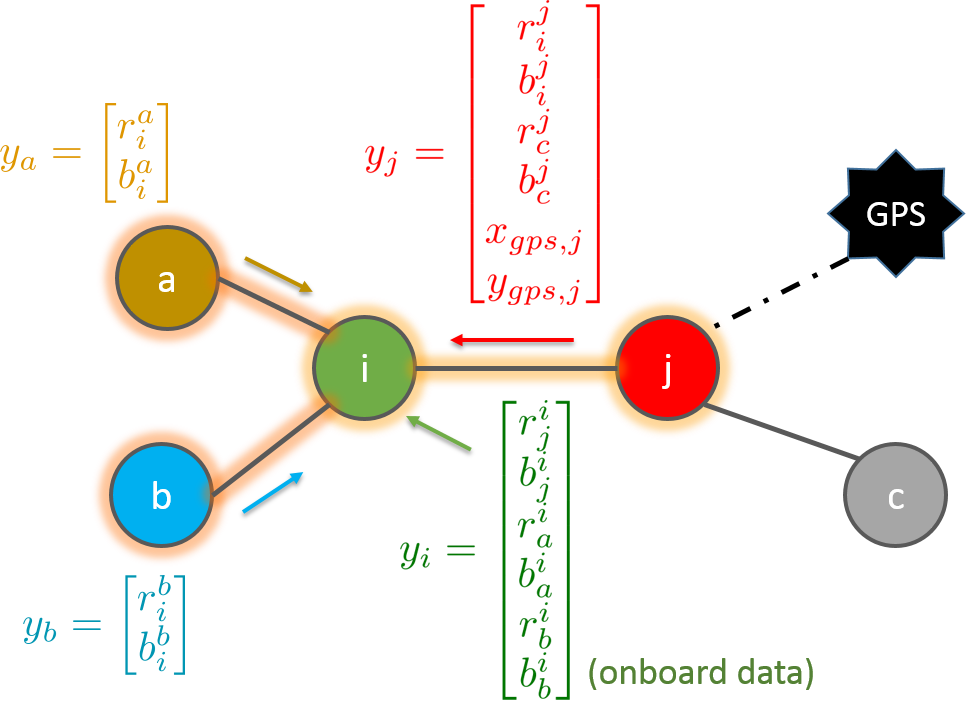}}
\vspace{0ex}
\subfigure[Implicit/explicit data flow to robot $j$]{
  \includegraphics[width=.49\linewidth]{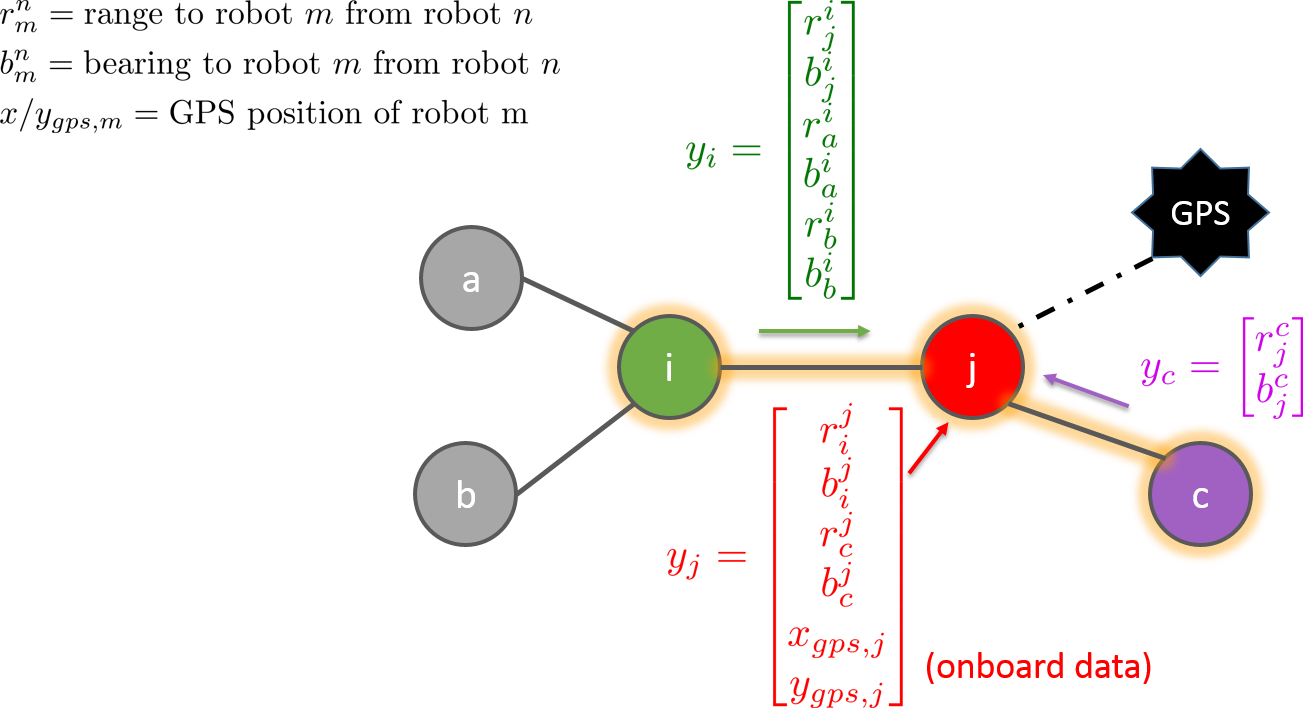}}
\vspace{0ex}
\subfigure[Covariance intersection information flow]{
  \includegraphics[width=.75\linewidth]{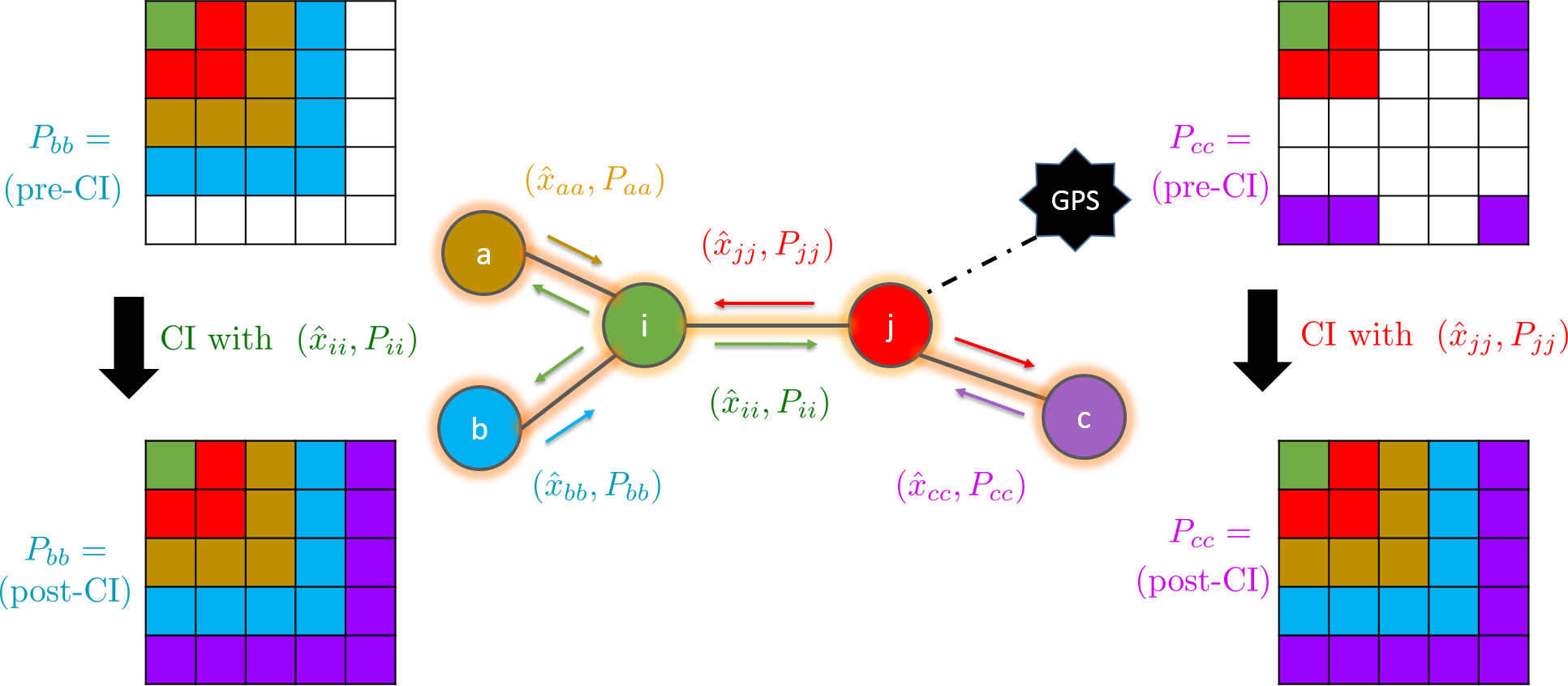}}
  \vspace{0ex}
\caption{Event-triggered CL information flow for $N=5$ network, where only $j$ receives GPS. 
}\label{fig:infoflow}
\end{figure*}
Figure ~\ref{fig:infoflow} (a) and (b) illustrate information flow according to the event-triggered decentralized CL approach, where 
any element of any $y$ vector can be censored to form implicit data via event-triggering. 
As before, the goal is for all of the robots to perform event-triggered CL to estimate the states of all robots in the network, 
each using the subset of explicit measurements received and the implicit information encoded in the lack of a measurement. 
However, it becomes clear that two issues can arise. 
 Firstly, in networks with $N > 2$ robots, the common reference state estimates $x_{ij}$ and $x_{ji}$ will eventually start
 to differ significantly from $x_{ii}$ and $x_{jj}$, since robots $i$ and $j$ can generally receive information from other robots that lie outside of $\neigh{i} \cap \neigh{j}$. 
 Secondly, in practice, each robot will need to accurately estimate the states of only \textit{some} robots in the network, e.g. as only this information may be required to execute certain coordinated tasks, or because other robots need this information relayed to them since they are too far away to directly communicate with/measure certain robots. 
 While the `relevant subset' that any robot cares about can change over time, the network topology clearly plays an important role in determining the observability of robot states: the states of robots that are not directly measured by $i$ 
 or by robots in $\neigh{i}$ are unobservable to $i$ using explicit or implicit measurements only. 
Both issues can be easily remedied if neighboring robots occasionally augment the event-triggered data sharing protocol with direct fusion of their Kalman filter estimates. 
  
 More formally, suppose each robot $i$ aims to ensure
 that $\trace(\covd{i}{i}{k}\diag(\alpha_i))$ remains less than a
 threshold parameter $\tau_{\text{goal}}^i$, for each $i \in
 \until{N}$. For simplicity, assume for now all agents have identical goals
 $\tau_{\text{goal}}^i=\tau_{\text{goal}}, \ \forall  i \in
 \until{N}$, 
 (Sec. \ref{se:heuristic} considers dynamically adjusted $\tau_{\text{goal}}^i$). 
 The preference vector
 $\alpha_i \in \realnonnegative^{Nd}$, $i\in \until{N}$, defines how
 sensitive robot $i$ is to uncertainty in a given robot's position
 components. For instance, one reasonable choice is that a robot would give more
 weight to the states of nearby/communicating robots than ones farther away, 
 since having more accurate knowledge of these states improves self-localization. 

 In the proposed event-triggering algorithm, 
 robot $i$ always fuses its own measurements and, depending
 on the state of the network, either implicit or explicit measurements
 from its neighbors.  
 If none of these measurements pertain to a
 generic robot $r$, robot $i$'s error covariance for the location of
 $r$ will grow unbounded due to process noise.  If $\alpha_i$ gives
 zero weight to the error in robot $r$'s location ($\alpha_i^r=0)$,
 then $i$ is not sensitive to unbounded estimation error of
 $r$ when minimizing $\trace(\covd{i}{i}{k}\diag(\alpha_i))$.
 However, if $\alpha_i^r>0$, in order to enforce that weighted
 covariance trace $\trace(\covd{i}{i}{k}\diag(\alpha_i)) <
 \tau_{\text{goal}}$, robot $i$ must keep its estimation error of
 robot $r$ bounded.  

Since it is assumed that $i$ never receives implicit or explicit measurements about $r$, an 
 event-based Covariance Intersection (CI) fusion mechanism is introduced
 as a way for $i$ to improve the estimation of any $j$ that it
 cannot receive measurements about.   
Note that this approach is novel compared to other decentralized CL algorithms, 
because robots communicate and fuse \emph{either} implicit/explicit observation data 
\emph{or} state estimates (means and covariances), rather than always sharing explicit measurements only, 
state estimates only, or both simultaneously.

\subsection{Covariance intersection} \label{sec:CI}
CI is a conservative, consistent method for directly fusing two state estimates 
(with possibly non-disjoint information sets) into a new state estimate that takes into account
the information in both estimates \cite{SJ-JU:07}. 
CI prevents double-counting of measurements in an ad-hoc, arbitrary communication
topology. Continuing the previous example, robot $i$ could improve its
estimation of the unmeasured robot $r$ by performing a CI with another 
robot $j$ that has a better estimate of $r$'s location. 
Fig. ~\ref{fig:infoflow} (c) shows how this information propagation 
leads to `filling in' of the corresponding blocks of the covariance matrix for
robots across the network, i.e. robots $a$ and $b$ receive information about 
robot $c$ (and its correlations to other robots) through CI with $i$, 
while $c$ receives information about $a$ and $b$ (and their correlations) through
CI with $j$. Note that even though $i$ cannot measure $c$ (and therefore 
cannot pass measurements pertaining to $c$ directly onto $a$ and $b$),
$i$ receives measurements information about $c$ from $j$ and also occasionally
performs CI with $j$, which also in turn periodically performs CI with $c$. 
Likewise, even though $j$ cannot measure $a$ or $b$, information about the latter pair
of robots eventually `percolates' back to $c$ via CI. 
Hence, as long as all robots are part of a connected network graph, dependency information for all robot
states in the network eventually reaches all robots via occasional neighbor-to-neighbor CI updates. 
However, in the interim, the event-triggering protocol is used to fuse measurements between neighbors. 
Note that all elements of estimates shown in (c) are explicitly shared during CI.  

Note that, since CI is a `single shot' decentralized fusion process, new information takes time to spread throughout the network. 
For instance, if robot $i$ receives new measurement information about $c$ via implicit/explicit measurement exchange
with $j$ at time $k$, then the information from time $k$ will reach $a$ and $b$ via CI at time $k'\geq k$ as part of 
$i$'s Kalman filter estimate at time $k'$. 

CI requires a parameter $\omega \in \realnonnegative$ that defines the 
relative weighting of the two estimates based on the relative quality of each
estimate. Algorithm~\ref{algo:cov-int} describes how one can find the
parameter $\omega^*$ that minimizes the weighted trace of the
fused covariance. This minimization can be solved efficiently with
gradient-free 1D optimization, e.g. bisection or 
golden section search \cite{Vanderplaats-Book-1984}.

\begin{algorithm}[htb]
  \caption{Covariance Intersection}
  \label{algo:cov-int}
  \begin{algorithmic}[1]
		\Require $\statepost{i}{i}{k}, \covpost{i}{i}{k}, \statepost{j}{j}{k}, \covpost{j}{j}{k},\alpha$
     \State $\omega^* \!\leftarrow\! \argmin_{\omega
      \in [0,1]} \trace((\omega \covpost{i}{i}{k} ^{-1} + (1-
    \omega)\covpost{j}{j}{k}^{-1})^{-1}\!\diag(\alpha))$
     \State $P_{\text{new}} \leftarrow (\omega^* \covpost{i}{i}{k}^{-1}
    + (1- \omega^*)\covpost{j}{j}{k}^{-1})^{-1}$ 
    \State $\mu_{\text{new}} \leftarrow P_{\text{new}}(\omega^* \covpost{i}{i}{k}^{-1}{\statepost{i}{i}{k}}
    + (1-\omega^*)\covpost{j}{j}{k}^{-1}{\statepost{j}{j}{k}})$ 
    \Return $\statepost{i}{i}{k} = \mu_{\text{new}}, \covpost{i}{i}{k} = P_{\text{new}}$
		
  \end{algorithmic}
\end{algorithm}

 It should be noted that CI has several drawbacks that make
 it inapplicable as the primary state estimate updating method 
 for strongly coupled CL.  
 First, it is more costly to send state estimates (mean and
 covariances) over a communication channel than it is to send a
 combination of explicit and implicit measurements to neighboring
 robots.  Second, the computation cost is greater than performing a
 Kalman filter update due to the additional optimization step required
 for finding the weighting parameter $\omega$. 
 Third, as alluded to above, CI is a conservative fusion process. 
 Thus, it is guaranteed to discard some new information to avoid data incest
 and produces suboptimal fusion results (whereas the proposed
 event-triggered scheme can closely approximate the idealized
 centralized fusion rule that is given the same information set reaching a particular robot). 
 Nonetheless, CI is guaranteed to be a consistent way to sporadically `reset' 
 common reference state estimates for ad hoc information dependencies  
 and when it is infeasible to directly send measurements throughout the network. 


\subsection{ Heuristic thresholding dynamics} \label{se:heuristic}
When the network diameter is large, a single $\tau_{\text{goal}}$ may not ensure that all robots' covariances actually remain below this threshold via combined event-triggered measurement fusion and CI. 
That is, there is no guarantee that triggering CI fusion updates for any robot 
whenever the weighted trace of the covariance exceeds $\tau_{\text{goal}}$ will result in
the posterior weighted covariance trace becoming less than
$\tau_{\text{goal}}$ for \emph{all} robots. 

To remedy this, heuristic balancing dynamics can be optionally introduced on the robots' triggering
thresholds. In this case, robot $i$ performs CI with its neighbors only
 when $\trace(\covd{i}{i}{k}\diag(\alpha_i)) > \tau^i_{\text{goal}} $,
 where $\tau_{\text{goal}}^i$ is now a dynamically varying quantity
 that helps ensure each robot actually achieves its desired uncertainty goal. 
Let $r_j(k)$ be the fraction of timesteps that
robot $j$ has triggered a Covariance Intersection up to timestep $k$.
For $\epsilon_1, \epsilon_2 \in \realnonnegative$, the dynamics for
agent $i$ are defined as:
\begin{align}\label{eq:CI-trigger}
  \tau_i(k+1)\! \!=\! \min\big(\tau_{\text{goal}},\tau_i(k) \!+\! \epsilon_1 \!\sum_{j \in
    \neigh{i}}\!\!(r_i(k)-r_j(k)) + \epsilon_2(\tau_{\text{goal}} \!-\!
  \tau_i)\big)\!.
\end{align}
The first argument of the minimization
ensures that the triggering threshold is never above the
goal~$\tau_{\text{goal}}$. 
The second argument contains three terms,  
the first of which enforces changes based on the previous value. 
The second term (agreement dynamics) cause robots 
with large triggering rates relative to their neighbors to increase
their threshold. Robots with triggering rates smaller than
their neighbors decrease their threshold (which increases their
triggering rate as they perform CI more often). 
Together, all of the terms in eq. \ref{eq:CI-trigger} (i.e., with $\epsilon_2=0$) 
cause the triggering rates to converge such that all robots'
weighted covariance traces are below~$\tau_{\text{goal}}$. The final
term penalizes deviations from~$\tau_{\text{goal}}$.  As seen in Example 2 of
Section~\ref{se:sims}, these dynamics are a heuristic that, when
properly tuned, allow the weighted
covariance trace to converge closer to $\tau_{\text{goal}}$ for those robots that are the least well-connected in the network.  
It is possible to formally tune these parameters offline using truth model simulations. 
Note that, due to the parameter $\alpha_i$, there can be a large variability in the frequency of robots' updates throughout the network. 
As seen later in Figure~\ref{fig:CIboth}, the overall result is that well-connected robots in the network increase their triggering rate to aid the less-connected ones. 
Hence, these balancing dynamics are one possible way to allow the information of interest to propagate more efficiently between robots. 

\section{Linear Model Simulation Results}\label{se:sims}
Simulations results were obtained to analyze the performance of the decentralized event-triggered CL algorithm for both simple one-dimensional, linear dynamics and nonlinear Dubins vehicle dynamics with nonlinear range and bearing measurements.  This section focuses the linear case to provide illustrative proof of concept examples, while Section~\ref{se:sims-nl} describes the nonlinear case to examine algorithm performance under more realistic application conditions. 

For simplicity, the following simulations consider several robotic
agents moving in one dimension with identical known motion control
law and dynamics:
$x(k+1) = x(k) + u(k) + v(k).$
Here, the additive process noise $v$ has zero mean and diagonal
covariance matrix has diagonal entries equal to $0.1~m^2$ .The agents
interact according to a graph line topology; i.e., agent $1$ can
communicate with $2$, agent $2$ can communicate with $1$ and $3$, etc.
Each agent can take GPS measurements of its own position with no bias and covariance $10~m^2$,
as well as relative distance to the agents that it can communicate with, with no bias and covariance
$1~m^2$.  The event-triggering parameter $\delta=0.75$. In the
following, we include two simulated experiments.  In the first, the
event-triggering strategy is shown to keep the trace of the covariance
bounded for three robots without using the CI reset mechanism. 
In the second, seven robots employ both the
event-triggered measurement propagation as well as CI resets.

\subsection*{Example 1: Three robots, constant thresholds}

\begin{figure*}[htb]
 \centering \subfigure[Robot $2$'s estimation errors]{
   \includegraphics[width=.4\linewidth]{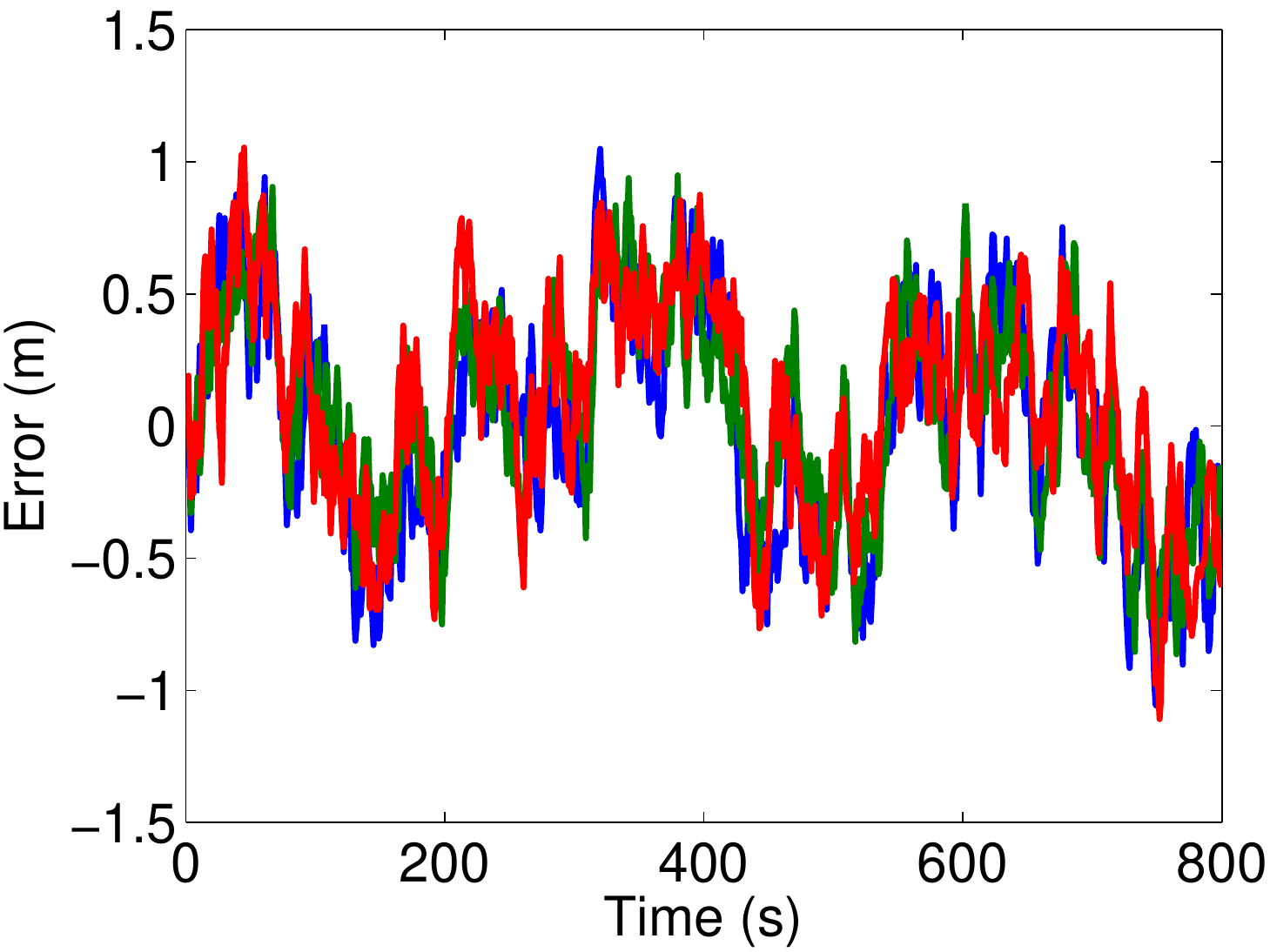}}
\vspace{0ex}
\subfigure[Trace of network state covariance]{
  \includegraphics[width=.4\linewidth]{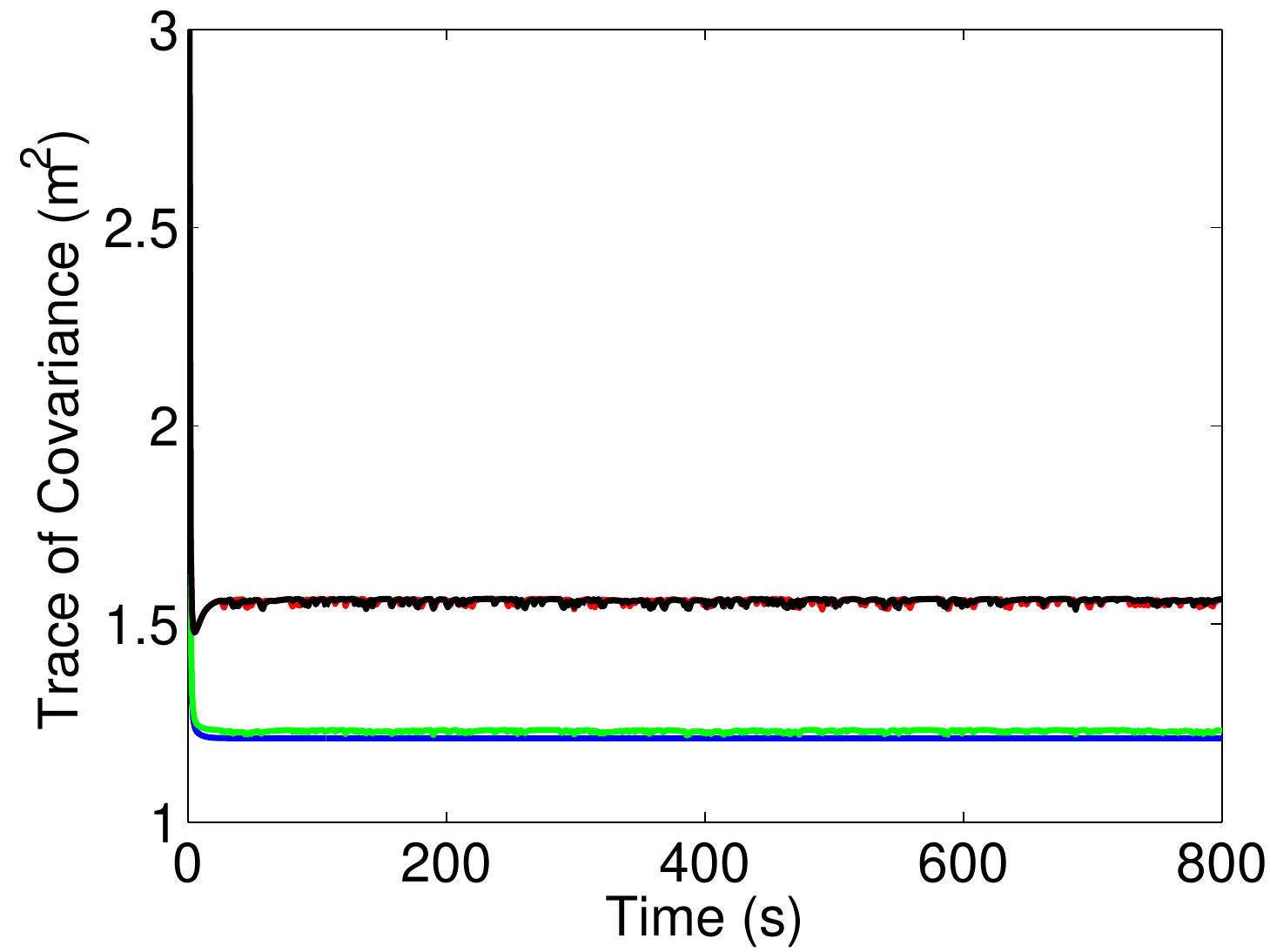}}
\vspace{0ex}
\subfigure[Effect of no implicit information]{
  \includegraphics[width=.4\linewidth]{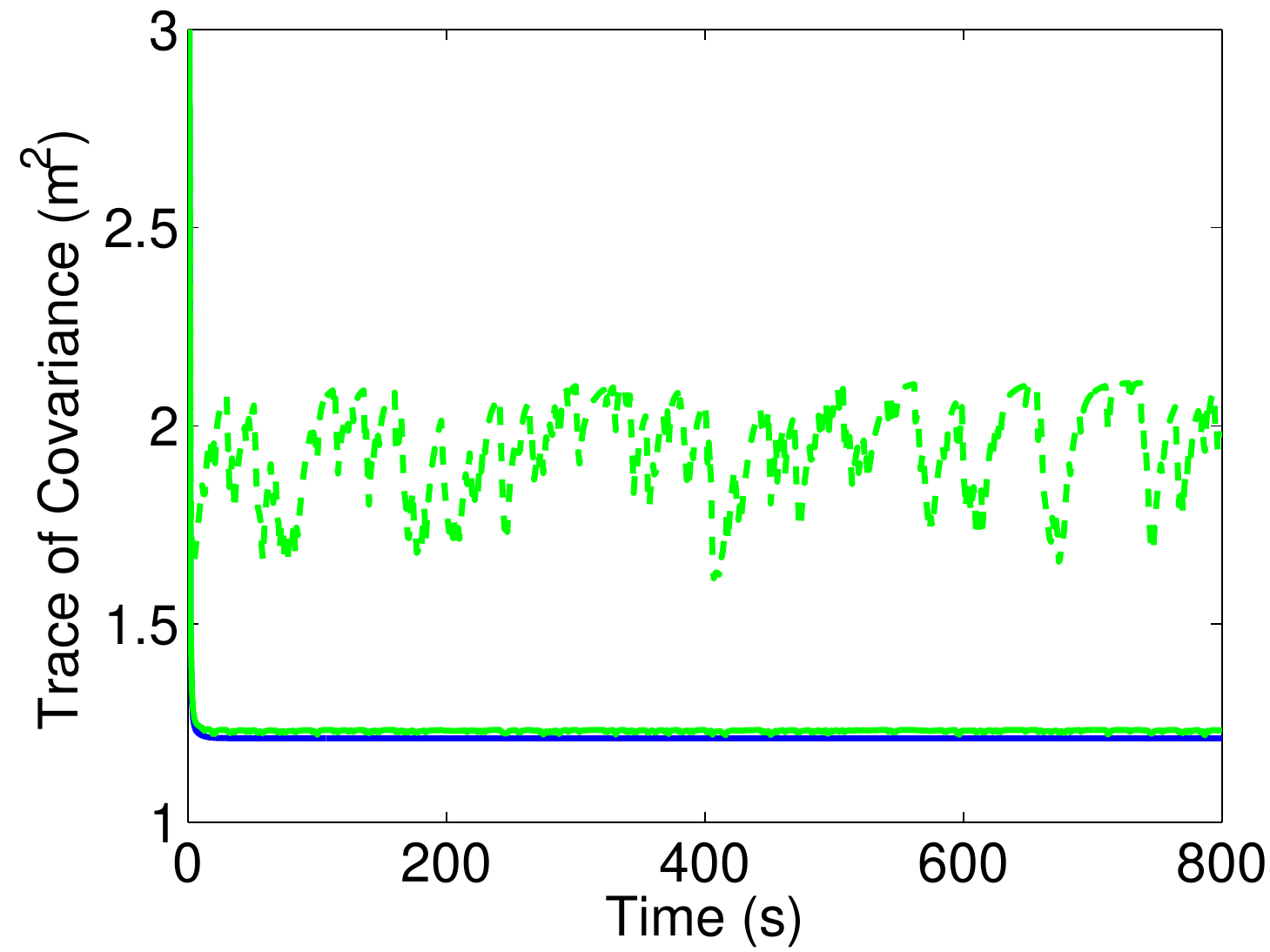}}
  \vspace{0ex}
 \subfigure[Fraction of messages
   sent]{\includegraphics[width=.4\linewidth]{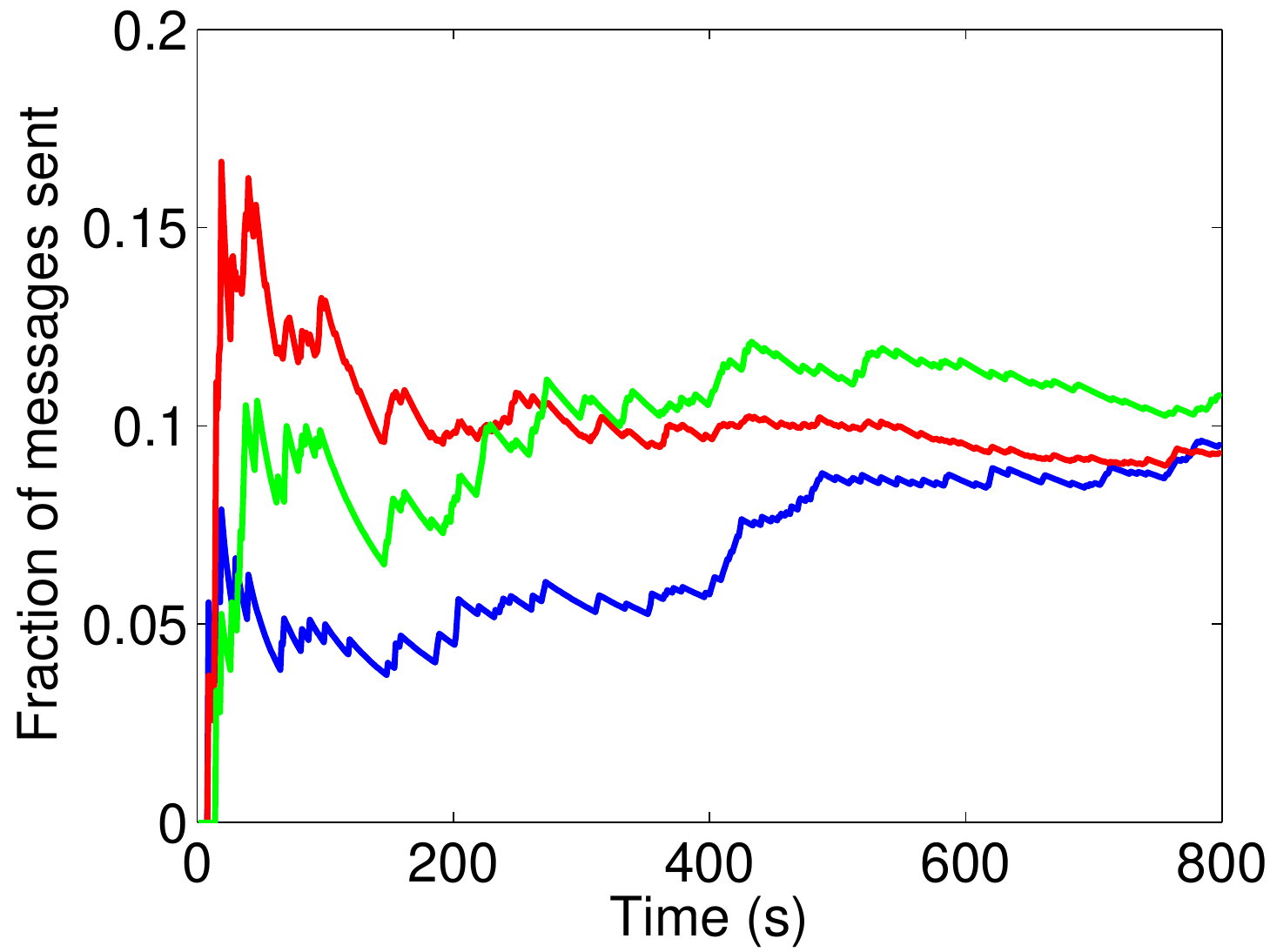}}

\caption{Results for Example 1.
}\label{fig:tracecov3}
\end{figure*}
For $N=3$ agents, Figure~\ref{fig:tracecov3}(a) depicts the the error between the true robot
locations and robot $2$'s filter estimates (red=1, green=2, blue =3). 
Note that it was able to track the true network state with small variance. 
For the same evolution, Figure~\ref{fig:tracecov3}(b) shows the trace
of the covariance matrix for each agents' estimate compared to that
obtained through a centralized Kalman filter that fuses the
information of all agents' measurements (red=1, green=2, black=3, blue= centralized). 
Note that since agent $2$ (green) is
the most connected, its covariance is nearly equal to the centralized
Kalman filter (blue).
Figure~\ref{fig:tracecov3}(c) shows the values of the trace of the
covariance for the case where agent $2$ does not fuse any of the
implicit information received from agents $1$ and $3$ (dashed green), versus the case where
robot $2$ fuses all implicit/explicit data (solid green) and versus the centralized estimate (blue). 
The implicit information clearly makes a large difference towards minimizing the
estimation error. 
Figure~\ref{fig:tracecov3}(d) shows the cumulative fraction of the total
measurements that were explicitly sent to other agents by other each agent (red=1, green=2, blue =3). 
Even though approximately $10\%$ of messages were sent, the overall
network estimation performance is not much worse than the centralized
Kalman filter.  This justifies the use of event-triggered measurements
to trade off between estimation performance and measurement
communication cost.  The threshold parameter $\delta$ can be tuned
according to a given application's relative importance between
performance and communication cost.  This figure only
  counts explicit measurements sent and does not include messages
  related to CI.  However, because all agents
  were able to directly or indirectly measure all other agents, no
  CI updates were required in this execution.

\subsection*{Example 2: Seven robots, with/without adaptive
  thresholding}

Now we consider the network to include $7$ agents under the same
assumptions as defined above (line graph communication topology,
agents can measure their own position and relative position to agents
on either side of it).  Because agents are not able to measure (either
directly or via a communicating neighbor) all other agents, the
position covariance of these robots would diverge in time.  Thus,
agents periodically perform CI with neighbors to
fuse their state estimates and gain information about agents they
cannot measure. 

\begin{figure*}[ht!]
 \centering \subfigure[Network covariance trace, $\tau_{\text{goal}}=5m^2$, no adaptive
 thresholding]{ \includegraphics[width=.4\linewidth]{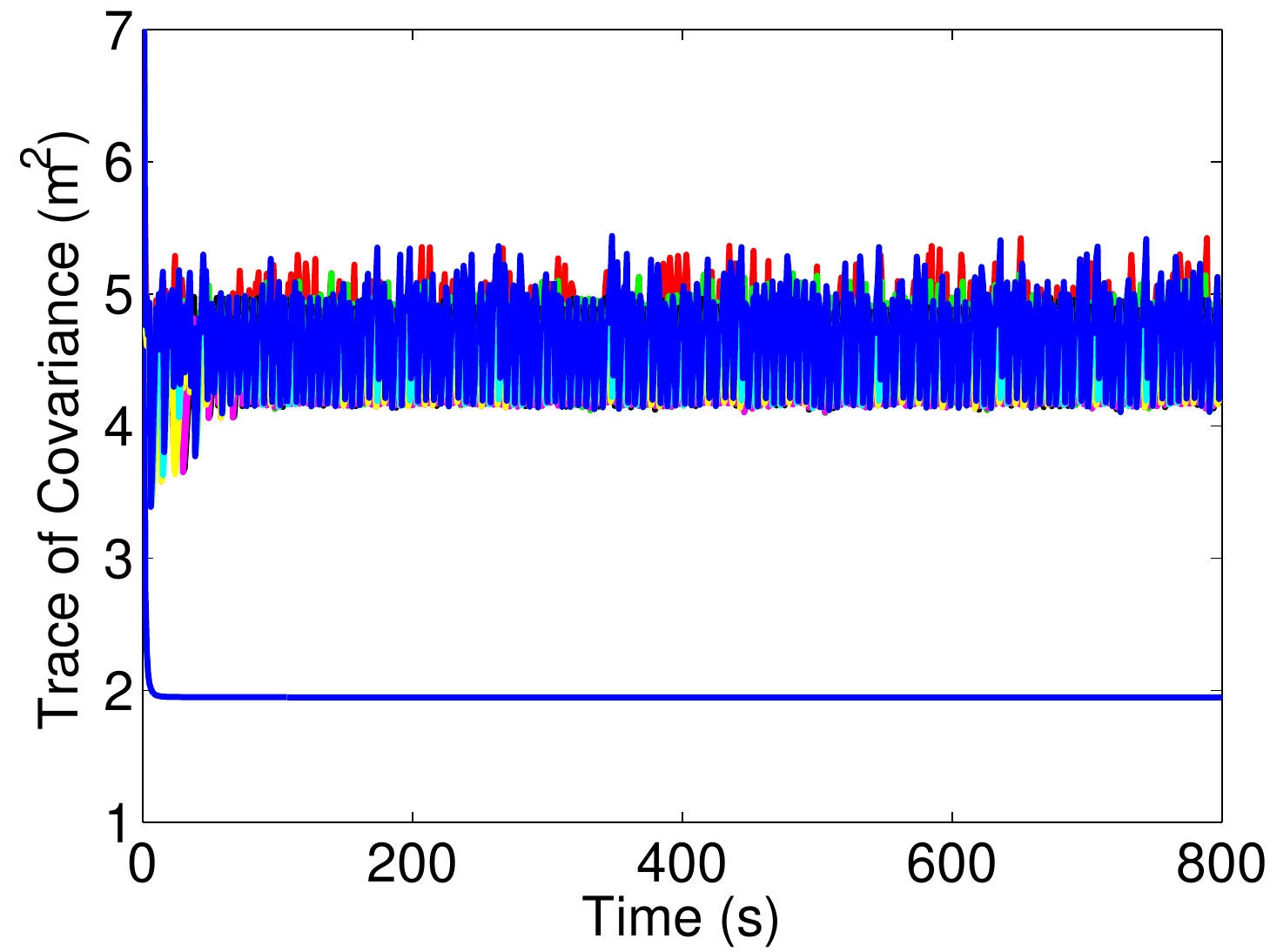}}
 \, \subfigure[ CI trigger rate, no adaptive
 thresholding]{\includegraphics[width=.4\linewidth]{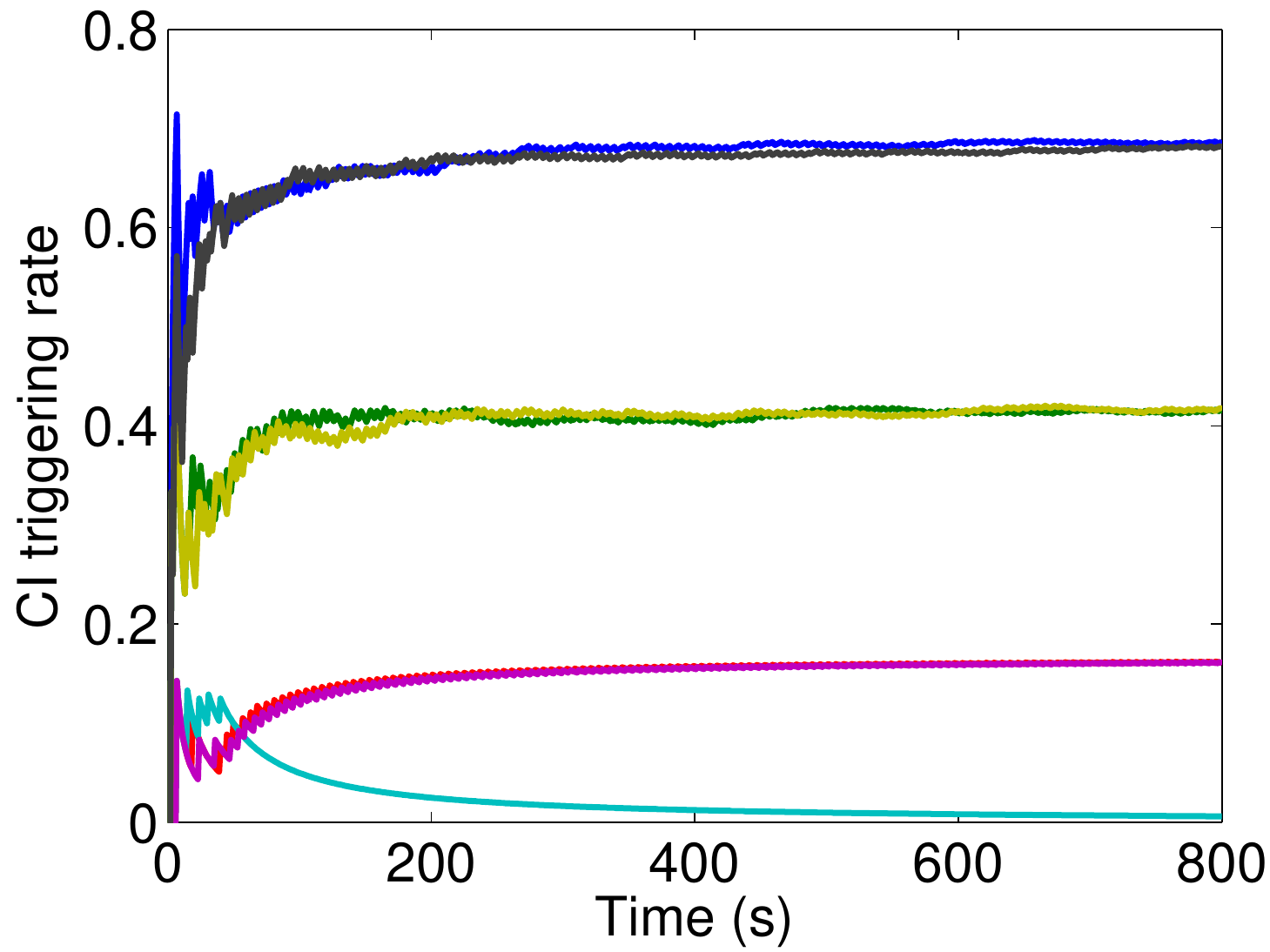}}
 \subfigure[ Network covariance trace, $\tau_{\text{goal}}=5m^2$, adaptive
 thresholding]{ \includegraphics[width=.4\linewidth]{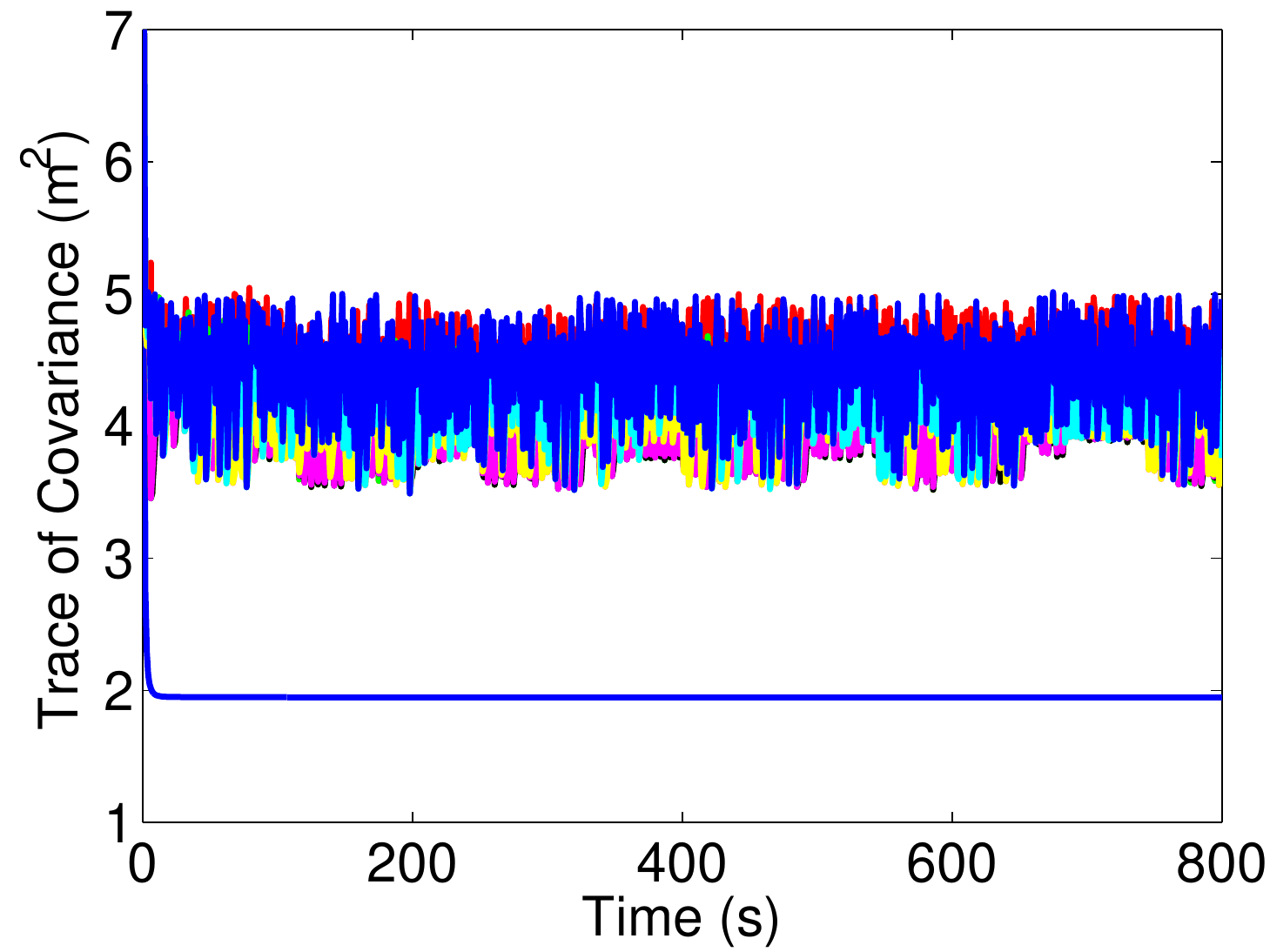}}
 \, \subfigure[ CI trigger rate, adaptive
 thresholding]{\includegraphics[width=.4\linewidth]{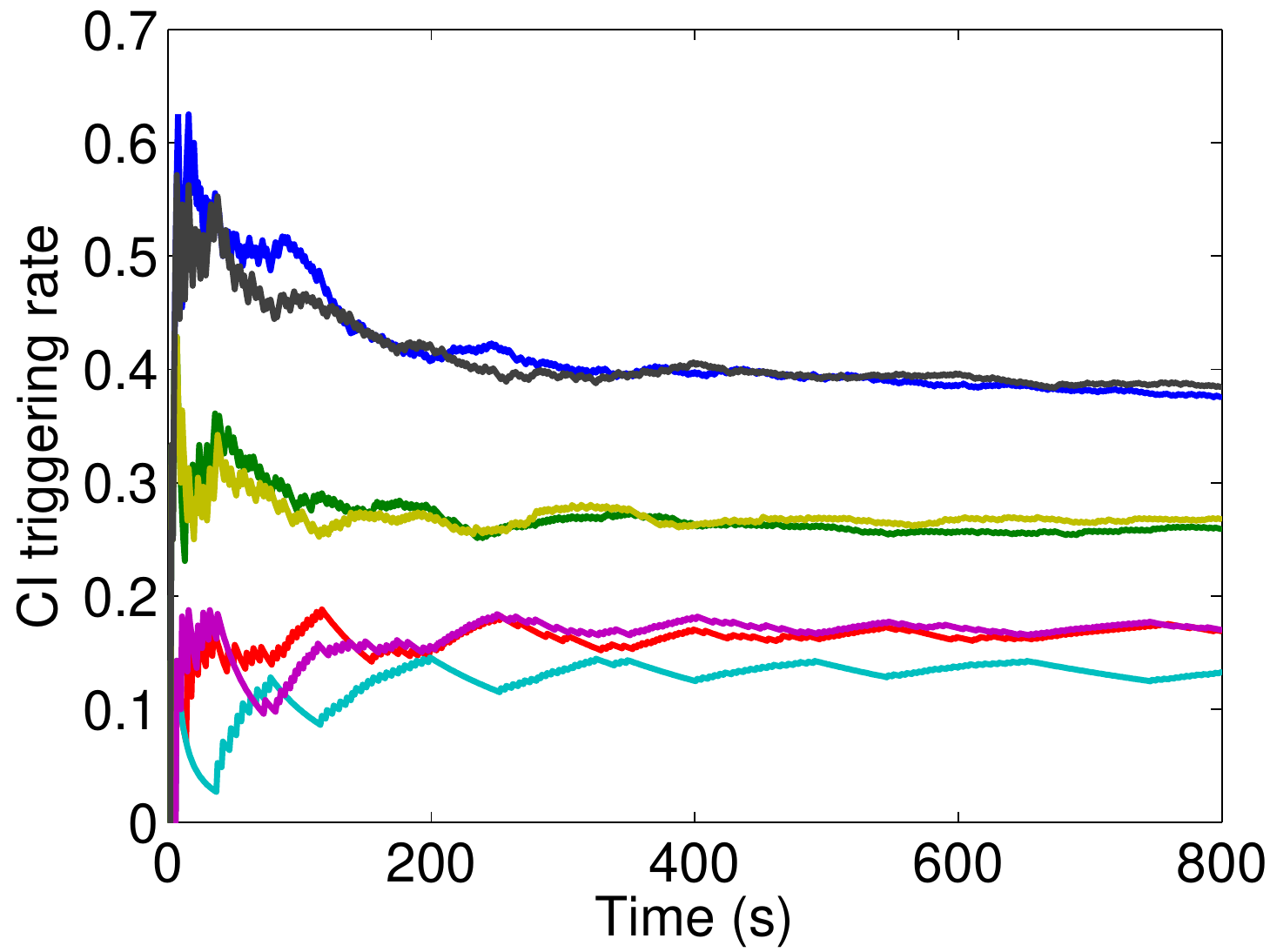}}

 \caption{Results for Example $2$.}\label{fig:CIboth}
\end{figure*}

Figure~\ref{fig:CIboth}(a) shows the trace of the network state covariance matrix estimated by each agent vs. time. 
Figure~\ref{fig:CIboth} (b) shows the CI triggering rate for each agent (number of CI updates divided by number of communication events) vs. time.  
In Figure~\ref{fig:CIboth} (a) and (b), the agents all
trigger a CI with their neighbors when the trace
of their covariance matrix is above the threshold of $\tau_i(k)
=\tau_{\text{goal}}^i=5~m^2$ (here, $\alpha_i = 1_{Nd}$). 
Note that some agents are more connected than others, 
and so not all are able to maintain their covariance trace below
$\tau_{\text{goal}}=5 m^2$.  As seen in Fig.~\ref{fig:CIboth}(b), 
since some agents are less connected (dark blue, black), they must perform CI fusion much more often, while the other agents perform CI only rarely. 
The robots that implement CI more frequently do not get enough information to lower their covariance
trace, because the robots which do not need it are not updating
frequently enough. 

Figures~\ref{fig:CIboth}(c) and (d) show the same metrics to illustrate the effect of 
the heuristic dynamics~\eqref{eq:CI-trigger} on the triggering threshold when $\epsilon_1= .1$ and $\epsilon_2 = .01$.  
By having some of the more connected agents(light blue, red, maroon)
lower their triggering threshold (and so trigger more frequently), we
see that all agents are able to maintain their state estimate
covariance below the desired threshold of $5~m^2$. Additionally, it can be
observed that the triggering rate for the less connected agents
(black, dark blue) decreases significantly because the more connected
agents increased their rates slightly. Setting $\epsilon_1$ affects the rate
of convergence.  In general, with $\epsilon_2=0$, agents converge to a
threshold less than $\tau_{\text{goal}}$.  By increasing $\epsilon_2$,
the threshold that agents converge to can be raised.  The parameters
used here were chosen based on hand tuning to achieve the best output
performance.  More formal analysis and strategies for automatic
adaptation are the subject of ongoing work.


\section{Nonlinear Model Simulation Results}\label{se:sims-nl}
This section analyzes the performance of the decentralized event-triggered CL algorithm for aerospace vehicle models and scenarios that are more realistic than those considered in the previous section. 
In these simulations, several factors are varied and the effects on the resulting CL performance are analyzed for two different simulated teams of multiple small fixed-wing aerial robots, each having 2D Dubins vehicles dynamics at the same constant altitude and nonlinear relative range and bearing measurement models. Roll and pitch variations for each vehicle are ignored for the sake of simplicity in these simulations, so that only planar translations, velocities, heading, and heading rates are considered. 
The factors varied in these simulations are: 
\begin{enumerate}[(i)]
	\item the tuning of the event-triggered innovation threshold $\delta$, to understand its effect on the resulting communication volume between robots; 
	\item the presence of an imperfect communication channel, i.e. where messages between robots are either completely received or dropped with some fixed probability;
	\item	vehicle maneuverability, to assess how the non-linear dynamics and measurement models impact the approximations made in the event-triggered EKF (which relies on linearization); 
	\item vehicle network topology, to assess how the combined flow of implicit/explicit data and state estimate information (via CI) impact estimation error convergence rates, especially if only one robot has access to GPS-like measurements. 
\end{enumerate}

In the following, the general nonlinear dynamics and measurement models are first described. 
The simulation studies for a 2-robot decentralized CL case are then discussed, under the assumption that both robots have GPS throughout in order to provide most of the insights for (i)-(iii) above. 
Finally, the results for 6-robot CL are discussed, where only a single designated robot has access to GPS throughout. 

\subsection{Nonlinear dynamics and measurement models}
For ease of notation in this section, we overload $x, y, \theta$ as the components of the state space where the subscript represents the robot. In the case of estimates, the superscript represents the robot that has that estimate.  For example, $y^j_i$ is robot $j$'s estimate of the $y$ location of robot $i$.  We also use $r$ and $b$ as the specific components of \emph{range} and \emph{bearing}, respectively.  Here, $r_i^j$ and $b_i^j$ is the range/bearing measurement of $i$ with respect to $j$. 
The Dubins vehicle dynamics for vehicle $i$ is given by:
\begin{align*}
x_i(k+1) =x_i(k) + V\cos(\theta_i(k)) \Delta t + v^x_i(k), \\
y_i(k+1) =y_i(k) + V\sin(\theta_i(k)) \Delta t + v^y_i(k), \\
\theta_i(k+1) = \theta_i(k) + u \Delta t + v^{\theta}_i(k),
\end{align*}
where $V$ is a constant velocity.  The measurement models for vehicle $i$ observing vehicle $j$ are given by:
\begin{subequations}
\begin{align}
r^j_i(k) = \sqrt{ ((x_i(k) - (x_j(k))^2 + ((y_i(k) - (y_j(k))^2}  + w^r_i(k) , \\
b^j_i(k) = \text{atan2}\Big (\frac{y_j(k) - y_i(k)} {x_j(k) - x_i(k)} \Big) +  w^b_i(k).
\end{align}
\label{eq:range-bearing}
\end{subequations}
Additionally, robots receive absolute (GPS-like) measurements of their location components $x$, $y$, and $\theta$ with additive noise:
\begin{subequations}
\begin{align}
g_i^x(k) = x_i(k) + w^x_i(k), \\
g_i^y(k) = y_i(k) + w^y_i(k), \\
g_i^{\theta}(k) = \theta_i(k) + w^{\theta}_i(k),
\end{align}
\label{eq:range-bearing}
\end{subequations}

\subsection{2-Agent Simulations}
The complete measurement vector for the 2-robot case is
\begin{align}
h_i(k) = [g_i^x(k),g_i^y(k),g_i^{\theta}(k),r^j_i(k),b^j_i(k)]^T.
\end{align}

\begin{figure}[h!]
 \centering 
   \includegraphics[width=.2\linewidth]{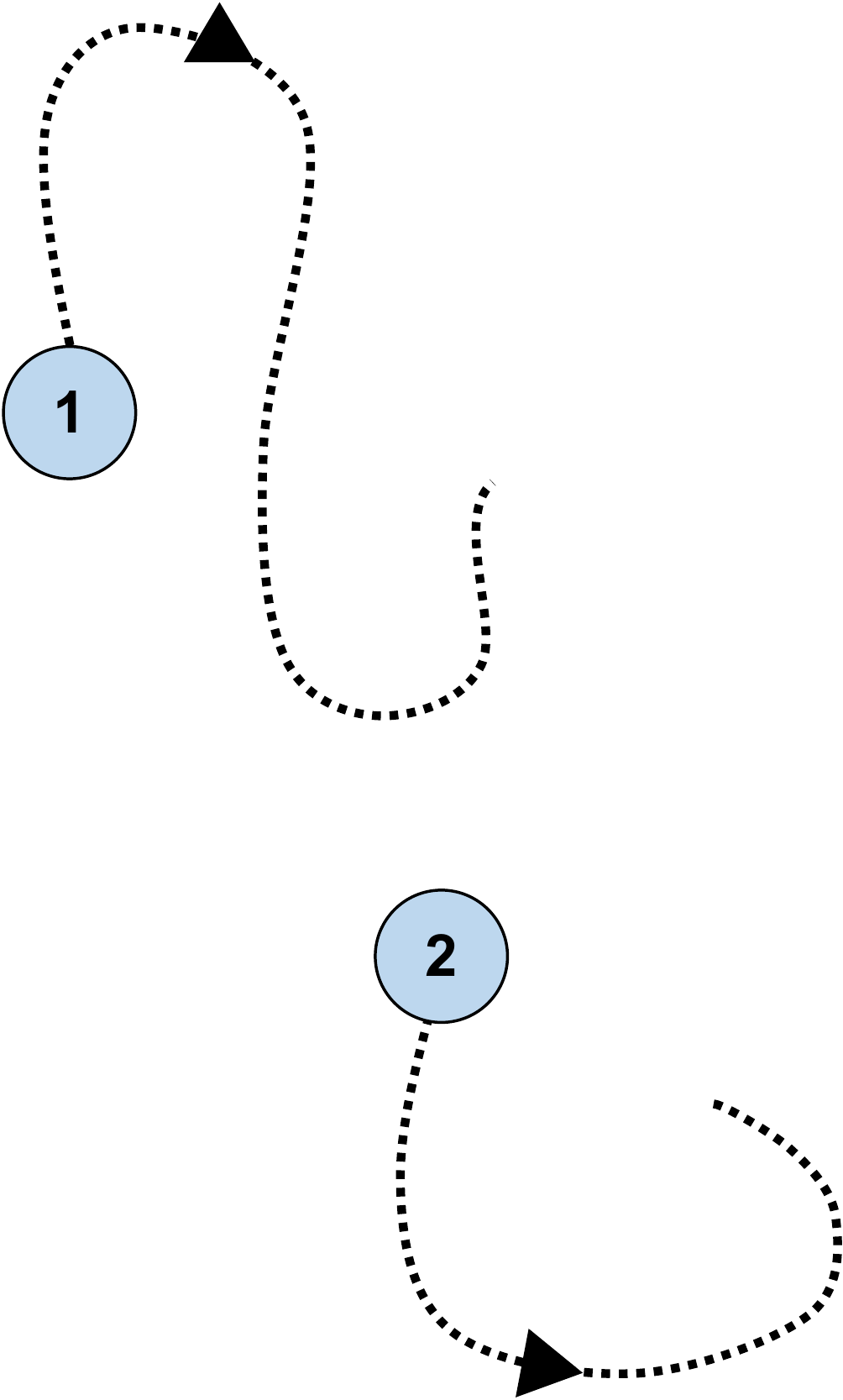}
\caption{Motions used in two robot simulations.
}\label{fig:2agent-scheme}
\end{figure}

Figure~\ref{fig:2agent-scheme} depicts the two robot simulations and we summarize the dynamics below.  The vehicles have a fixed speed of $1$~m/s and the control for bearing $\theta$ is $\text{sin}(0.5t(k)+\pi)$~rad/s for robot $1$ and $\text{sin}(0.1t(k))$ rad/s for robot $2$. The simulation runs for $10$ seconds and the discretization timestep is $0.05$ seconds. The vehicle dynamics noise covariance matrix is
\begin{align*}
Q=\begin{bmatrix}
0.01 &0 &0\\
0 &0.01 &0\\
0 &0 &0.001
\end{bmatrix} 
\end{align*}
and the measurement error variances are: $\rho_r=0.05 \ \text{m}^2$ (range), $\rho_b=0.05 \ \text{rad}^2$ (bearing), $\rho_{GPS}=1 \ \text{m}^2$ (GPS-position), and $\rho_{GPS}=1 \ \text{rad}^2$ (GPS-orientation).  The robots start at initial conditions of 
\begin{align*}
x_1=\begin{bmatrix}
-2 \\ 
12 \\ 
2\pi/3
\end{bmatrix}, \ 
x_2=\begin{bmatrix}
0\\ 
5\\ 
-\pi/2
\end{bmatrix} 
\end{align*}
with initial covariance estimates $P_1=P_2=\text{I}_{3 \times 3}$. 

In this section, the simulations allow for the possibility of a communication channel where measurements occasionally are dropped.  For simplicity, a model parameterized by a communication success probability ($CP$) is used, where each measurement explicitly sent will be received with probability $CP$.

\subsubsection*{Example 1: Effects of communication rate}
Here we present simulations on the effects of performance and consistency as a function of the communication threshold $\delta$ when the communication success probability $CP$ is $1$.

\begin{figure}[h!]
 \centering 
   \includegraphics[width=.32\linewidth]{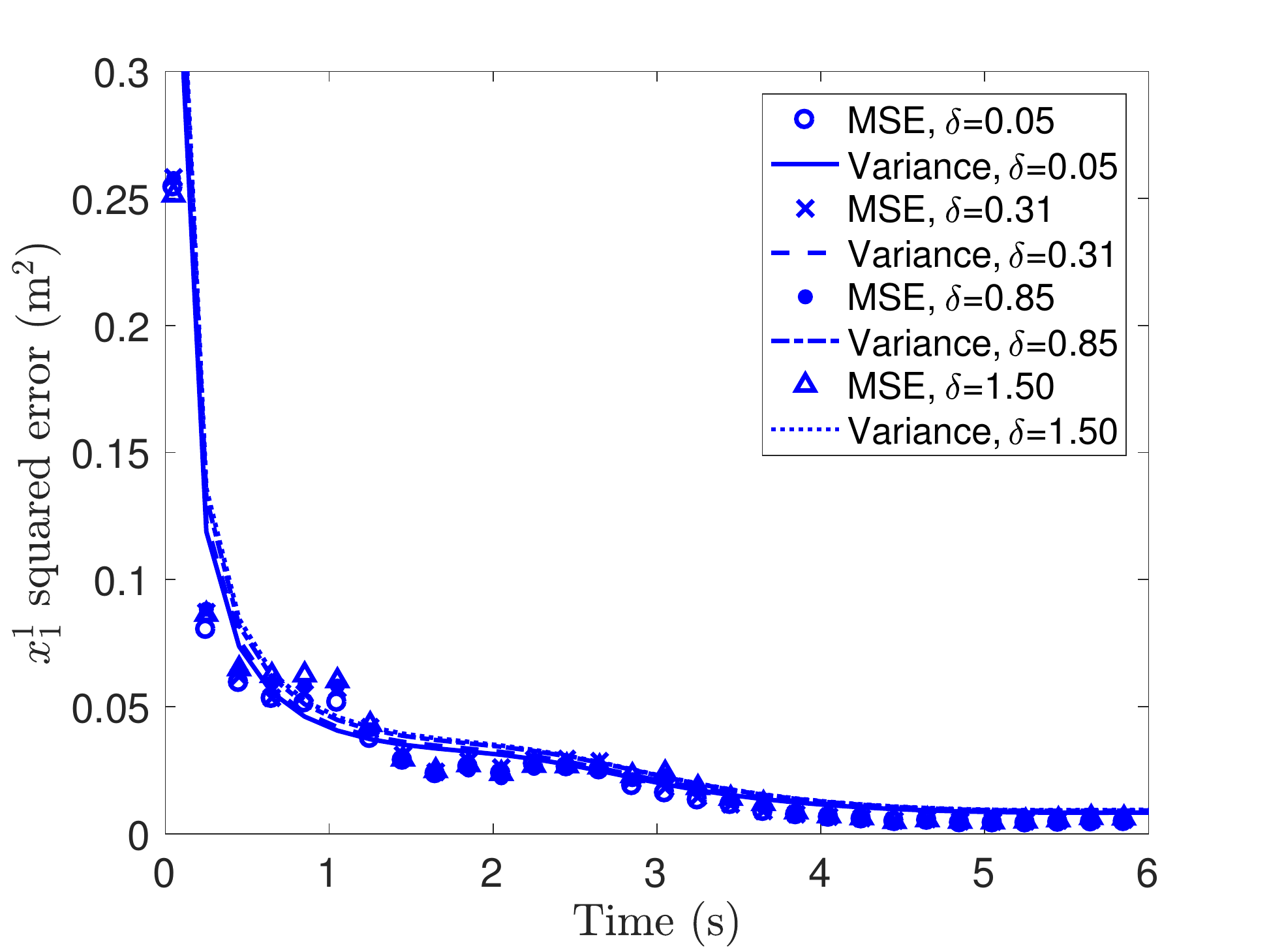}
   \includegraphics[width=.32\linewidth]{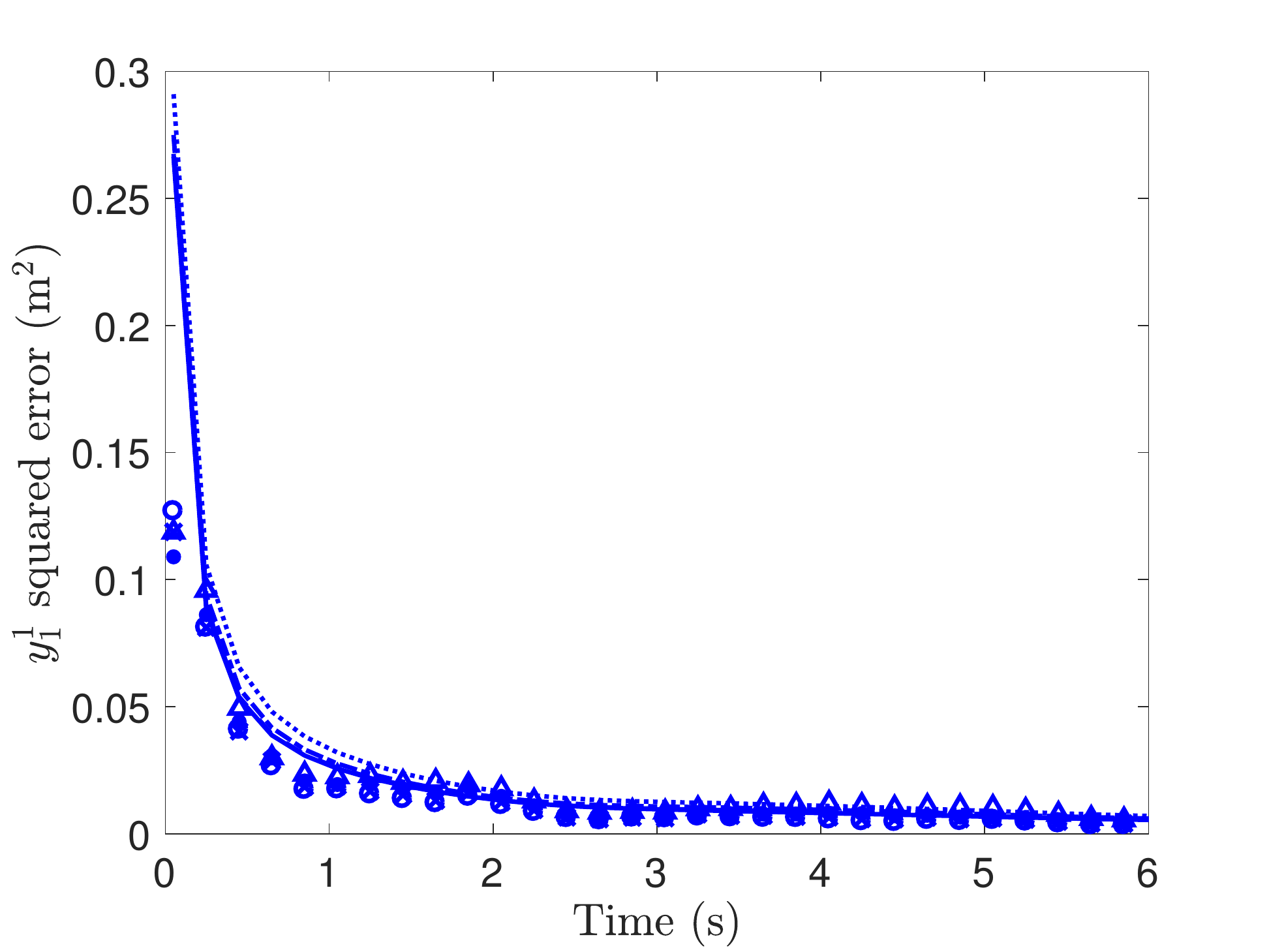}
   \includegraphics[width=.32\linewidth]{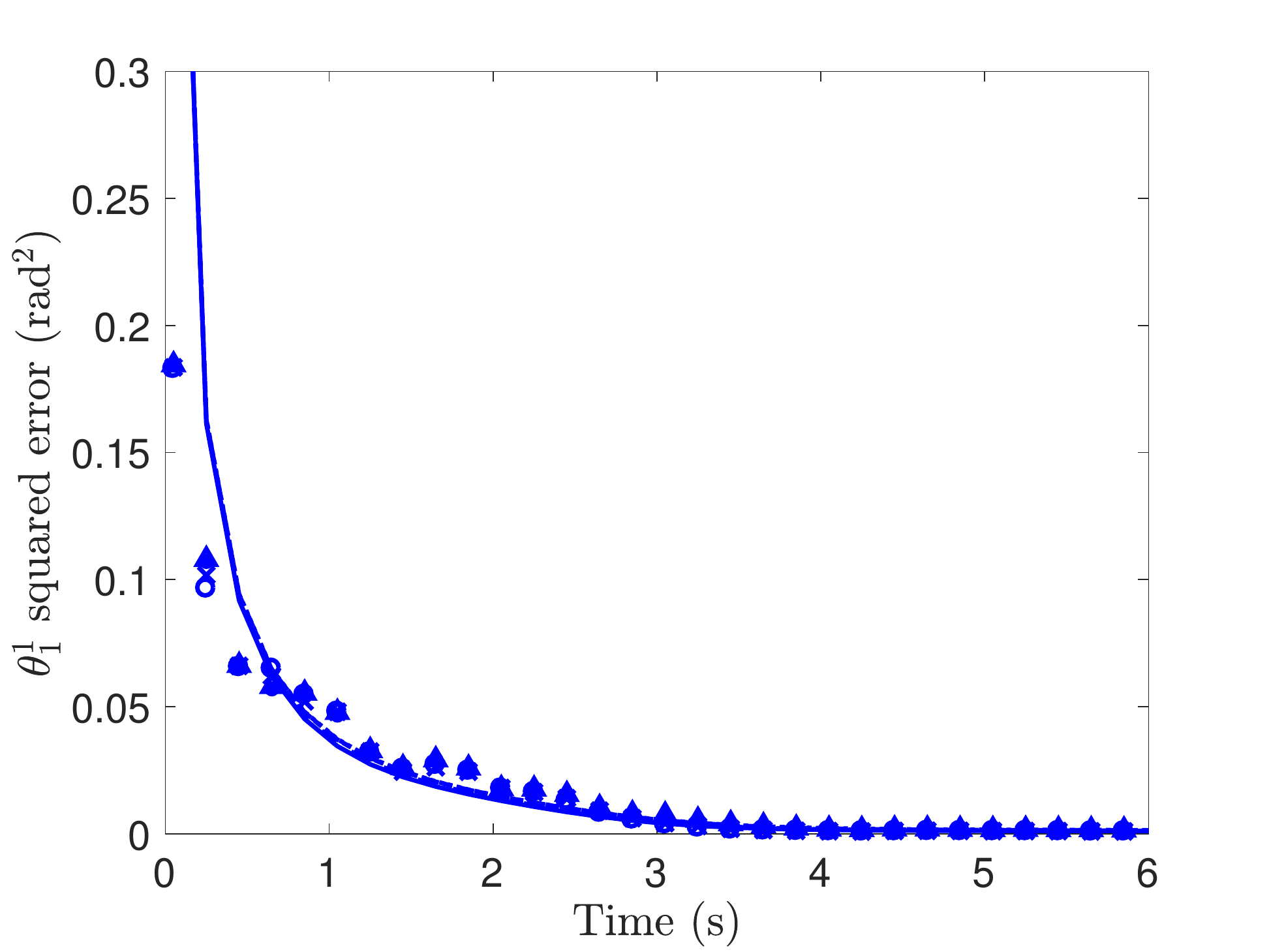}
   \caption{MSE for robot $1$ tracking its own pose.}
   \label{fig:consistency-100CS-own}
\end{figure}

\begin{figure}[h!]
   \includegraphics[width=.32\linewidth]{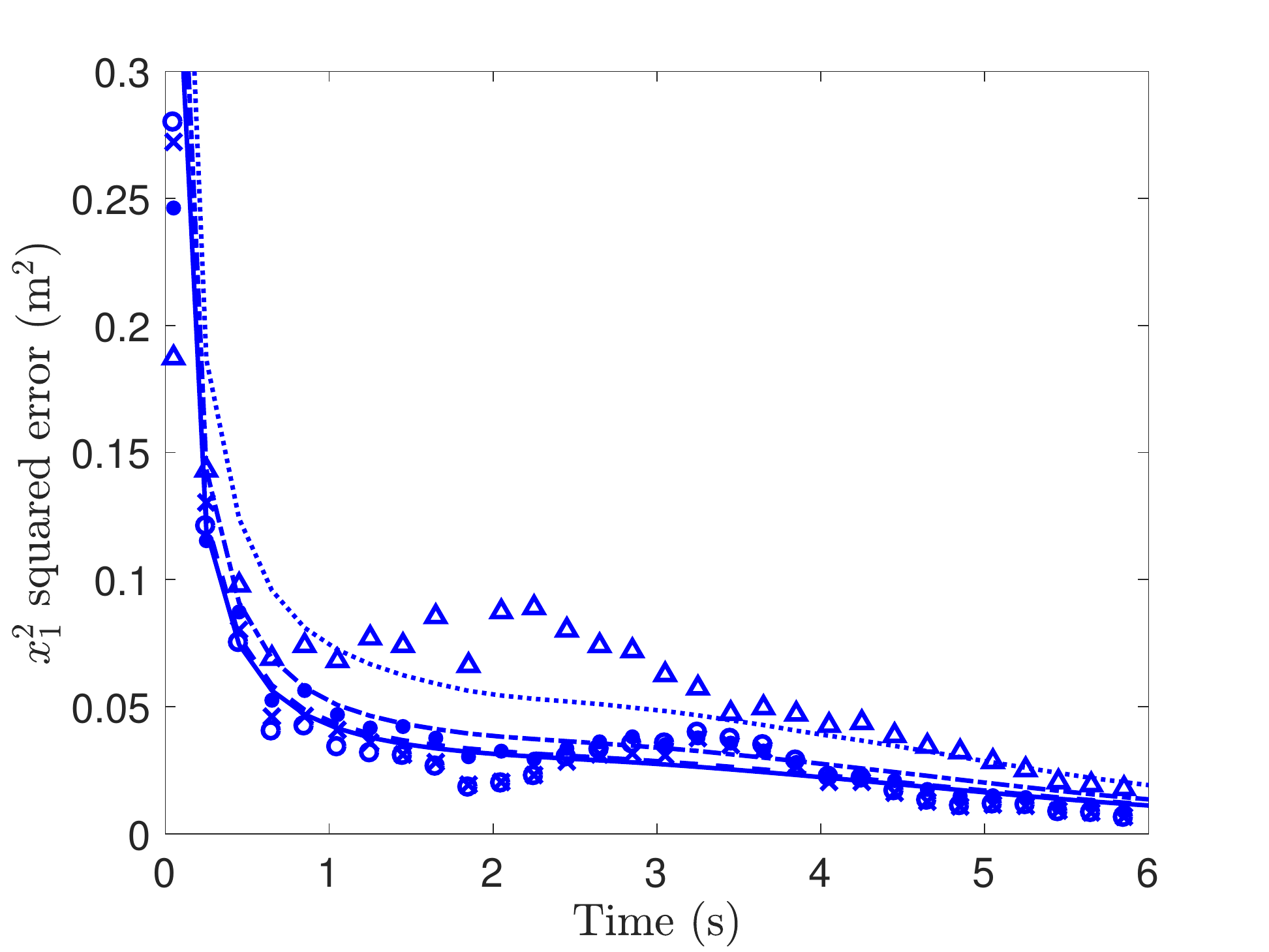}
   \includegraphics[width=.32\linewidth]{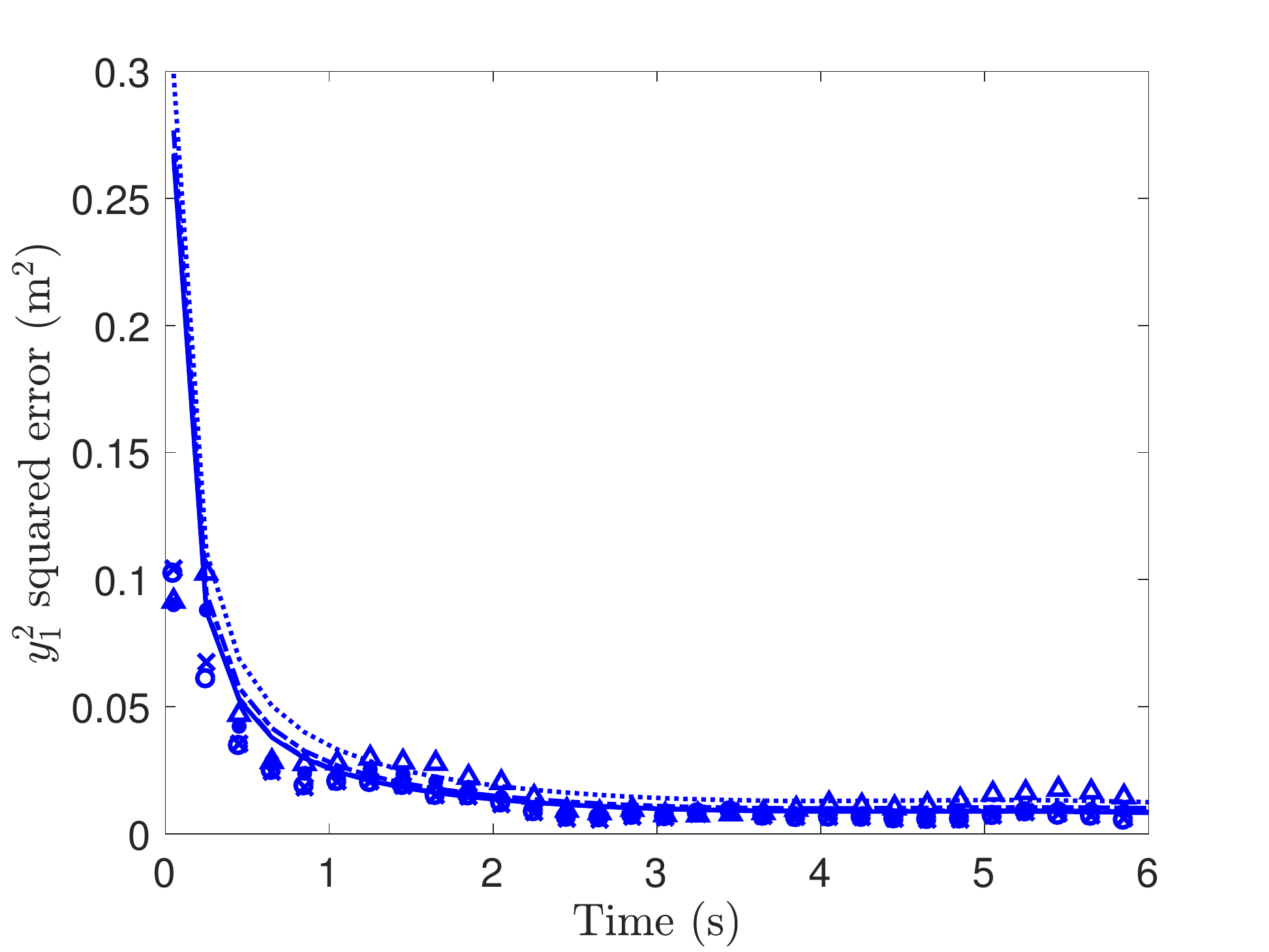}
   \includegraphics[width=.32\linewidth]{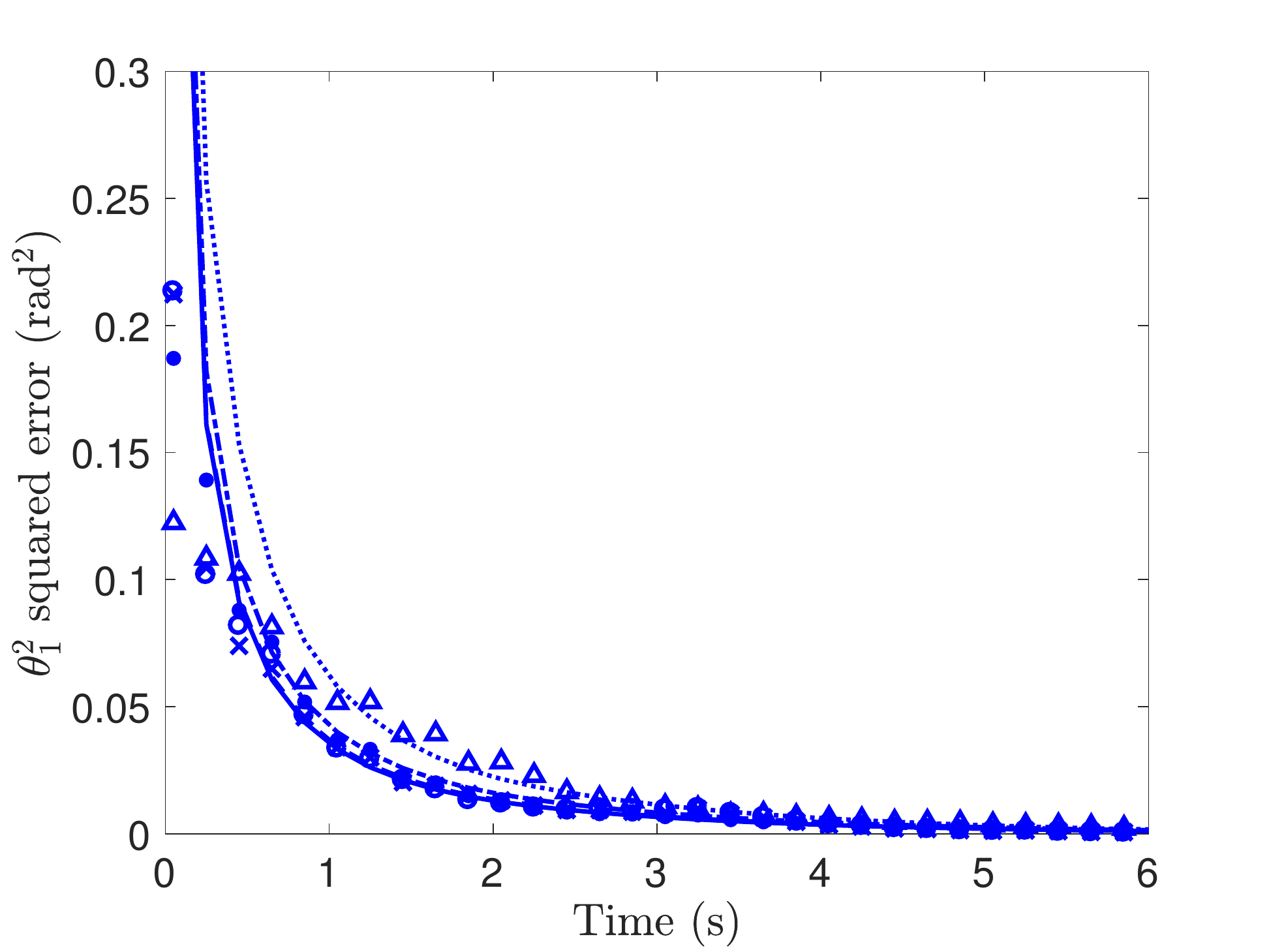}
   \caption{MSE for robot $1$ tracking robot $2$'s pose.}
   \label{fig:consistency-100CS-mut}
\end{figure}

From Figures~\ref{fig:consistency-100CS-own} and \ref{fig:consistency-100CS-mut}, one can see that for a $2$ robot setup with robot $1$ tracking both its own location as well as robot $2$'s state, there is no apparent loss of consistency as one increases $\delta$ (decreases the average explicit communication rate) because the mean squared error matches the predicted covariance showing that even with the nonlinear dynamics and measurement models, our algorithm remains consistent.

\begin{figure}[h!]
\centering
\includegraphics[width=.55\linewidth]{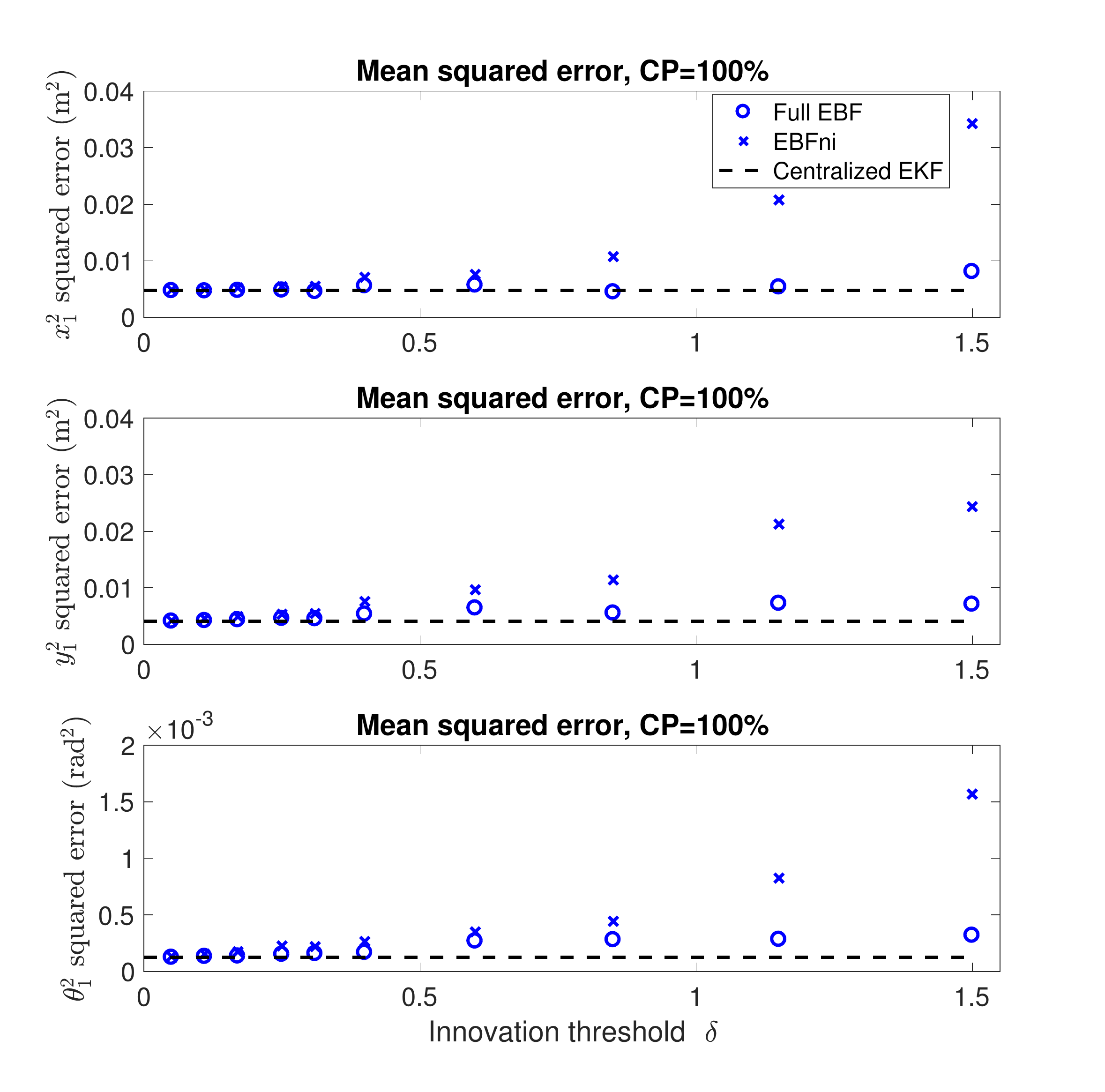}
\vspace{0ex}
\caption{MSE of full event-based filter vs. filter that fuses only explicit data. 
}
\label{fig:event-vs-non}
\end{figure}

\begin{figure}[h!]
\centering
\includegraphics[width=.5\linewidth]{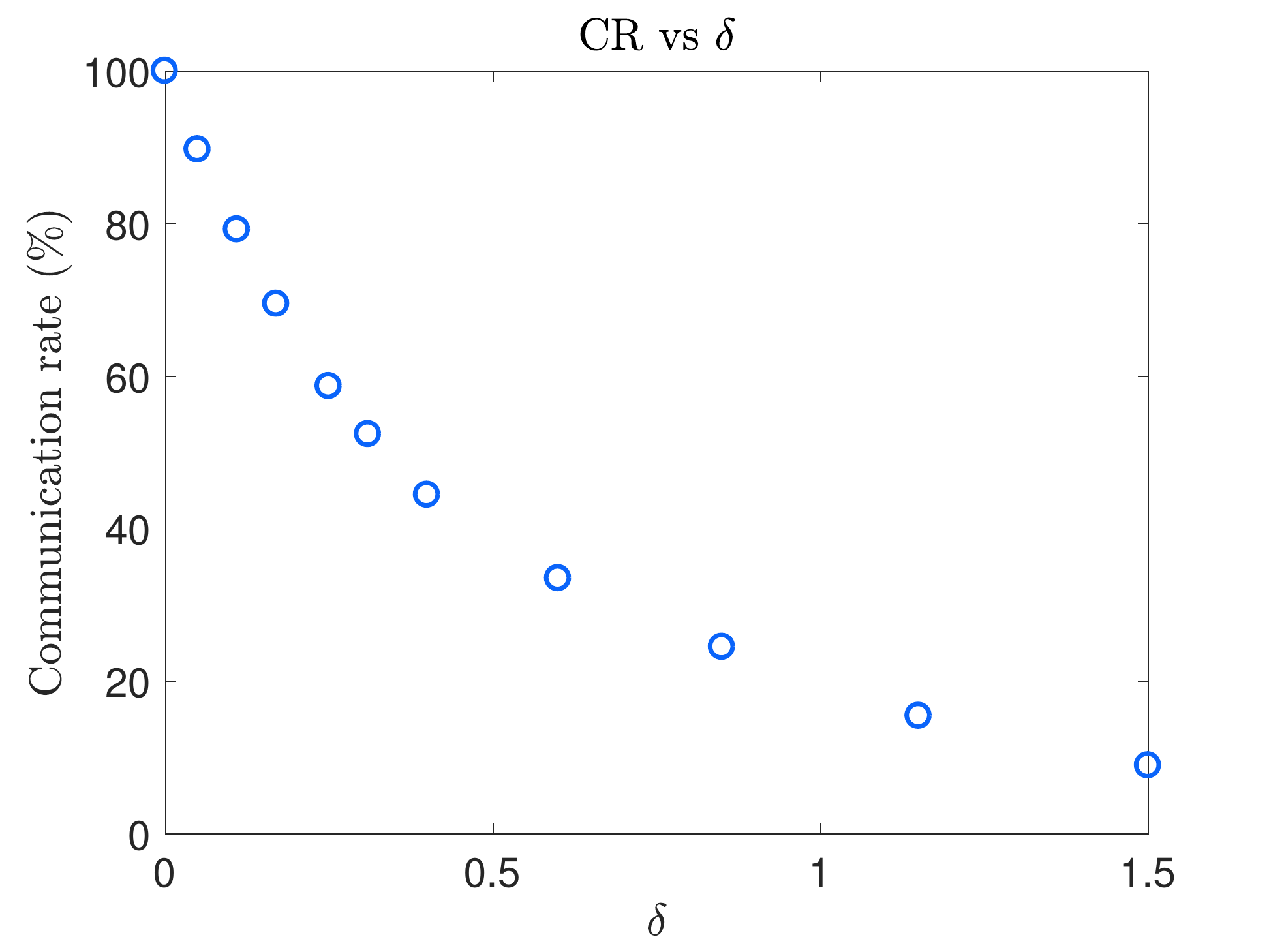}
\caption{Average communication rate between robots 1 and 2 as a function of $\delta$.
}
\label{fig:cr-vs-delta}
\end{figure}

We also compare the MSE of our event-based algorithm with the event-based filter where implicit measurements are not fused in Figure~\ref{fig:event-vs-non} for a two-robot setup.  One can see that as $\delta$ increases and so fewer measurements are sent, our algorithm steadily outperforms the version without the implicit information showing that fusing the implicit information in the measurements not sent improves performance. Furthermore, for small $\delta$ values, both filter's performance will be very close to an EKF that fuses all measurements explicitly ($\delta = 0$).  For larger $\delta$ values, our algorithm has moderate increases in MSE for the benefit of requiring much fewer measurements sent. Figure~\ref{fig:cr-vs-delta} illustrates the non-linear relationship for the average communication rate between robots 1 and 2 versus $\delta$.

%
%

\subsubsection*{Example 2: Effect of communication probability}
\begin{figure}[h!]
 \centering \subfigure[]{
   \includegraphics[width=.32\linewidth]{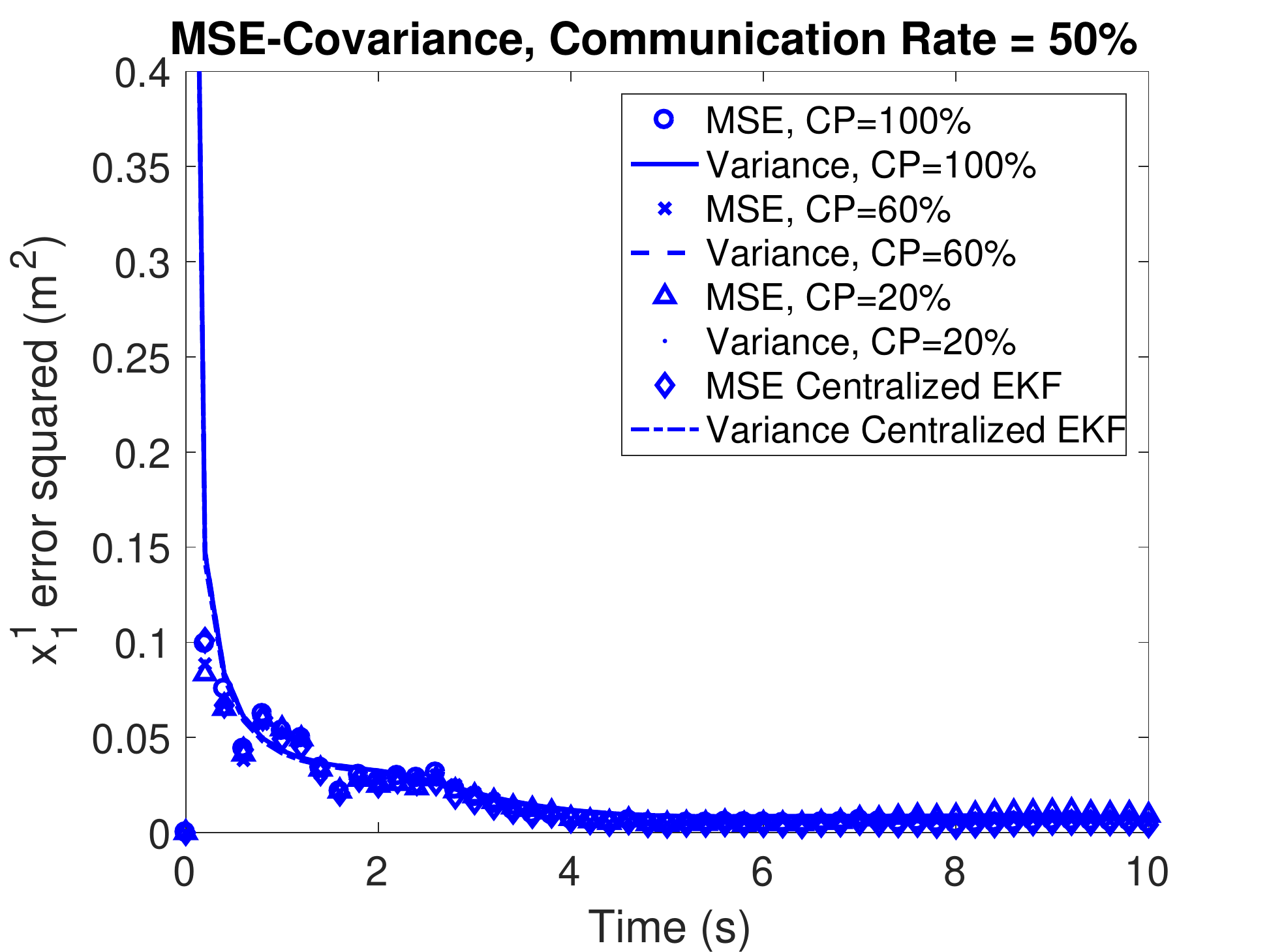}}
\vspace{0ex}
\subfigure[]{
  \includegraphics[width=.32\linewidth]{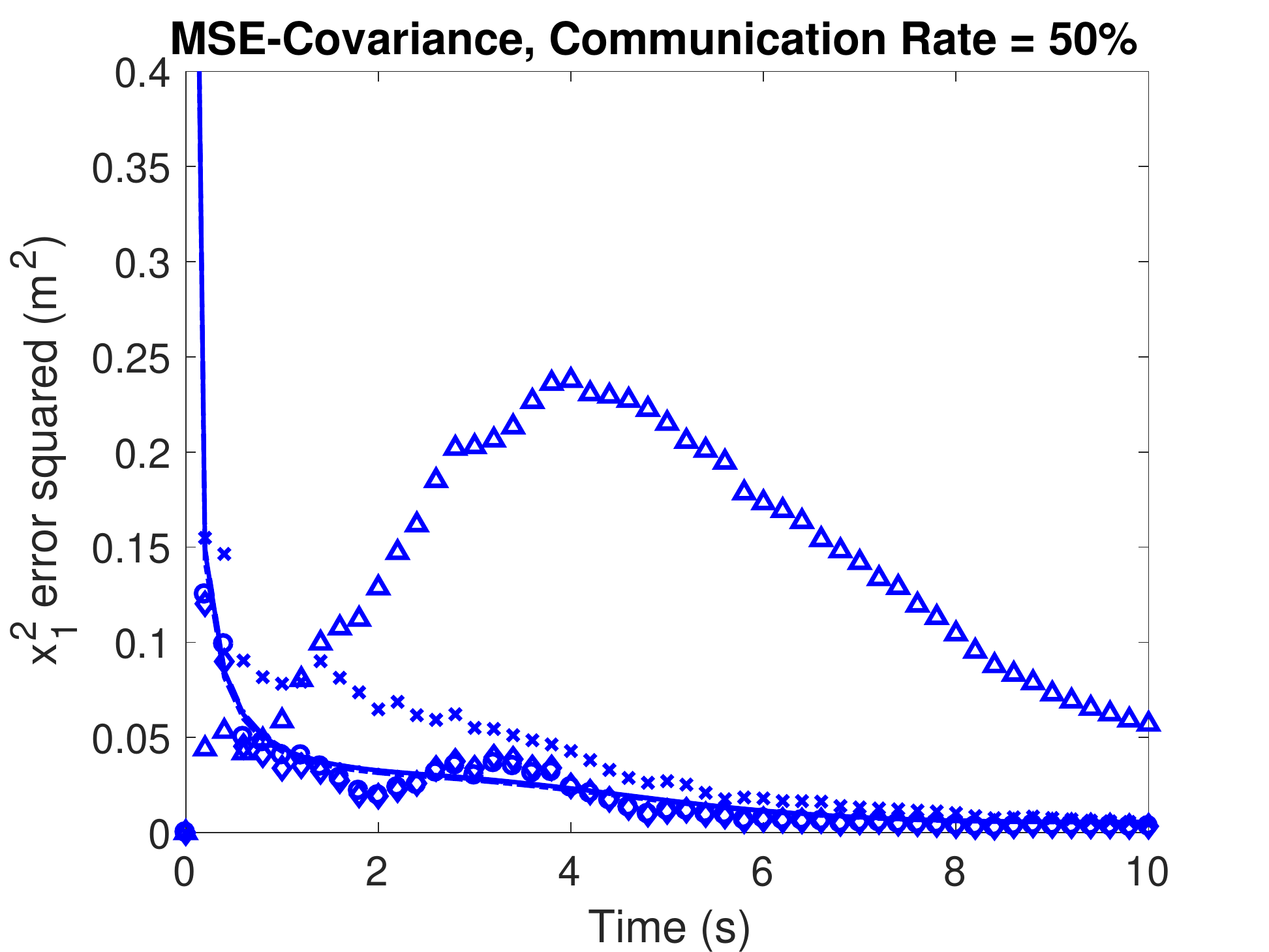}}
\vspace{0ex}
\subfigure[]{
  \includegraphics[width=.32\linewidth]{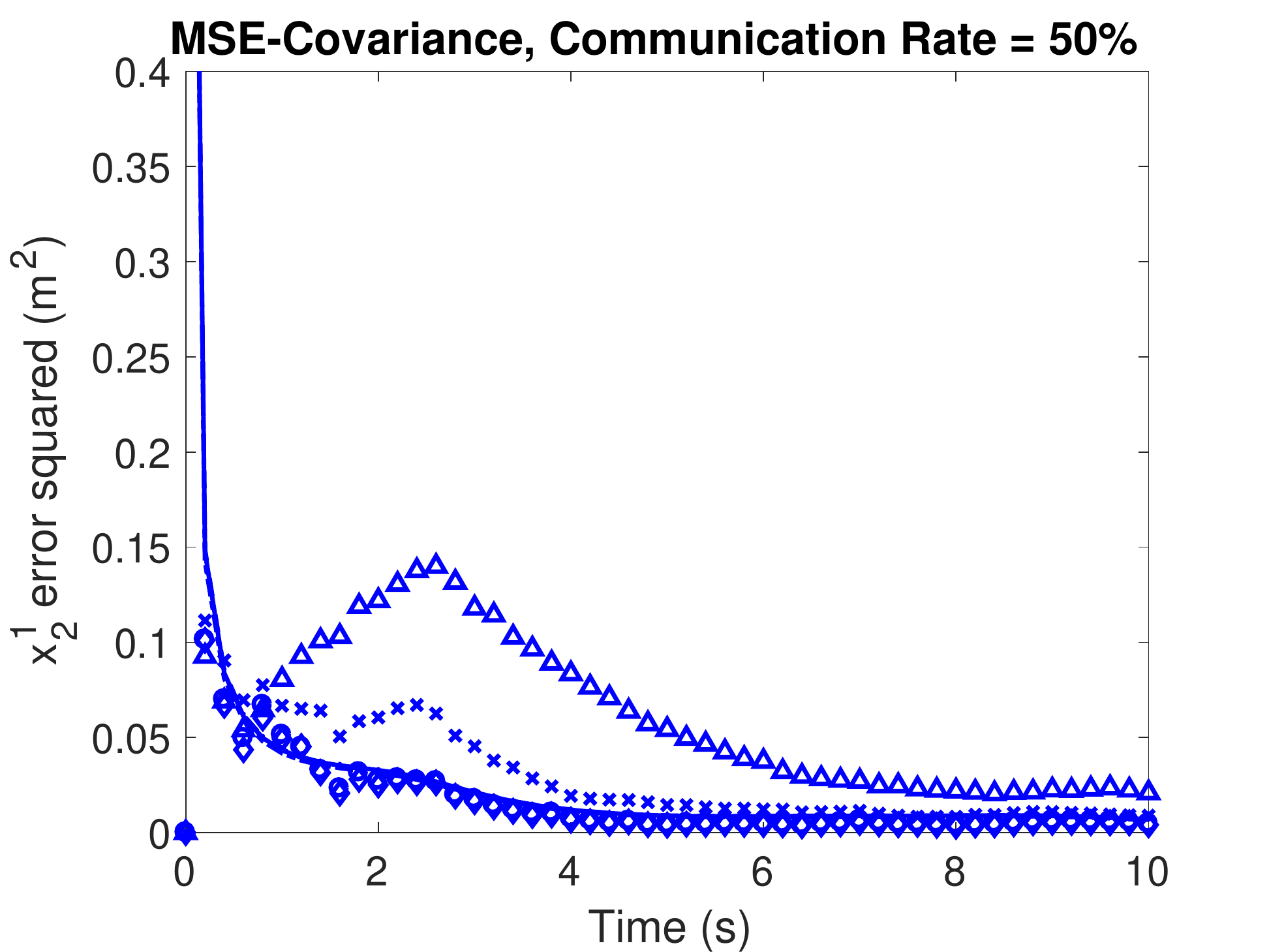}}
\vspace{0ex}
\caption{ Results for 50\% average explicit data throughput, with random losses from  $CP<100$. 
}\label{fig:consistency-50CR}
\end{figure}

In this example, we examine the effect of communication channels where measurement drops are possible.  Since this algorithm relies on the knowledge that a measurement not received was intentionally censored, we investigate the robustness of our algorithm to environments where this assumption is violated. Figure~\ref{fig:consistency-50CR} depicts the performance of two robots with 50\% measurements being explicitly sent on average for a fixed $\delta$, but where a variable number of messages are dropped with some Communication Probability $CP$. Figure~\ref{fig:consistency-50CR}(a) shows robot $1$'s estimate of its own $x$ location.  Since it takes and fuses measurements of this quantity at every time step, the performance is unaffected by the increased $CP$.  However Figure~\ref{fig:consistency-50CR}(b) and (c), show that robot $1$'s estimate of robot $2$'s $x$ location as well as robot $2$'s estimate of robot $1$'s $x$ location is dependent on $CP$. As one decreases $CP$, there is an increasing gap between the predicted covariance and the true MSE.

\begin{figure}[h!]
 \centering \subfigure[]{
   \includegraphics[width=.38\linewidth]{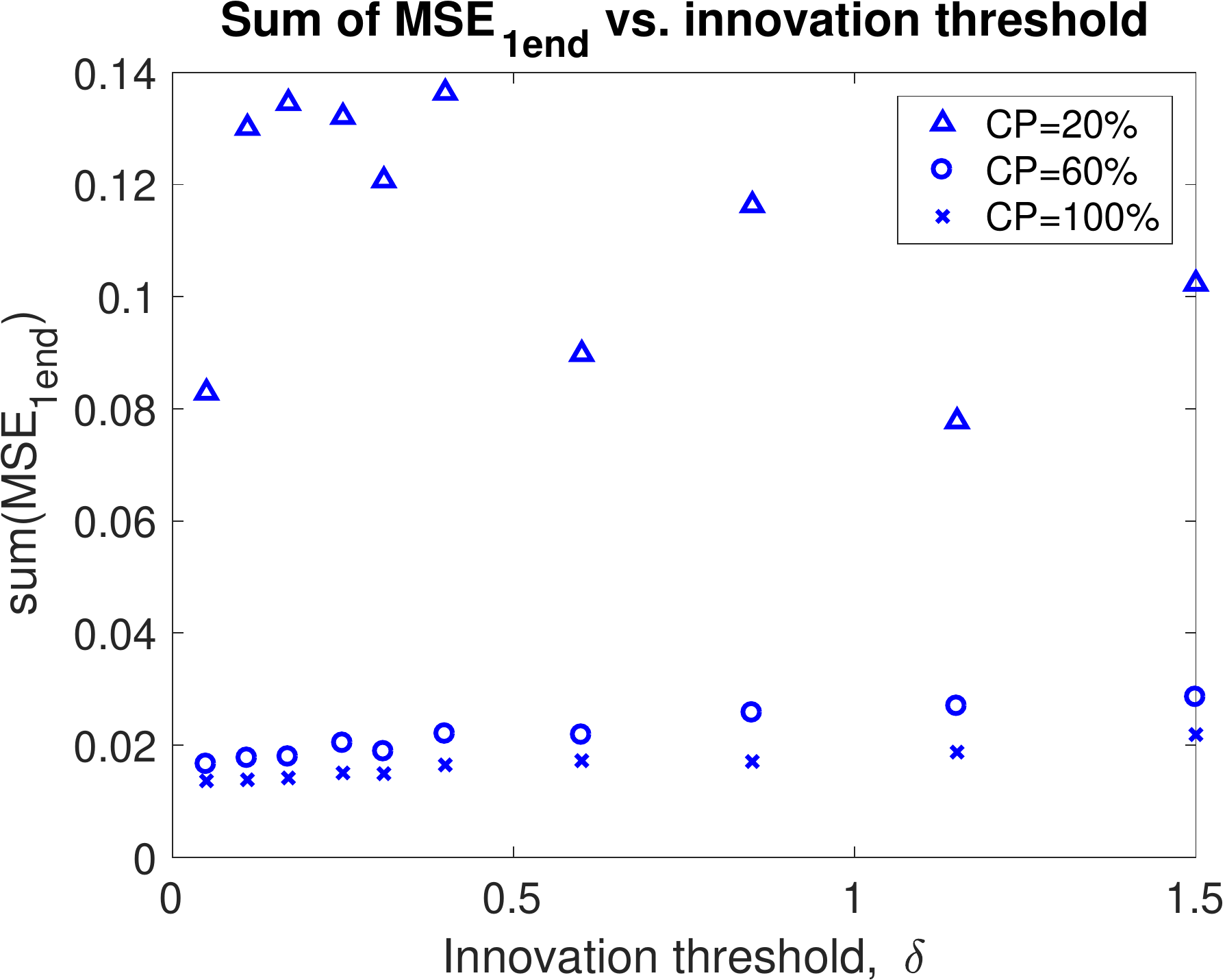}}
\vspace{0ex}
\subfigure[]{
  \includegraphics[width=.38\linewidth]{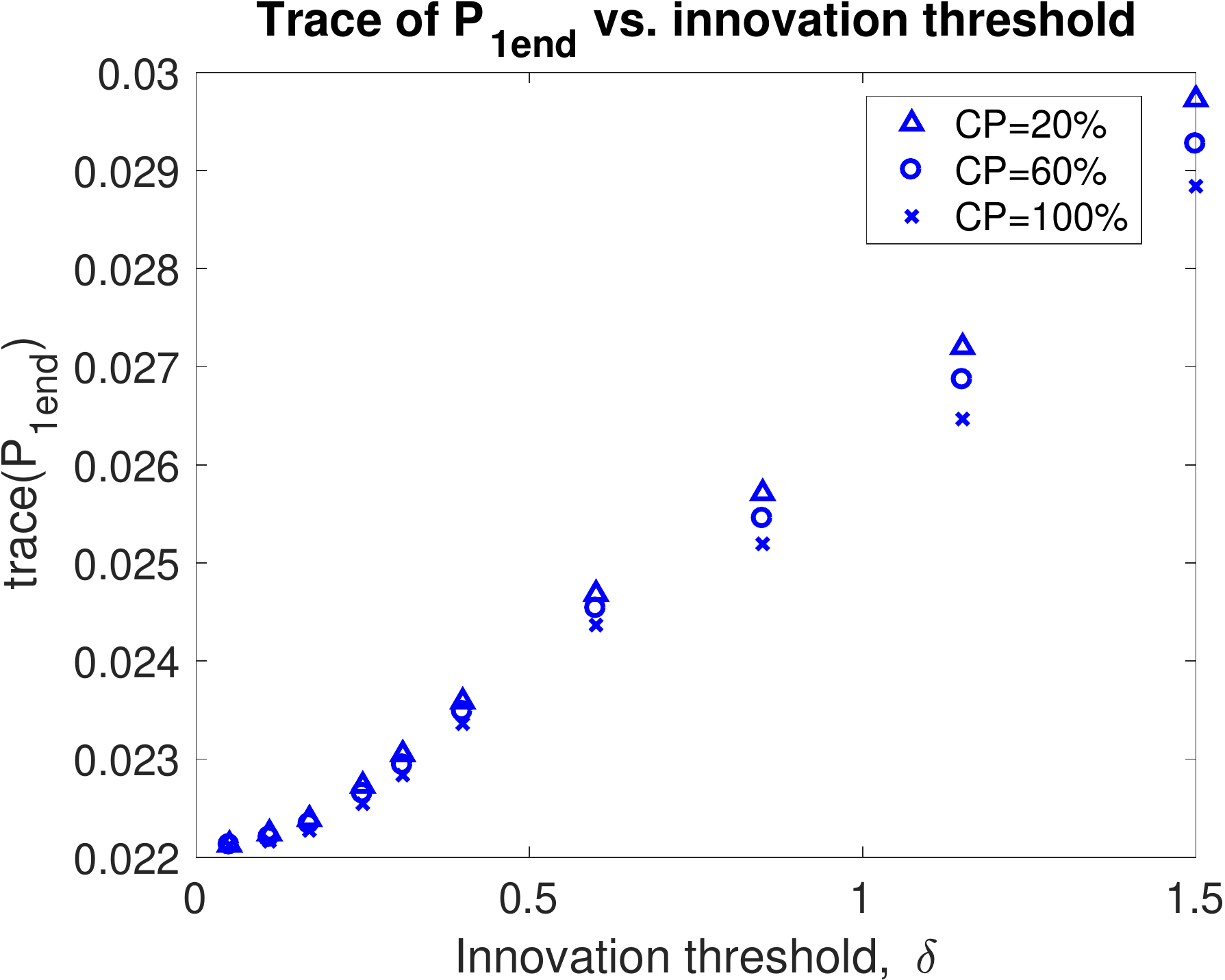}}
\vspace{0ex}
\caption{ Actual vs. estimated error covariance traces vs. $\delta$ for imperfect communications.
}\label{fig:PMSE-vs-Threshold}
\end{figure}

\begin{figure}[htb]
 \centering 
   \includegraphics[width=.5\linewidth]{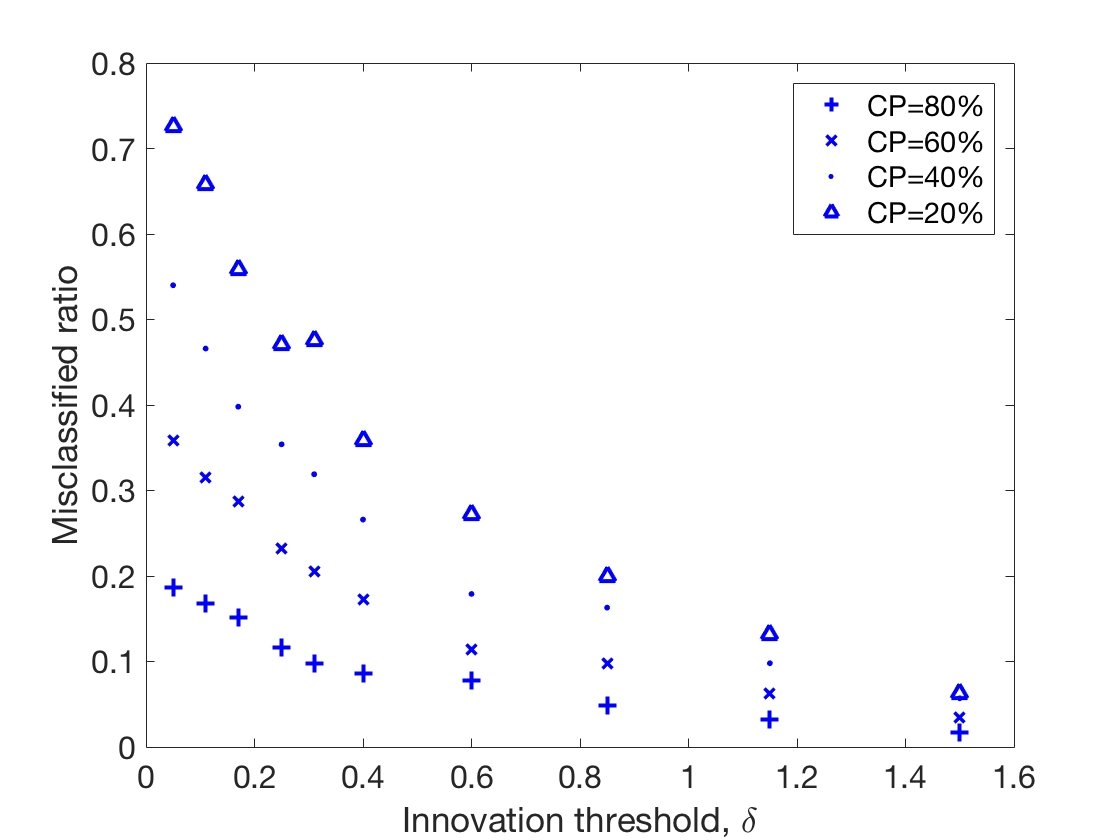}
\caption{Confusion ratio for imperfect communication scenarios. 
}\label{fig:misclas}
\end{figure}

Figure ~\ref{fig:PMSE-vs-Threshold}(a) depicts the final sum of MSEs for robot $1$ for several $CP$ values (30 sample runs each) , showing that the error increases as more measurements are dropped; Figure (b) shows the trace of the estimated covariance matrix for the same trials. Note the difference in scale between the plots and that for $CP=20\%$, the filter is overconfident in the filter's estimates.  This is because the filter is interpreting dropped measurements as implicit measurements. Figure~\ref{fig:misclas} depicts the confusion ratio as a function of the threshold parameter $\delta$ and Communication Probability ($CP$) showing that as $\delta$ increases, fewer measurements are misclassified. 

Figure~\ref{fig:misclas} depicts the confusion ratio, i.e. the number of explicit measurements dropped divided by the number of total measurements (implicit and explicit).  As $\delta$ increases, fewer measurements are sent explicitly, and therefore, cannot be dropped; this leads to a reduction in the number of dropped data elements that are misinterpreted as implicit measurements.

\begin{figure}[h!]
 \centering 
\subfigure[]{
  \includegraphics[width=.55\linewidth]{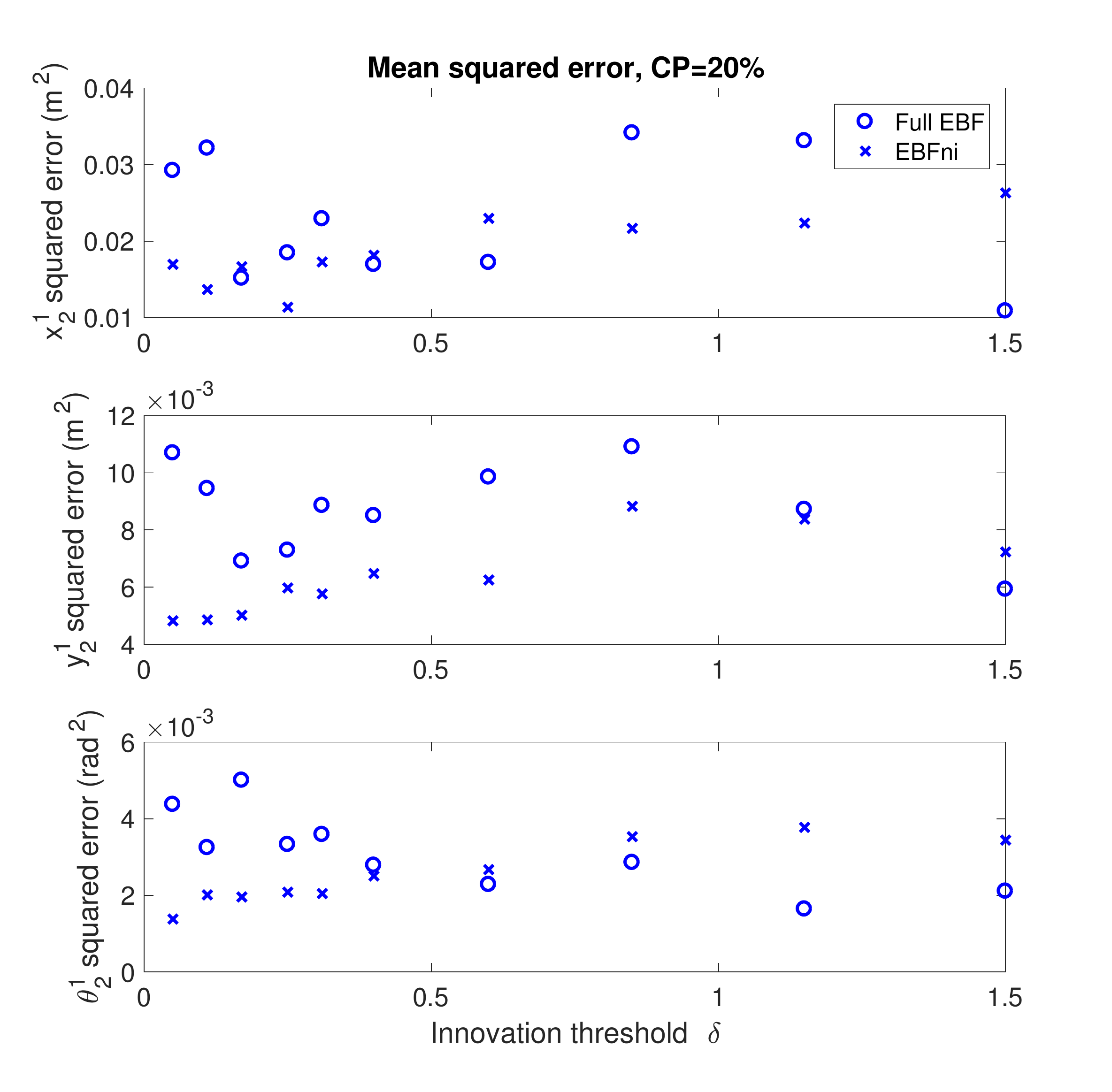}}
\vspace{0ex}
\subfigure[]{
  \includegraphics[width=.55\linewidth]{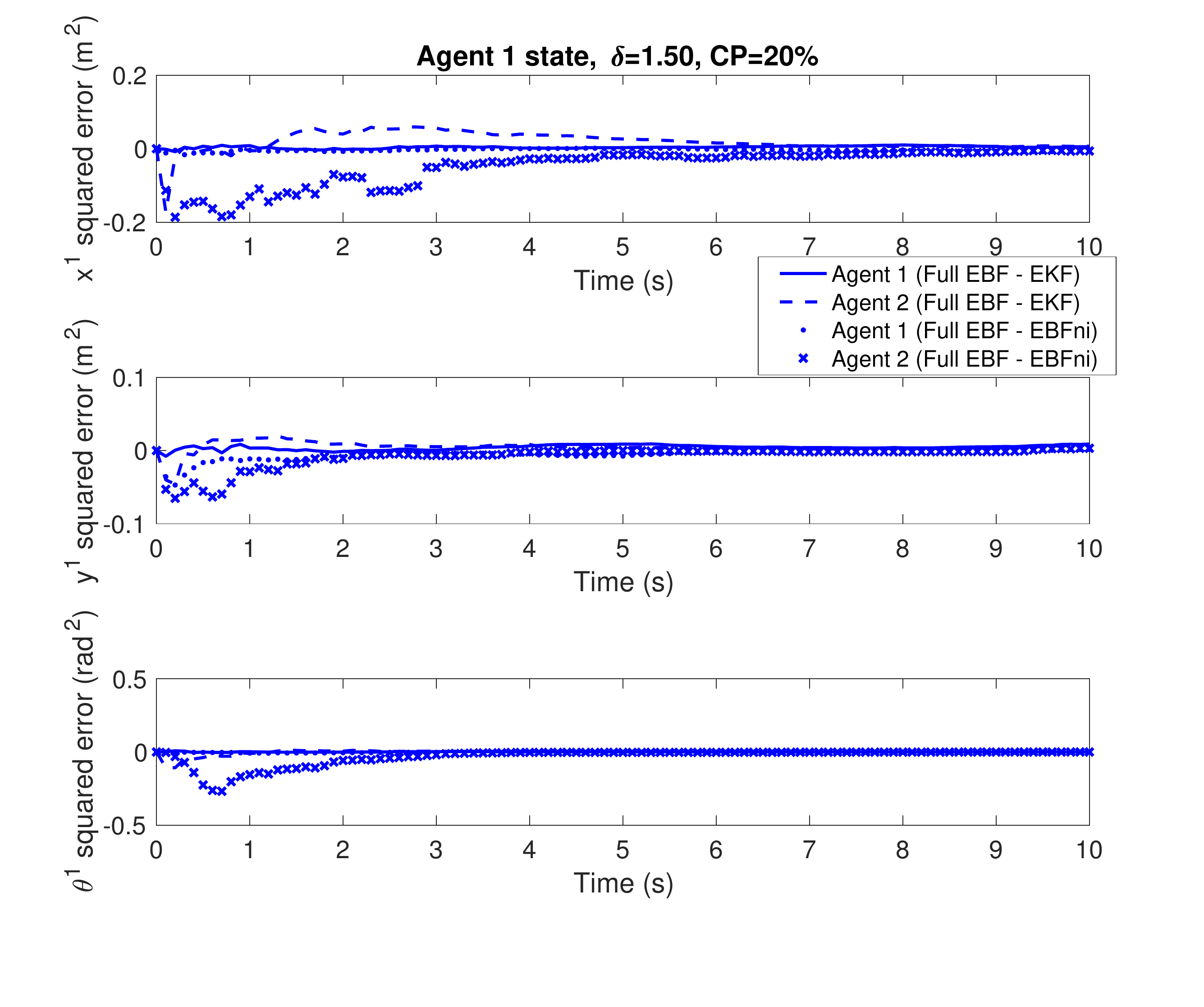}}
\caption{ Imperfect communication scenario results for full event-based filter vs. explicit-data-only filter.
}
\label{fig:ours-vs-noneg}
\end{figure}

Figure~\ref{fig:ours-vs-noneg} compares our event-based filter (Full EBF) with one which only fuses explicit data that is sent by the Full EBF; the $CP=20\%$  for both filters. 
Figure ~\ref{fig:ours-vs-noneg} (a) shows the MSE for robot 2's estimate of robot 1 states (30 sample runs) at different $\delta$'s. 
Figure ~\ref{fig:ours-vs-noneg} (b) shows a plot of squared errors vs. time for typical sample run. 
 These results contrasts with Figure~\ref{fig:event-vs-non}, which highlights the improved performance of fusing the implicit information when $CP$ is $1$.  Figure~\ref{fig:ours-vs-noneg}(a) and (b) both show a limitations of our event-based filter.  Since the $CP$ is only $20\%$, some of the messages are dropped but our filter interprets them as being intentionally implicitly sent, i.e. misinterpreted as seen in Figure~\ref{fig:misclas}.  This leads to larger errors in Full EBF and smaller errors when `implicit information' (i.e. real implicit information or dropped measurements) is ignored (EBFni).  Figure~\ref{fig:ours-vs-noneg}(a) also implies that for low $CP$, increasing innovation parameter $\delta$ improves the performance.  This seems to be explained by the fact that increasing $\delta$ decreases the chance the an explicit measurement is dropped, and therefore, misinterpreted as an implicit measurement, as seen above in Figure~\ref{fig:misclas}.

\subsubsection*{Example 3: Effect of communication topology}

\begin{figure}[htb]
 \centering 
   \includegraphics[width=.75\linewidth]{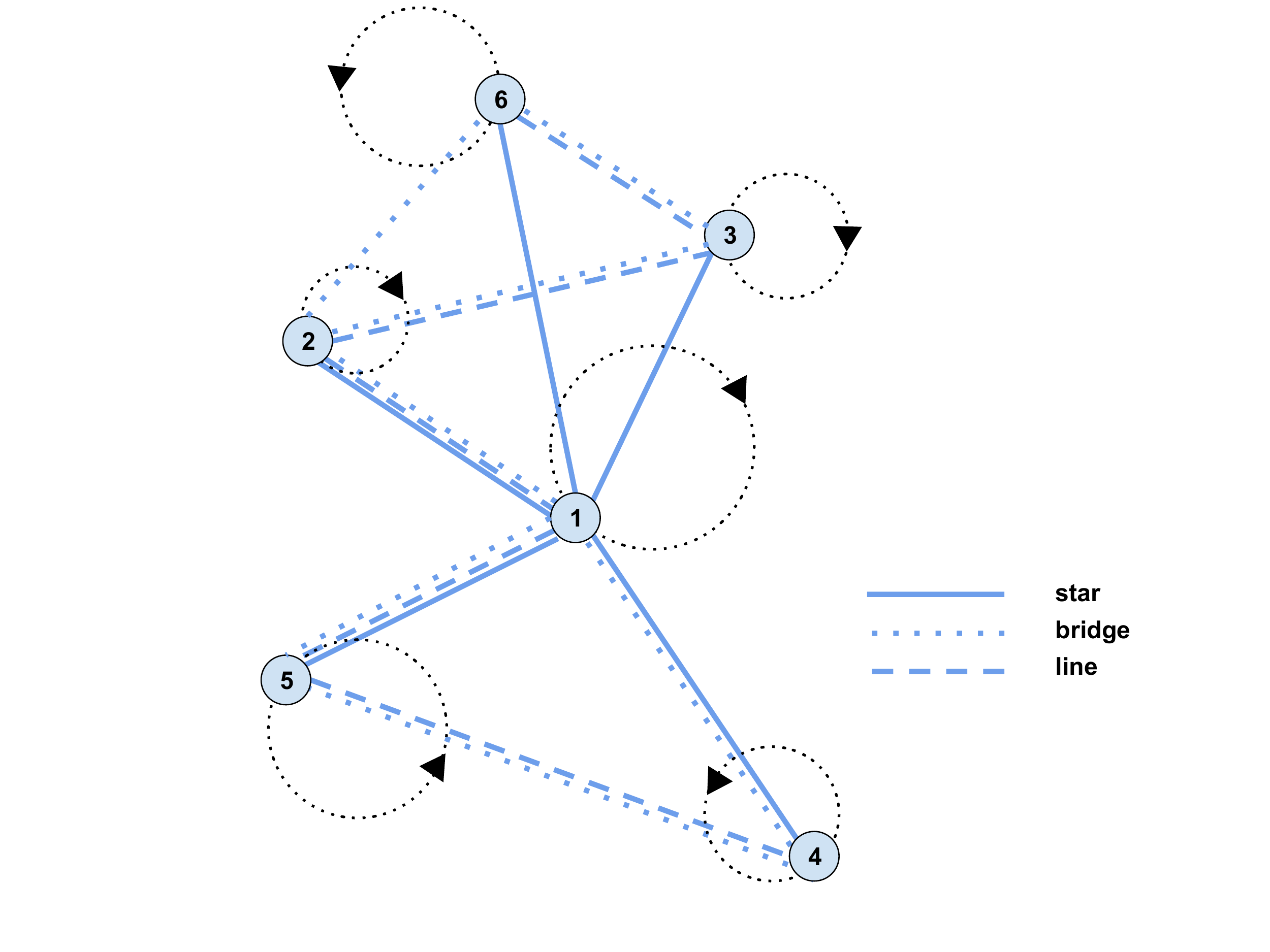}
\caption{Motions for 6-robot scenario with three different topologies (star, bridge, chain).
}\label{fig:6agent-scheme}
\end{figure}

In this section, the performance of the event-based filter for a larger network consisting of 6 robots is studied. Since our ultimate goal is to see the accuracy and consistency of this algorithm in scenarios which are as realistic as possible, additional elements such as multiple robots, different communication graphs and non-ubiquitous GPS are added in the simulations. Figure~\ref{fig:6agent-scheme} depicts the 6-robot simulations for the three different graphs or topologies: star, bridge and line.

As before, the robots' dynamics model is Dubins.  The robots' fixed velocities and turning rates are given by 
\begin{align*}
u_1&=\begin{bmatrix}
1 \text{m/s}\\
0.5 \text{rad/s}
\end{bmatrix}, \ 
u_2=\begin{bmatrix}
0.5 \text{m/s}\\
1 \text{rad/s}
\end{bmatrix}, \ 
u_3=\begin{bmatrix}
1 \text{m/s}\\
1 \text{rad/s}
\end{bmatrix}  \\
u_4&=\begin{bmatrix}
0.5 \text{m/s}\\
0.5 \text{rad/s}
\end{bmatrix},
u_5=\begin{bmatrix}
0.7 \text{m/s}\\
0.1 \text{rad/s}
\end{bmatrix},
u_6=\begin{bmatrix}
0.5 \text{m/s}\\
0.5 \text{rad/s},
\end{bmatrix} 
\end{align*}
held constant throughout the simulations. The simulation duration is $10$ seconds and the time step is $0.1$ seconds. The dynamics process covariance matrix is
\begin{align*}
Q=\begin{bmatrix}
0.01 &0 &0\\
0 &0.01 &0\\
0 &0 &0.001
\end{bmatrix}.
\end{align*}
The measurement error variances are $\rho_r=0.05 \text{m}^2$~(range),
 $\rho_b=0.05 \text{rad}^2$ (bearing), $\rho_{GPS}=1\text{m}^2$ (GPS-position), and $\rho_{GPS}=1\text{rad}^2$ (GPS-orientation).
The initial conditions are  
\begin{align*}
x_1=\begin{bmatrix}
0 \\ 
0 \\ 
0 
\end{bmatrix}, \ \ 
x_2=\begin{bmatrix}
-5 \\ 
7 \\ 
\pi/2
\end{bmatrix}, \ \ 
x_3=\begin{bmatrix}
5 \\ 
12 \\ 
\pi/2 
\end{bmatrix}, \ \ 
&x_4=\begin{bmatrix}
5 \\ 
-12 \\ 
0 
\end{bmatrix}, \ \ 
x_5=\begin{bmatrix}
-5 \\ 
-7 \\ 
0 
\end{bmatrix}, \ \ 
x_6=\begin{bmatrix}
0\\ 
17\\ 
-\pi/2
\end{bmatrix}, \\
P_1=P_2= &\cdots = P_6=\begin{bmatrix}
1 &0 &0\\
0 &1 &0\\
0 &0 &0.1
\end{bmatrix}.
\end{align*}

The simulation results show that communication topology plays an important role in the performance of the algorithm. Two main aspects of the 6-robots simulations, observability and covariance intersection effects, are considered next.

\subsubsection*{Example 3.A: Observability issues}
Depending on the communication topology and GPS availability to the robots, the state of the whole network in an absolute reference frame may be impossible to determine. The ability to pin down a subset of robots does not necessarily lead to the ability to do so for all team members. In our algorithm, measurements are shared between communicating pairs of robots, but they are not passed over to additional members. 

Observability of the full network state is heavily dependent on the measurement sharing topology. For instance, in the bridge topology the 2 subsets of robots have full communication internally (every robot talks with the rest of the robots in the subset), but there is only one link between the two groups. This means if that link fails or is lost, the two groups become blind to one another. A more extreme case is the chain topology, where the network diameter is largest. For this graph, if information is desired to go from the robot at one edge to the robot at the other edge, all robots in between have to successfully receive and transmit that information at some point, which takes time and makes the system more vulnerable to data drops. An example of such valuable information that would potentially be shared amongst all robots is GPS localization, since the problem of interest is that of robot localization in a global reference frame. On the other side, the star graph has a smaller network diameter, which benefits the system as a whole, but is dangerous for robots individually, since if one of the links breaks, an robot becomes isolated. Complexity increases in situations where GPS is limited or restricted to certain robots and, overall, the particular application dictates which topology should be used, and different communication graphs work better in different scenarios. 

\begin{figure}[h!]
 \centering 
\subfigure[]{
  \includegraphics[width=.45\linewidth]{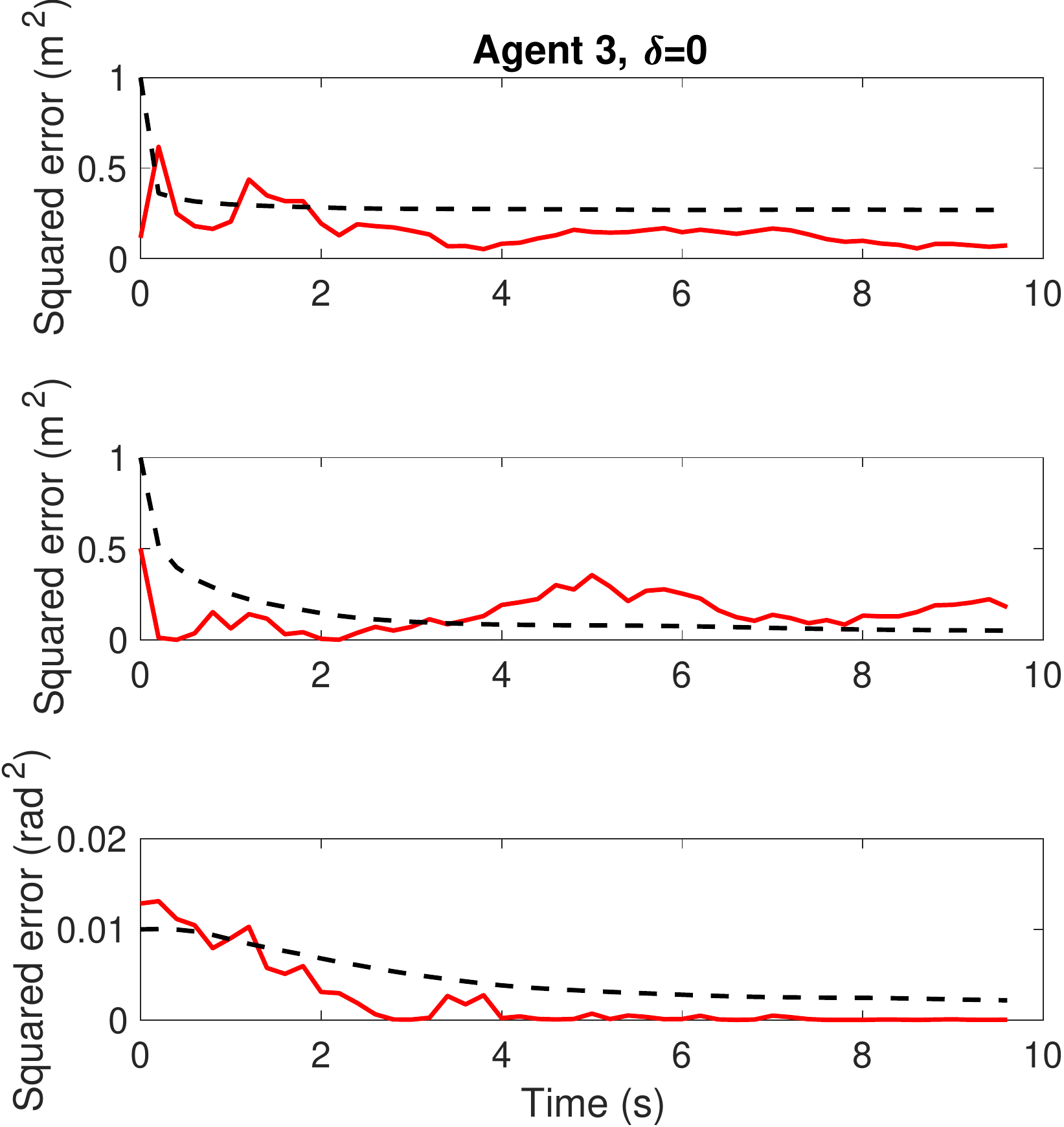}}
\subfigure[]{
  \includegraphics[width=.45\linewidth]{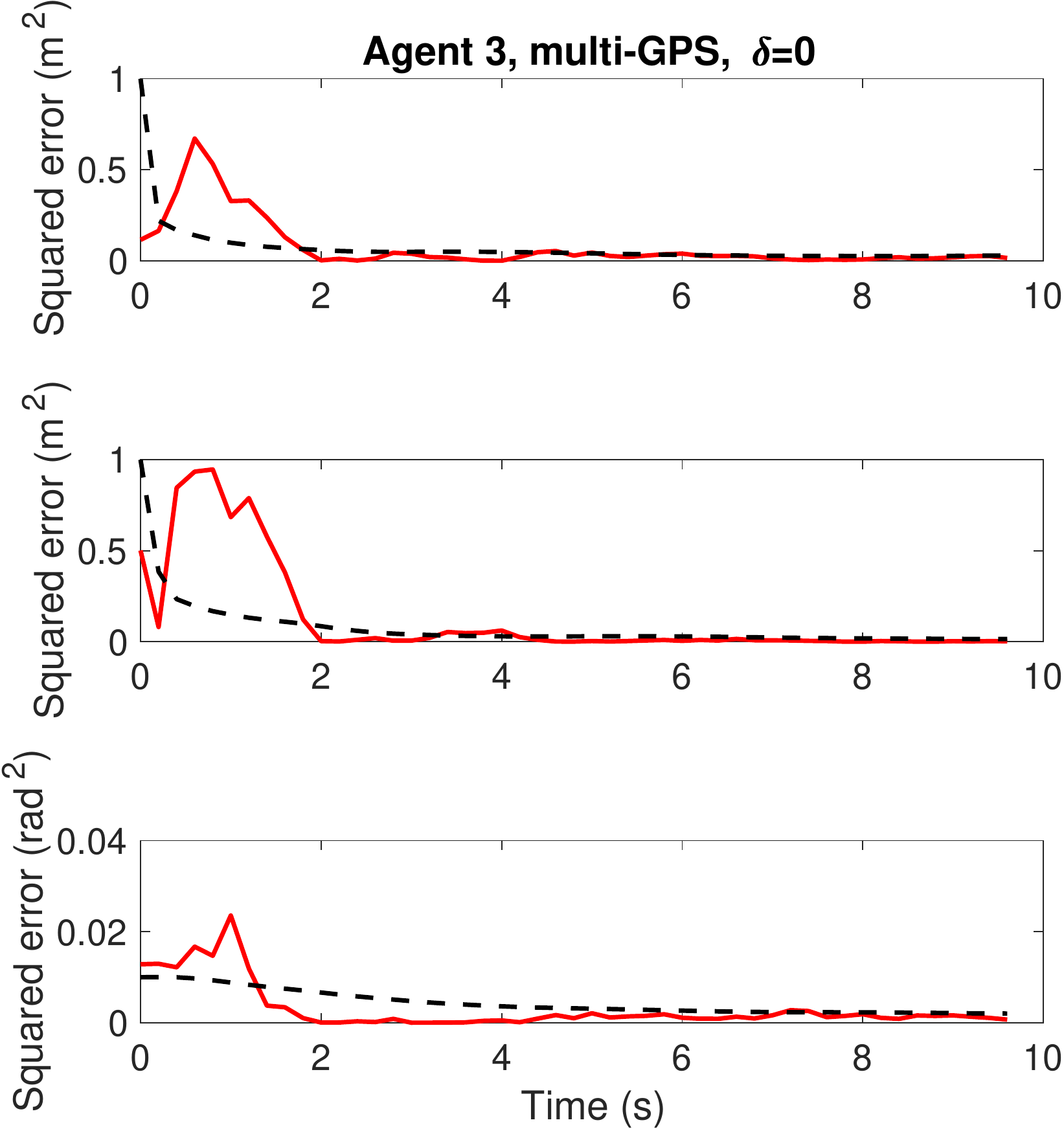}}
\caption{Robot 3's estimates in chain graph: (a) one GPS robot; (b) two GPS robots.}
\label{fig:one_vs_multi}
\end{figure}

There are different ways to cope with observability issues. Well connected networks present fewer problems when it comes to this due to more links existing between robots, at the expense of having increased communication costs and, possibly, having to use higher bandwidths and more expensive hardware. Another option is to have GPS or other forms of absolute positioning available to more robots, or to robots in strategical positions of the network. As an example, let us examine a case with a chain graph. Here, it can be seen that even if the robot with GPS access changes, most robots are still going to be out of reach, not being able to benefit from that information. For the chain graph, Figure \ref{fig:one_vs_multi} (a) shows a case where only one of the robots (robot 4, in the edge of the graph) gets GPS measurements. Here, the components of the estimated state drift and the filter is unable to correct it if only one robot gets GPS measurements. In (b) this behavior is corrected by adding additional robots (robots 1 and 6) that receive GPS measurements as well.

However, in some cases, such as UUV's, this may be impractical, since vehicles have to surface to get absolute measurements, thus incurring in additional energy consumption and losing valuable mission time. In more extreme scenarios this practice is not even possible, depending on mission requirements, GPS signal availability or hardware used. Another method, studied here, is covariance intersection (or CI), and allows robot pairs to fuse their state estimates and covariances with the goal of reducing uncertainty and obtaining an estimate that outperforms the other two. In this way, accurate state estimates from an robot that benefits from absolute measurements flow through the network.

\subsubsection*{ Example 3.B Effect of covariance intersection}
Covariance intersection is used as a way for robots to synchronize their estimates and reduce their covariance matrices. It is worth noting that CI involves higher communication and computational costs than performing a traditional Kalman measurement update, so it is not used as the main method for data sharing and fusion, but rather to overcome the problems associated with less observable networks by triggering it when the trace of the covariance matrix for a particular robot exceeds a certain threshold. The value of this threshold is a design choice in which different metrics and problem requirements have to be considered.

\begin{figure}[b!]
 \centering \subfigure[]{
   \includegraphics[width=.4\linewidth]{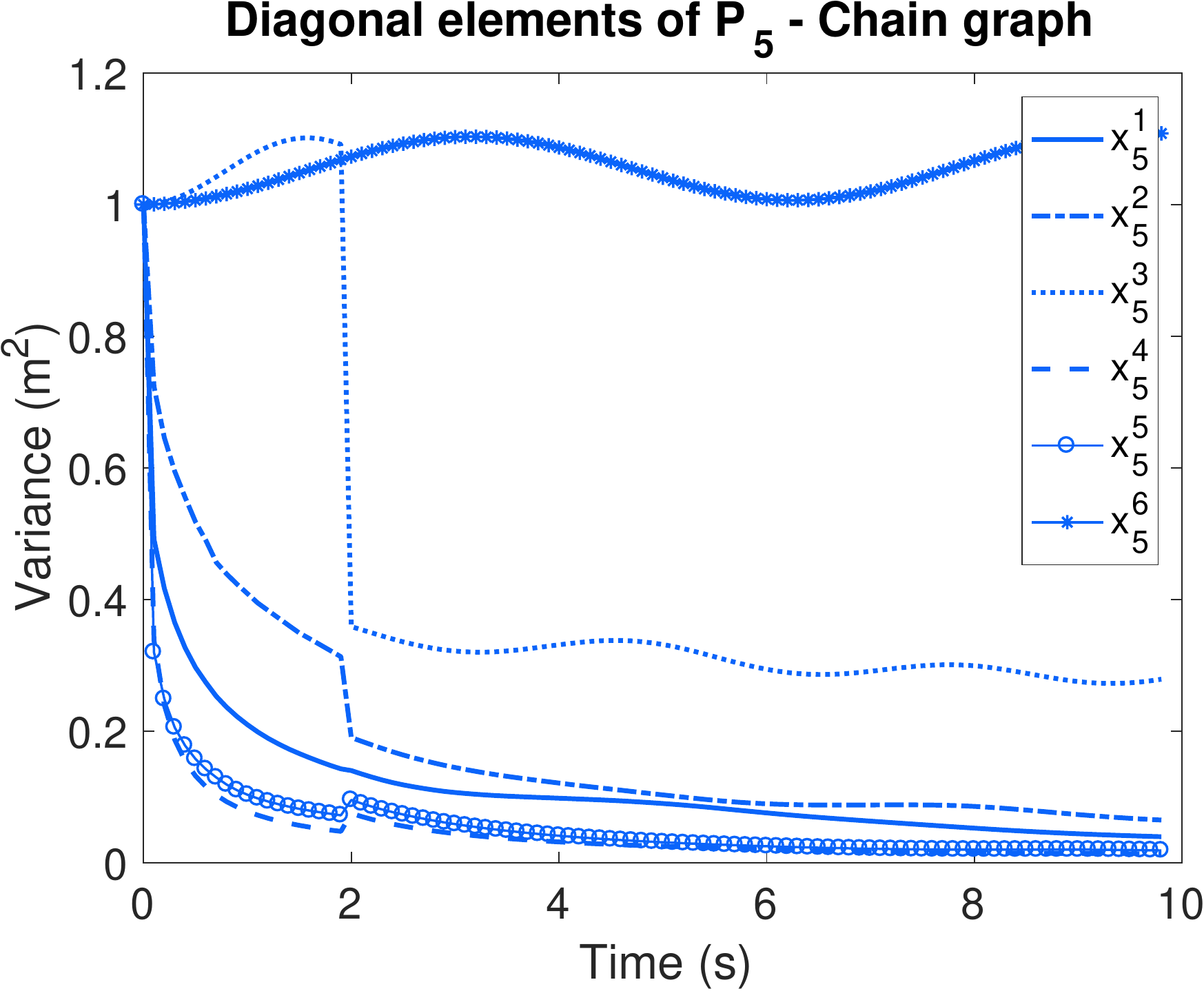}}
   \subfigure[]{
    \includegraphics[width=.4\linewidth]{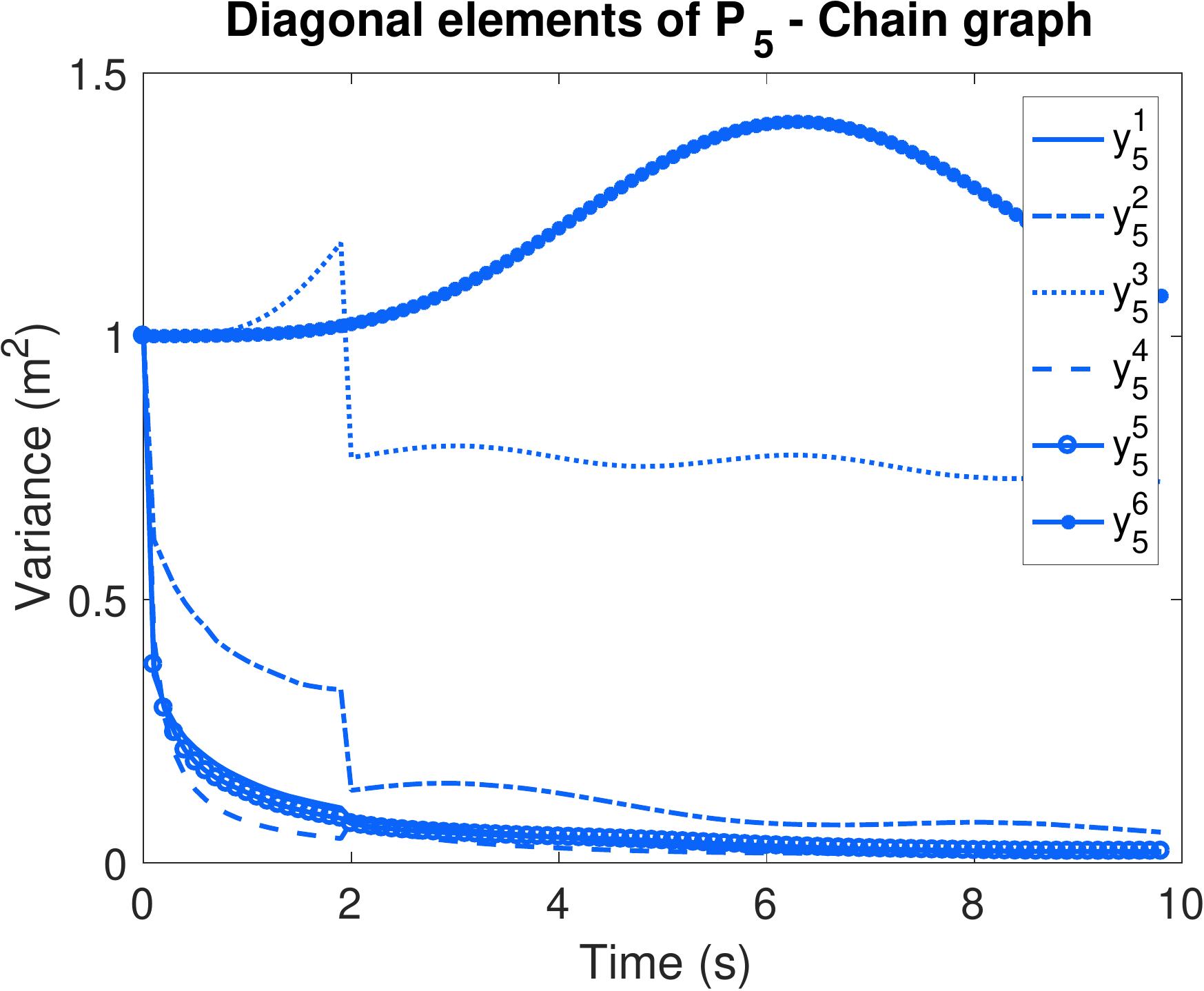}}
    \subfigure[]{
     \includegraphics[width=.4\linewidth]{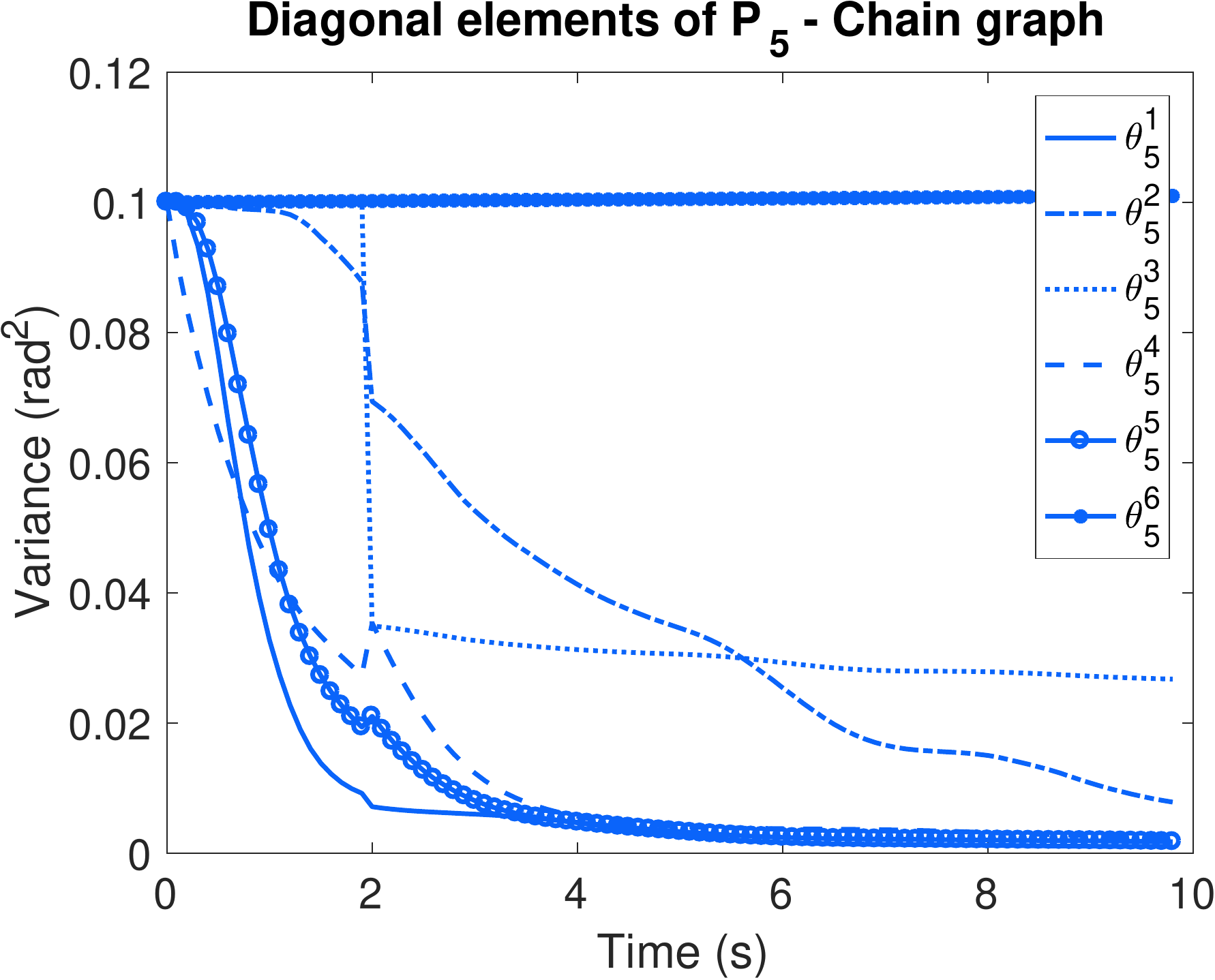}}
\caption{Typical variances estimated by robot 5 in chain graph; sudden spikes are CI updates.}
\label{fig:ci_fusion}
\end{figure}

Figure \ref{fig:ci_fusion} shows the effect that performing CI has on diagonal elements of the covariance matrix of an arbitrary robot. The sinusoidal shape of the curves is caused by the circular motion of the robots. This figure depicts a simulation where the robots communicate with one another following a chain-shaped (or line) graph, as can be seen in Figure \ref{fig:6agent-scheme}. By introducing additional correlations between robots' states, the resulting covariance matrix is generally filled with non-zero values in corresponding off-diagonal elements. CI effects manifest by generally sudden jumps in the covariance values, as can be seen around $t=4\text{s}$ or $t=5.2\text{s}$. These correlations are passed from robot to robot every time that CI is performed. However, states corresponding to two robots that are separated by several links in the communication graph will not become correlated instantaneously. Instead, it will take intermediate robots to perform CI successively for a few time steps. Delays in the propagation of covariance intersection correlations are hard to see in these simulations, since the network is relatively small.

Another interesting aspect of these simulations is the fact that the covariance matrix hits an upper bound even in cases where there are very sparse or no measurements containing information about specific robots. Then, every time covariance intersection is performed new correlations between states are added, which reflects in this bound adopting a different value. Figure \ref{fig:ci_fusion} shows the variances of the states of all robots, as estimated by robot 5. As can be seen, there is a direct correspondance between the number of links separating robot 5 and the other robots and how large the variances are -- for example, robot 6 is the farthest away from 5 (in terms of links or connections) and the associated variance for 6 is the largest, whereas robot 2 is measured by 5 directly, so its covariance is small.

\begin{figure}[b!]
 \centering \subfigure[]{
   \includegraphics[width=.475\linewidth]{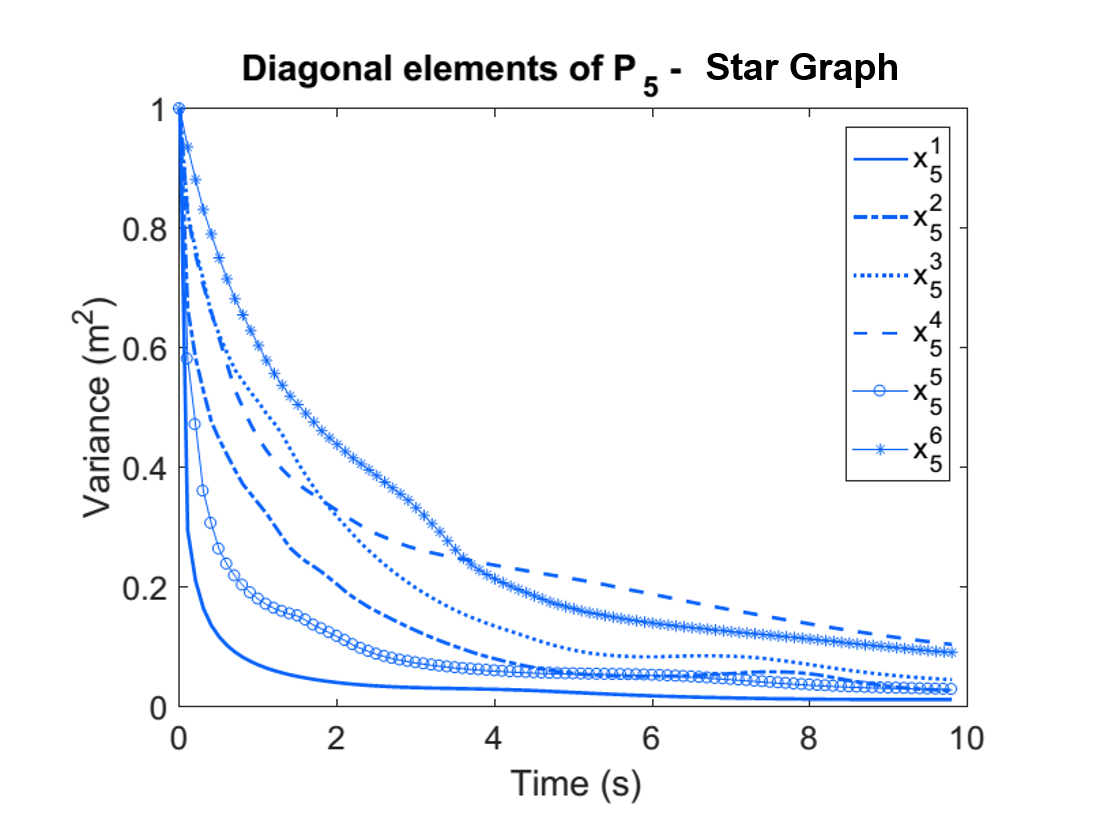}}
   \subfigure[]{
    \includegraphics[width=.475\linewidth]{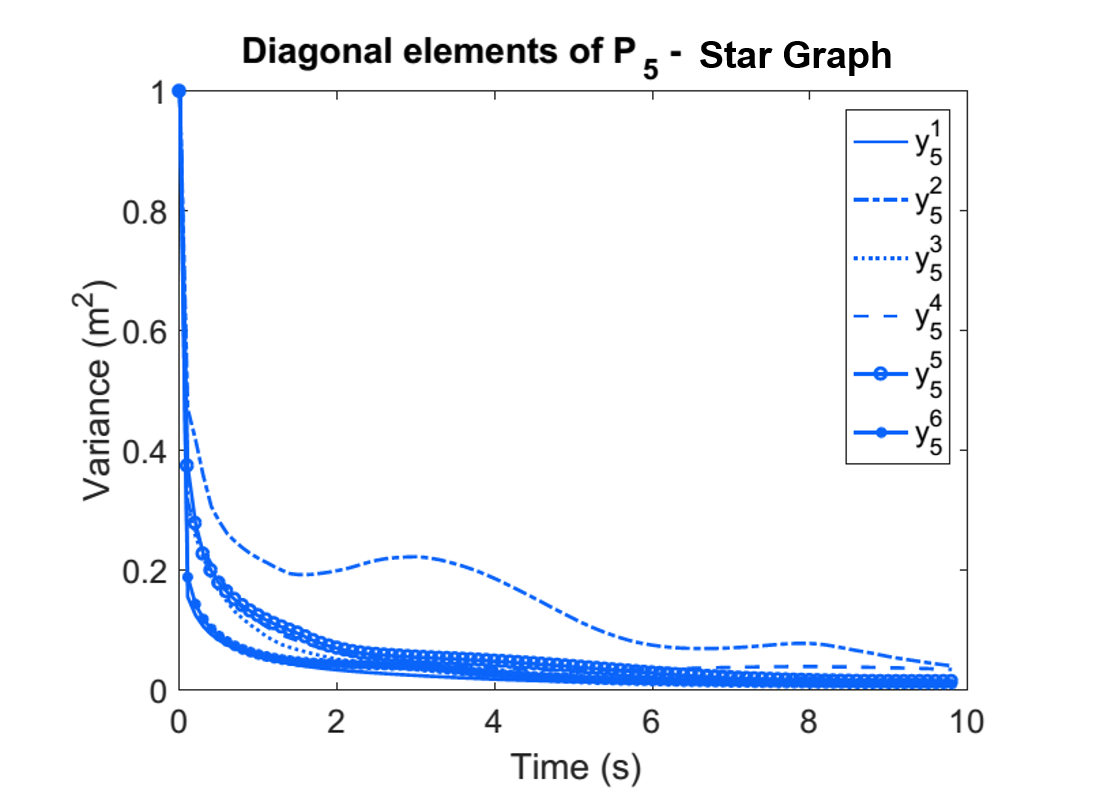}}
    \subfigure[]{
     \includegraphics[width=.475\linewidth]{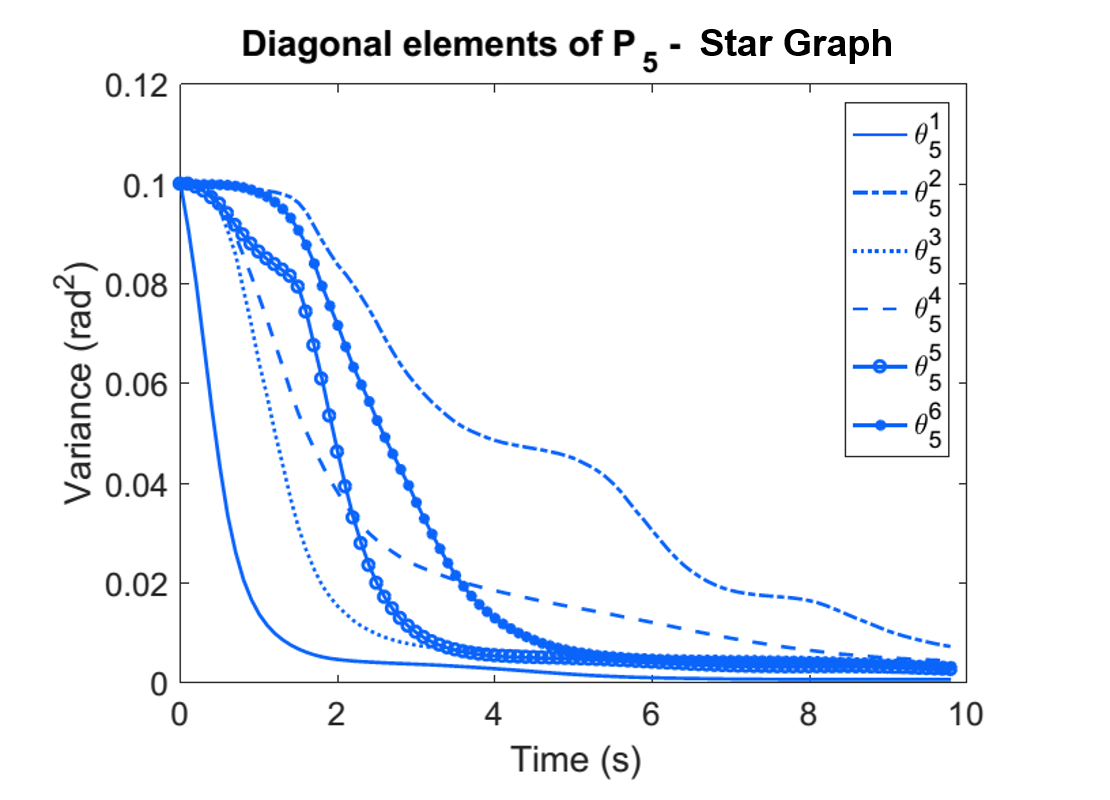}}
\caption{Typical variances estimated by robot 5 in star graph; no CI updates occur.}
\label{fig:ci_fusion_star}
\end{figure}

Covariance intersection is not needed in networks where all robots' states are measured by or shared between one another, since if the filter is properly tuned it will eventually converge. Figure \ref{fig:ci_fusion_star}, where robots communicate following a star-shaped graph, shows that because the only robot that is receiving GPS measurements (robot 1) acts as a hub and is able to share them with the rest of the network, all other states can be uniquely estimated by virtue of pinning down robot 1. On the other hand, if GPS measurements were provided only to one of the robots that act as leaf nodes (that is, all robots except for robot 1), it would become necessary to perform CI to prevent the estimates from drifting away from the true states. 

One important conclusion in multiple robot scenarios is that the number of robots that have access to GPS measurements, as well as their position and ability to communicate with other robots, affects the overall performance of the filter. Additionally, in view of the results it becomes clear that CI plays a dominant role in cooperative localization, and correlations between robots states introduced early on in the simulations have a long-lasting effect that allows a reduction in communication costs without strong penalizations on filter performance. This is particularly useful in poorly connected graphs, although it comes at a price -- performing CI more often brings about higher communication costs, which is counterproductive, so an optimal combination of these two values needs to be obtained. This is a problem of its own that would be worth studying in future works.

\section{Conclusions}
This paper presented a novel decentralized Cooperative Localization algorithm for a
team of robotic robots to estimate the state of the network using sparse 
event-triggered measurement communication. 
Since robots all know the event-triggering condition, censored measurements can be interpreted
and fused as set-valued information into state estimates at low communication cost. 
The algorithm also occasionally employs Covariance Intersection estimate fusion, to keep the weighted trace of the error covariance bounded when robots cannot directly measure all other robots. 
Since it has higher communication and computation cost as compared to sending a combination of explicit/implicit measurements, the Covariance Intersection process is only triggered sporadically as required by the performance goal. 
Simulations showed the effectiveness of the mixed implicit/explicit data fusion strategy for linear and nonlinear dynamics and measurement models using extended Kalman filters. 
Simulations also showed that mixed implicit/explicit measurement fusion can provide performance that is nearly equivalent to total explicit data communication and fusion, but requires far less data to be exchanged between robots (under ideal communication conditions).  
This work also studied measurement erasure over lossy communication channels has on the consistency and performance, due to the algorithmic assumption of lossless communications. 
Results revealed that estimation consistency is improved in lossy communication channels as the innovation threshold parameter size is increased, at the cost of elevated mean squared error.  
The simulations also explored several communication topologies and illustrated the benefit of sporadic Covariance Intersection when absolute GPS measurements are not available to all robots in certain topologies. 
These simulations results focused on relatively small teams of robots, but nevertheless provide useful benchmarks for typical applications of Cooperative Localization with intelligent autonomous vehicle systems (as opposed to swarms of robots with highly limited sensing and computing power). 

Open directions for follow-on work include determining if Bayesian channel filters can be used
when the communication topology is fixed to further improve
performance~\cite{SG-HRDW:94}; extending the algorithm to correlated
non-scalar measurements; investigating Covariance Intersection schemes
that allow the fusion of only a subset of the network's states~
\cite{SJJ-JKU:97, Ahmed-FUSION-2016} so each robot may only estimate its own and its neighbors' locations; and formally analyzing the behavior of the thresholding dynamics. 
Experimental comparisons to other decentralized Cooperative Localization strategies should also be performed to formally assess the relative strengths and weaknesses for different application scenarios. 

The proposed algorithm could also be extended to handle communication dropouts that are incorrectly assumed to be implicit information, since our simulations have confirmed that performance degrades when this assumption is violated under heavy communication losses. For instance, the fact that it will be statistically highly unlikely for an entire measurement vector to be censored via event-triggered estimation can be used by robots to formally detect when an explicit measurement has been completely dropped by senders. 
It can also be assumed that the robots' control is unknown a priori but is also sent using an event-triggering strategy; 
then, it may be possible to investigate how the interplay of both mechanisms affects performance. 

Another relevant direction is formal analysis of event-triggered cooperative estimator performance in the presence of measurement outliers and corrupted measurements. These types of measurements were ignored here, but can still arise in many practical scenarios. Outliers correspond to measurement data whose random error statistics are not accurately described by the estimator's underlying data likelihood model. These can often be attributed to artifacts of noise model mismatch, e.g. if random sensor errors are modeled as Gaussian, but truly follow a non-Gaussian distribution such as a multi-modal/heavy-tailed Gaussian mixture model. 
Corrupt measurements are those which arise from non-ideal (though not necessarily unpredictable) sensor behavior, e.g. signal saturation or cut-off above some known threshold. 
Outliers and corrupt data could cause explicit measurement updates to occur more often than they should in an event-triggered estimator (such that the filter will be `surprised' more often than its own model predicts), and as a result communication savings would be compromised. This could lead to large errors in the a posteriori state estimation statistics following either implicit or explicit measurement updates, since the Kalman gain will not be based on the correct noise statistics. 
There are at least two different ways to handle such data: via deterministic heuristic gating methods (hard classifications of measurement innovations); or via formal probabilistic data fusion, which use more accurate models of all potential sensor error sources to automatically weight sensor information in Bayesian state estimator updates \cite{Blackman-AESMag-2004}. Since the effectiveness of these methods in an event-triggered setting is highly scenario-dependent (and since outliers/corrupt sensor data are not inherent to the basic decentralized cooperative localization problem), a detailed description and analysis is left for future work. 

Formal analysis of stability and convergence characteristics for the decentralized event-triggered cooperative localization algorithm can also be further examined in the case of linear vehicle dynamics and sensor models. To our knowledge, no formal stability results have yet been established for decentralized event-triggered state estimators of the type considered here. 
However, work by \cite{Khojasteh-arxiv-2018} and \cite{Trimpe-CDC-2014} provide promising results for closely related problems that could be adapted and applied for linear systems. 
Stability and convergence guarantees are unlikely to exist for nonlinear systems via the EKF, since the EKF relies heavily on the validity of linearization assumptions (which will vary by application and operating conditions). The EKF has been long established as a core algorithm in the robot state estimation and navigation literature despite this limitation. 
Yet, it should be understood that, like all EKFs, the event-triggered EKF developed here for cooperative localization requires proper initialization and tuning to ensure good performance and stable/non-diverging behavior, i.e. such that the state estimates about which dynamics and measurements are repeatedly re-linearized remain close to the actual system state.

\section*{Acknowledgments}
David Iglesias was supported by a fellowship from the Balsells Fellowship Program. 

\section*{References}
\bibliography{alias,Main-add}

 \end{document}